\documentclass[12pt,twoside]{article}
\usepackage{epsfig}
\usepackage{makeidx}
\usepackage{varioref}
\usepackage{amssymb}
\usepackage{longtable}

\input dina4.sty

\usepackage{url}

\newlength{\mybibitemsep}
\newcommand\setmybibitemsep[1]{\setlength{\mybibitemsep}{#1}}

\setmybibitemsep{0.5\topsep}

\newcommand\includenetreferences{y}

\newcommand\referencessize{\normalsize}

\newcommand\mybibbaselinestretch{0.96}

\newcommand\mybibsection[1]{\section*{#1}\if 
\addcontentslineofbibsection\addcontentsline{toc}{section}{#1}\fi}
\newcommand\addcontentslineofbibsection{y}

\newcommand\mybibtitle[1]{{\em #1\@.}}
\newcommand\mybibhardnodate
[1]{{\def\myfirstarg{#1}\ifx\empty\myfirstarg{}\else#1.\ \fi}}
\newcommand\mybibsoft
[2]{{\def\myfirstarg{#1}\def\mysecondarg{#2}\ifx\empty\myfirstarg{}\else
\if y\includenetreferences\writemynetsource#1\else\if x\includenetreferences
\ifx\empty\mysecondarg\writemynetsource#1\else{}\fi\else
\if n\includenetreferences{}\else\typeout
{WARNING includenetreferences from headerbiblio.tex has silly definition}
\fi\fi\fi\fi}}

\def\writemynetsource[#1,#2,#3,#4://#5]{{\tt\sloppy\ 
\url{#4://#5} \discretionary
{(\ignorespaces#2\,\ignorespaces#1\mbox{$\!$},\mbox{$\!$}}%
{\ignorespaces#3)\mbox{$\!$}.}%
{(\ignorespaces#2\,\ignorespaces#1\mbox{$\!$},\,\ignorespaces
#3)\mbox{$\!$}.}}}

\catcode`\@=11

\newcommand\resetlongbibstyle{%
\def\bibitem{\@ifnextchar[\@lbibitem\@bibitem}
\def\@lbibitem[##1]##2{\item[\@biblabel{##1}\hfill]\if@filesw
      {\let\protect\noexpand
       \immediate
       \write\@auxout{\string\bibcite{##2}{##1}}}\fi\ignorespaces}
\def\@bibitem##1{\item\if@filesw \immediate\write\@auxout
       {\string\bibcite{##1}{\the\value{\@listctr}}}\fi\ignorespaces}
\def\bibcite{\@newl@bel b}
\let\citation\@gobble
\let\bibdata=\@gobble
\let\bibstyle=\@gobble
\def\bibliography##1{%
  \if@filesw
    \immediate\write\@auxout{\string\bibdata{##1}}%
  \fi
  \@input@{\jobname.bbl}}
\def\bibliographystyle##1{%
  \ifx\@begindocumenthook\@undefined\else
    \expandafter\AtBeginDocument
  \fi
    {\if@filesw
       \immediate\write\@auxout{\string\bibstyle{##1}}%
     \fi}}
\def\nocite##1{\@bsphack
  \@for\@citeb:=##1\do{%
    \edef\@citeb{\expandafter\@firstofone\@citeb}%
    \if@filesw\immediate\write\@auxout{\string\citation{\@citeb}}\fi
    \@ifundefined{b@\@citeb}{\G@refundefinedtrue
        \@latex@warning{Citation `\@citeb' undefined}}{}}%
  \@esphack}
\expandafter\let\csname b@*\endcsname\@empty
\DeclareRobustCommand\cite{%
  \@ifnextchar [{\@tempswatrue\@citex}{\@tempswafalse\@citex[]}}
\DeclareRobustCommand\citet{%
  \@ifnextchar [{\@tempswatrue\@citex}{\@tempswafalse\@citex[]}}
\def\@tempswafalse{\let\if@tempswa\iffalse}
\def\@tempswatrue{\let\if@tempswa\iftrue}
\let\if@tempswa\iffalse
\def\@cite##1##2{##1\if@tempswa , ##2\fi}
\def\@citex[##1]##2{%
  \let\@citea\@empty
  \@cite{\@for\@citeb:=##2\do
    {\@citea\def\@citea{,\penalty\@m\ }%
     \edef\@citeb{\expandafter\@firstofone\@citeb}%
     \if@filesw\immediate\write\@auxout{\string\citation{\@citeb}}\fi
     \@ifundefined{b@\@citeb}{\mbox{\reset@font\bfseries ?}%
       \G@refundefinedtrue
       \@latex@warning
         {Citation `\@citeb' on page \thepage \space undefined}}%
       {\csname b@\@citeb\endcsname}}}{##1}}
\def\@biblabel##1{##1}
\def\mybibitem##1##2##3##4##5##6##7##8##9
{\item
 \if@filesw
      {\let\protect\noexpand
       \immediate
       \write\@auxout{\string\bibcite{##1}{##7\discretionary
{}{}{\,}(##3##9)}}}\fi\ignorespaces
##2 (##3##9). \mybibtitle{##4} \mybibhardnodate{##5}\mybibsoft
{##6}{##5}$\!\!$\par}
\def\thebibliography##1{\mybibsection{\refname}%
\@mkboth{\uppercase{\refname}}{\uppercase{\refname}}
\def\baselinestretch{\mybibbaselinestretch}%
\list{}{\labelwidth\z@
    \leftmargin 1.5pc
    \itemindent-\leftmargin}
    \referencessize
    \parindent\z@
    \parskip\mybibitemsep\relax
    \def\newblock{\hskip .11em plus .33em minus .07em}
    \sloppy\clubpenalty4000\widowpenalty4000
    \sfcode`\.=1000\relax}
\let\endthebibliography=\endlist
}

\newcommand\resetshortbibstyle{%
\def\bibitem{\@ifnextchar[\@lbibitem\@bibitem}
\def\@lbibitem[##1]##2{\item[\@biblabel{##1}\hfill]\if@filesw
      {\let\protect\noexpand
       \immediate
       \write\@auxout{\string\bibcite{##2}{##1}}}\fi\ignorespaces}
\def\@bibitem##1{\item\if@filesw \immediate\write\@auxout
       {\string\bibcite{##1}{\the\value{\@listctr}}}\fi\ignorespaces}
\def\bibcite{\@newl@bel b}
\let\citation\@gobble
\def\@citex[##1]##2{%
  \let\@citea\@empty
  \@cite{\@for\@citeb:=##2\do
    {\@citea\def\@citea{,\penalty\@m\ }%
     \edef\@citeb{\expandafter\@firstofone\@citeb}%
     \if@filesw\immediate\write\@auxout{\string\citation{\@citeb}}\fi
     \@ifundefined{b@\@citeb}{\mbox{\reset@font\bfseries ?}%
       \G@refundefinedtrue
       \@latex@warning
         {Citation `\@citeb' on page \thepage \space undefined}}%
       {\hbox{\csname b@\@citeb\endcsname}}}}{##1}}
\let\bibdata=\@gobble
\let\bibstyle=\@gobble
\def\bibliography##1{%
  \if@filesw
    \immediate\write\@auxout{\string\bibdata{##1}}%
  \fi
  \@input@{\jobname.bbl}}
\def\bibliographystyle##1{%
  \ifx\@begindocumenthook\@undefined\else
    \expandafter\AtBeginDocument
  \fi
    {\if@filesw
       \immediate\write\@auxout{\string\bibstyle{##1}}%
     \fi}}
\def\nocite##1{\@bsphack
  \@for\@citeb:=##1\do{%
    \edef\@citeb{\expandafter\@firstofone\@citeb}%
    \if@filesw\immediate\write\@auxout{\string\citation{\@citeb}}\fi
    \@ifundefined{b@\@citeb}{\G@refundefinedtrue
        \@latex@warning{Citation `\@citeb' undefined}}{}}%
  \@esphack}
\expandafter\let\csname b@*\endcsname\@empty
\def\@cite##1##2{[{##1\if@tempswa , ##2\fi}]}
\def\@biblabel##1{[##1]}
\DeclareRobustCommand\cite{%
  \@ifnextchar [{\@tempswatrue\@citex}{\@tempswafalse\@citex[]}}
\def\@tempswafalse{\let\if@tempswa\iffalse}
\def\@tempswatrue{\let\if@tempswa\iftrue}
\let\if@tempswa\iffalse
\def\mybibitem##1##2##3##4##5##6##7##8##9
{\bibitem[##8##9]{##1}##2 (##3). 
\mybibtitle{##4} \mybibhardnodate{##5}\mybibsoft{##6}{##5}$\!\!$\par}
\def\thebibliography##1{\mybibsection{\refname}%
\@mkboth{\uppercase{\refname}}{\uppercase{\refname}}%
\def\baselinestretch{\mybibbaselinestretch}%
\referencessize
\list
 {[\arabic{enumi}]}
 {\settowidth\labelwidth{[##1]}\leftmargin\labelwidth
 \advance\leftmargin\labelsep
 \usecounter{enumi}}
 \parskip\mybibitemsep\relax
 \def\newblock{\hskip .11em plus .33em minus .07em}
 \sloppy\clubpenalty4000\widowpenalty4000
 \sfcode`\.=1000\relax}
\let\endthebibliography=\endlist
}

\newcommand\resetnumberbibstyle{%
\def\bibitem{\@ifnextchar[\@lbibitem\@bibitem}
\def\@lbibitem[##1]##2{\item[\@biblabel{##1}\hfill]\if@filesw
      {\let\protect\noexpand
       \immediate
       \write\@auxout{\string\bibcite{##2}{##1}}}\fi\ignorespaces}
\def\@bibitem##1{\item\if@filesw \immediate\write\@auxout
       {\string\bibcite{##1}{\the\value{\@listctr}}}\fi\ignorespaces}
\def\bibcite{\@newl@bel b}
\let\citation\@gobble
\def\@citex[##1]##2{%
  \let\@citea\@empty
  \@cite{\@for\@citeb:=##2\do
    {\@citea\def\@citea{,\penalty\@m\ }%
     \edef\@citeb{\expandafter\@firstofone\@citeb}%
     \if@filesw\immediate\write\@auxout{\string\citation{\@citeb}}\fi
     \@ifundefined{b@\@citeb}{\mbox{\reset@font\bfseries ?}%
       \G@refundefinedtrue
       \@latex@warning
         {Citation `\@citeb' on page \thepage \space undefined}}%
       {\hbox{\csname b@\@citeb\endcsname}}}}{##1}}
\let\bibdata=\@gobble
\let\bibstyle=\@gobble
\def\bibliography##1{%
  \if@filesw
    \immediate\write\@auxout{\string\bibdata{##1}}%
  \fi
  \@input@{\jobname.bbl}}
\def\bibliographystyle##1{%
  \ifx\@begindocumenthook\@undefined\else
    \expandafter\AtBeginDocument
  \fi
    {\if@filesw
       \immediate\write\@auxout{\string\bibstyle{##1}}%
     \fi}}
\def\nocite##1{\@bsphack
  \@for\@citeb:=##1\do{%
    \edef\@citeb{\expandafter\@firstofone\@citeb}%
    \if@filesw\immediate\write\@auxout{\string\citation{\@citeb}}\fi
    \@ifundefined{b@\@citeb}{\G@refundefinedtrue
        \@latex@warning{Citation `\@citeb' undefined}}{}}%
  \@esphack}
\expandafter\let\csname b@*\endcsname\@empty
\def\@cite##1##2{[{##1\if@tempswa , ##2\fi}]}
\def\@biblabel##1{[##1]}
\DeclareRobustCommand\cite{%
  \@ifnextchar [{\@tempswatrue\@citex}{\@tempswafalse\@citex[]}}
\def\@tempswafalse{\let\if@tempswa\iffalse}
\def\@tempswatrue{\let\if@tempswa\iftrue}
\let\if@tempswa\iffalse
\def\mybibitem##1##2##3##4##5##6##7##8##9
{\bibitem{##1}##2 (##3). 
\mybibtitle{##4} \mybibhardnodate{##5}\mybibsoft{##6}{##5}$\!\!$\par}
\def\thebibliography##1{\mybibsection{\refname}%
\@mkboth{\uppercase{\refname}}{\uppercase{\refname}}
\def\baselinestretch{\mybibbaselinestretch}%
\referencessize
\list
 {[\arabic{enumi}]}
 {\settowidth\labelwidth{[##1]}\leftmargin\labelwidth
 \advance\leftmargin\labelsep
 \usecounter{enumi}}
 \parskip\mybibitemsep\relax
 \def\newblock{\hskip .11em plus .33em minus .07em}
 \sloppy\clubpenalty4000\widowpenalty4000
 \sfcode`\.=1000\relax}
\let\endthebibliography=\endlist
}

\newcommand\setshortbibstyle {\AtBeginDocument{\resetshortbibstyle }}

\catcode`\@=12

\input headertheorem
\input headerhot
\usepackage{amssymb}

\mathchardef\Gammaoffont="7000
\mathchardef\Gamma="0100
\mathchardef\Deltaoffont="7001
\mathchardef\Delta="0101
\mathchardef\Thetaoffont="7002
\mathchardef\Theta="0102
\mathchardef\Lambdaoffont="7003
\mathchardef\Lambda="0103
\mathchardef\Xioffont="7004
\mathchardef\Xi="0104
\mathchardef\Pioffont="7005
\mathchardef\Pi="0105
\mathchardef\Sigmaoffont="7006
\mathchardef\Sigma="0106
\mathchardef\Upsilonoffont="7007
\mathchardef\Upsilon="0107
\mathchardef\Phioffont="7008
\mathchardef\Phi="0108
\mathchardef\Psioffont="7009
\mathchardef\Psi="0109
\mathchardef\Omegaoffont="700A
\mathchardef\Omega="010A
\mathchardef\itype="017B

\catcode`\@=11

\gdef\allowhyphens{\penalty\@M \hskip\z@skip}

\gdef\set@low@box#1{\setbox\tw@\hbox{,}\setbox\z@\hbox{#1}\dimen\z@\ht\z@
     \advance\dimen\z@ -\ht\tw@
     \setbox\z@\hbox{\lower\dimen\z@ \box\z@}\ht\z@\ht\tw@ \dp\z@\dp\tw@ }
\gdef\set@low@boxsingle#1{\setbox\tw@\hbox{\rm,}\setbox\z@\hbox{#1}\dimen\z@\ht\z@
     \advance\dimen\z@ -\ht\tw@
     \setbox\z@\hbox{\lower\dimen\z@ \box\z@}\ht\z@\ht\tw@ \dp\z@\dp\tw@ }

\gdef\@glqq{%
\ifhmode\edef\@SF{\spacefactor\the\spacefactor}%
\else\let\@SF\empty
\fi
\CheckFamily\font\fraknomath\ifSameFamily ``\relax
\else\CheckFamily\font\swab\ifSameFamily ``\relax
\else\leavevmode\set@low@box{''}\box\z@\kern-.04em\allowhyphens\@SF\relax
\fi\fi}
\gdef\glqq{\protect\@glqq\kern+.07em}
\gdef\@grqq{%
\ifhmode\edef\@SF{\spacefactor\the\spacefactor}%
\else\let\@SF\empty 
\fi 
\CheckFamily\font\fraknomath\ifSameFamily ''\relax
\else\CheckFamily\font\swab\ifSameFamily ''\relax
\else\kern+.07em``\kern.07em\@SF\relax
\fi\fi}
\gdef\grqq{\protect\@grqq}
\gdef\@glq{{\ifhmode \edef\@SF{\spacefactor\the\spacefactor}\else
     \let\@SF\empty \fi \leavevmode
     \set@low@boxsingle{'\/}\box\z@\kern-.04em\allowhyphens\@SF\relax}}
\gdef\glq{\protect\@glq\kern+.07em}
\gdef\@grq{\ifhmode \edef\@SF{\spacefactor\the\spacefactor}\else
     \let\@SF\empty \fi \kern-.0125em`\kern.07em\@SF\relax}
\gdef\grq{\protect\@grq}

\catcode`\@=12

\makeatletter
\if \@ptsize 0
   \newfont{\scriptscriptscriptgoth}{ygoth scaled 760}
   \newfont{\scriptscriptgoth}{ygoth scaled 833}
   \newfont{\scriptgoth}{ygoth scaled 912}
   \newfont{\gothnomath}{ygoth}
   \newfont{\Goth}{ygoth scaled \magstephalf}
   \newfont{\GOth}{ygoth scaled \magstep1}
   \newfont{\GOTh}{ygoth scaled \magstep2}
   \newfont{\GOTH}{ygoth scaled \magstep3}

   \newfont{\scriptscriptscriptswab}{yswab scaled 760}
   \newfont{\scriptscriptswab}{yswab scaled 833}
   \newfont{\scriptswab}{yswab scaled 912}
   \newfont{\swab}{yswab}
   \newfont{\Swab}{yswab scaled \magstephalf}
   \newfont{\SWab}{yswab scaled \magstep1}
   \newfont{\SWAb}{yswab scaled \magstep2}
   \newfont{\SWAB}{yswab scaled \magstep3}

   \newfont{\scriptscriptscriptfrak}{yfrak scaled 760}
   \newfont{\scriptscriptfrak}{yfrak scaled 833}
   \newfont{\scriptfrak}{yfrak scaled 912}
   \newfont{\fraknomath}{yfrak}
   \newfont{\Frak}{yfrak scaled \magstephalf}
   \newfont{\FRak}{yfrak scaled \magstep1}
   \newfont{\FRAk}{yfrak scaled \magstep2}
   \newfont{\FRAK}{yfrak scaled \magstep3}

   \newfont{\init}{yinit}
   \newfont{\Init}{yinit scaled \magstephalf}
   \newfont{\INit}{yinit scaled \magstep1}
   \newfont{\INIt}{yinit scaled \magstep2}
   \newfont{\INIT}{yinit scaled \magstep3}
\fi
\if \@ptsize 1
   \newfont{\scriptscriptscriptgoth}{ygoth scaled 833}
   \newfont{\scriptscriptgoth}{ygoth scaled 912}
   \newfont{\scriptgoth}{ygoth}
   \newfont{\gothnomath}{ygoth scaled \magstephalf}
   \newfont{\Goth}{ygoth scaled \magstep1}
   \newfont{\GOth}{ygoth scaled \magstep2}
   \newfont{\GOTh}{ygoth scaled \magstep3}
   \newfont{\GOTH}{ygoth scaled \magstep4}

   \newfont{\scriptscriptscriptswab}{yswab scaled 833}
   \newfont{\scriptscriptswab}{yswab scaled 912}
   \newfont{\scriptswab}{yswab}
   \newfont{\swab}{yswab scaled \magstephalf}
   \newfont{\Swab}{yswab scaled \magstep1}
   \newfont{\SWab}{yswab scaled \magstep2}
   \newfont{\SWAb}{yswab scaled \magstep3}
   \newfont{\SWAB}{yswab scaled \magstep4}

   \newfont{\scriptscriptscriptfrak}{yfrak scaled 833}
   \newfont{\scriptscriptfrak}{yfrak scaled 912}
   \newfont{\scriptfrak}{yfrak}
   \newfont{\fraknomath}{yfrak scaled \magstephalf}
   \newfont{\Frak}{yfrak scaled \magstep1}
   \newfont{\FRak}{yfrak scaled \magstep2}
   \newfont{\FRAk}{yfrak scaled \magstep3}
   \newfont{\FRAK}{yfrak scaled \magstep4}

   \newfont{\init}{yinit scaled \magstephalf}
   \newfont{\Init}{yinit scaled \magstep1}
   \newfont{\INit}{yinit scaled \magstep2}
   \newfont{\INIt}{yinit scaled \magstep3}
   \newfont{\INIT}{yinit scaled \magstep4}
\fi
\if \@ptsize 2
   \newfont{\scriptscriptscriptgoth}{ygoth scaled 912}
   \newfont{\scriptscriptgoth}{ygoth}
   \newfont{\scriptgoth}{ygoth scaled \magstephalf}
   \newfont{\gothnomath}{ygoth scaled \magstep1}
   \newfont{\Goth}{ygoth scaled \magstep2}
   \newfont{\GOth}{ygoth scaled \magstep3}
   \newfont{\GOTh}{ygoth scaled \magstep4}
   \newfont{\GOTH}{ygoth scaled \magstep5}

   \newfont{\scriptscriptscriptswab}{yswab scaled 912}
   \newfont{\scriptscriptswab}{yswab}
   \newfont{\scriptswab}{yswab scaled \magstephalf}
   \newfont{\swab}{yswab scaled \magstep1}
   \newfont{\Swab}{yswab scaled \magstep2}
   \newfont{\SWab}{yswab scaled \magstep3}
   \newfont{\SWAb}{yswab scaled \magstep4}
   \newfont{\SWAB}{yswab scaled \magstep5}

   \newfont{\scriptscriptscriptfrak}{yfrak scaled 912}
   \newfont{\scriptscriptfrak}{yfrak}
   \newfont{\scriptfrak}{yfrak scaled \magstephalf}
   \newfont{\fraknomath}{yfrak scaled \magstep1}
   \newfont{\Frak}{yfrak scaled \magstep2}
   \newfont{\FRak}{yfrak scaled \magstep3}
   \newfont{\FRAk}{yfrak scaled \magstep4}
   \newfont{\FRAK}{yfrak scaled \magstep5}

   \newfont{\init}{yinit scaled \magstep1}
   \newfont{\Init}{yinit scaled \magstep2}
   \newfont{\INit}{yinit scaled \magstep3}
   \newfont{\INIt}{yinit scaled \magstep4}
   \newfont{\INIT}{yinit scaled \magstep5}
\fi

\makeatother

\newif\ifSameFamily
\def\CheckFamily#1#2{\GetFamilyName{#1}\ArgOne
        \GetFamilyName{#2}\ArgTwo
        \ifx\ArgOne\ArgTwo\SameFamilytrue\else\SameFamilyfalse\fi}
\def\GetFamilyName#1{\edef\Tempa{#1}\def\Tempb{#1}\ifx\Tempa\Tempb
        \edef\Tempa{\fontname#1}\fi
        \edef\Tempa{\Tempa\space}%
        \expandafter\iGetFamilyName\Tempa\\}
\def\iGetFamilyName#1 #2\\#3{\def#3{#1}}
\def\DefFontName#1#2{{\escapechar-1\expandafter\expandafter\expandafter
        \iDefFontName\expandafter{\csname#2\endcsname}%
        \xdef#1{\expandafter\string\Tempa}}}
\def\iDefFontName{\def\Tempa}

\newcommand\unprotectedae
{\CheckFamily\font\fraknomath\ifSameFamily *a\else
 \CheckFamily\font\swab\ifSameFamily\char'212\else\"a\fi\fi}
\newcommand\unprotectedoe
{\CheckFamily\font\fraknomath\ifSameFamily 
*o\else\CheckFamily\font\swab\ifSameFamily\char'232\else\"o\fi\fi}
\newcommand\unprotectedue
{\CheckFamily\font\fraknomath\ifSameFamily 
*u\else\CheckFamily\font\swab\ifSameFamily\char'237\else\"u\fi\fi}

\DefFontName\eccclarge{eccc1200}
\DefFontName\eccc{eccc1000}
\DefFontName\ecccsmall{eccc0900}
\DefFontName\ecccfootnotesize{eccc0800}

\newcommand\unprotectedes
{\CheckFamily\font\fraknomath\ifSameFamily\char'215\else
\CheckFamily\font\swab\ifSameFamily\char'215\else  
s\fi\fi}

\newcommand\unprotectedesi
{\CheckFamily\font\fraknomath\ifSameFamily\char'215\else
\CheckFamily\font\swab\ifSameFamily\char'215\else  
\mbox{s\hskip.05em}\fi\fi
\-\ignorespaces}

\newcommand\unprotectedmyparagraphsymbol
{\CheckFamily\font\fraknomath\ifSameFamily 
\char'244\else\CheckFamily\font\swab\ifSameFamily
\char'244\else\S\fi\fi}

\renewcommand\ae{\protect\unprotectedae}
\renewcommand\oe{\protect\unprotectedoe}
\newcommand\ue  {\protect\unprotectedue}

\newcommand\es  {\protect\unprotectedes}

\newcommand\esi {\protect\unprotectedesi}  

\newcommand\myparagraphsymbol{\protect\unprotectedmyparagraphsymbol}

\newcommand\namefont{}

\newcommand\footroom{\raisebox{-1.5ex}{\rule{0ex}{.5ex}}}

\newcommand\smallfootroom{\raisebox{-1.0ex}{\rule{0ex}{3ex}}}

\newcommand\headroom{\rule{0ex}{2.8ex}}

\newcommand\claus   {Clau\es}

\newcommand\jan     {Jan}

\newcommand\joerg   {J\oe rg}

\newcommand\paul    {Paul}
\newcommand\peter   {Peter}

\newcommand\ruediger{R\ue\-di\-ger}

\newcommand\ulrich  {Ul\-rich}

\newcommand\volker  {Vol\-ker}

\newcommand\bush            {Bush}
\newcommand\bushname        {Vannevar \bush}

\newcommand\kuehler         {K\ue h\-ler}
\newcommand\kuehlername     {\ulrich\ \kuehler}

\newcommand\lenart          {{\namefont Lenart}}
\newcommand\lenartname      {{\namefont Mihaly \lenart}}

\newcommand\lunde           {Lunde}
\newcommand\lundename       {\ruediger\ \lunde}

\newcommand\mattick         {Mattick}
\newcommand\mattickname     {\volker\ \mattick}

\newcommand\padawitz        {Pada\-witz}
\newcommand\padawitzname    {\peter\ \padawitz}

\newcommand\pasztor         {Pasztor}
\newcommand\pasztorname     {Ana \pasztor}

\newcommand\wirth           {{\namefont Wirth}}
\newcommand\wirthname       {{\namefont\claus-\peter\ \wirth}}

\newcommand\ASF  {ASF}
\newcommand\asfplus{\ASF\/\math{^+}}

\newcommand\FB   {FB}

\newcommand\FBinfshort{\FB\ Informatik}

\newcommand\ff   {\mbox{}{ff.}}  

\newcommand\defi {Definition} 

\newcommand\Nov  {Nov.}

\newcommand\getittotheright[1]  
{\hfill\mbox{}\penalty 100\mbox{\ \,}\nolinebreak
\hfill\hfill\hfill\nolinebreak\mbox{#1}\ignorespaces}

\newcommand\uni  {Uni\-ver\-si\-t\ae t}
\newcommand\Univ {Univ.}

\newcommand\WWW  {WWW}
\newcommand\www  {\url{http://www.ags.uni-sb.de/~cp}}

\newcommand\cf   {cf.}

\newcommand\Comm {Comm.}

\newcommand\Conf {Conf.}

\newcommand\CS   {Computer \Sci}

\newcommand\eg   {e.g.}
\newcommand\Eg   {E.g.}

\newcommand\etc  {\&c.}

\newcommand\ie   {i.e.}
\newcommand\Ie   {I.e.}

\newcommand\uiff {\ iff\ }
\newcommand\udiff{\ if\ }

\newcommand\ITP  {Inductive Theorem Proving}

\newcommand\Oct  {Oct.}
\newcommand\p    {p.}
\newcommand\pp   {pp.}
\newcommand\PP[2]{\pp\,\ignorespaces#1--\ignorespaces#2}

\newcommand\Proc {Proc.}

\newcommand\resp {resp.}

\newcommand\sect {\myparagraphsymbol} 

\newcommand\Sci  {Sci.}

\newcommand\wrt  {w.r.t.}

\newcommand\litsectref[1]{\sect\,#1} 

\newcommand\litdefiref[1]{\defi\,#1}
\newcommand\Examplename{Ex\-am\-ple}

\newcommand\defiref[1]{\litdefiref{\ref{#1}}}

\newcommand\sectref[1]{\litsectref{\ref{#1}}}

\newcommand\nthpositioner[2]
{#1\raisebox{0.52ex}{\scriptsize\hspace{0.07em}#2}}
\newcommand\nth[1]{\nthtinypositioner{#1}{\nthstring{#1}}}
\newcommand\nthtinypositioner[2]{#1\raisebox{0.52ex}{\tiny\hspace{0.07em}#2}}

\newcommand\mthpositioner[2]
{\math{#1}\raisebox{0.52ex}{\scriptsize\hspace{0.07em}#2}}
\newcommand\modulointocountzero[2]
{\count1=#1
 \count2=#2
 \count0=\count1
 \divide  \count0 by \count2
 \multiply\count0 by-\count2
 \advance \count0 by \count1}
\newcommand\absolutevalueintocountzero[1]
{\count0=#1
 \ifnum\count0<0\multiply\count0 by -1\fi}
\newcommand\nthstring[1]
{\def\myargone{#1}\ifcat a\myargone th\else\nthstringnochar{#1}\fi}
\newcommand\nthstringnochar[1]
{\absolutevalueintocountzero{#1}%
 \modulointocountzero{\count0}{100} 
 \ifnum\count0>9\ifnum\count0<20 th\else\stupidnthstring\fi
                                   \else\stupidnthstring\fi}
\newcommand\stupidnthstring
{\modulointocountzero{\count0}{10}
 \ifnum\count0=1 \hskip-0.2em st\else
 \ifnum\count0=2 nd\else
 \ifnum\count0=3 rd\else 
                 th\fi\fi\fi}

\newcommand\writeascents
[1]{\count4=#1
\ifnum\count4<0 
-\multiply\count4 by -1\fi
\modulointocountzero{\count4}{10}
\divide\count4 by 10
\count3=\the\count0
\modulointocountzero{\count4}{10}
\divide\count4 by 10
\the\count4
.\the\count0
\the\count3
}

\newcommand\frenchnthstring[1]
{\def\myargone{#1}\ifcat a\myargone th\else\frenchnthstringnochar{#1}\fi}
\newcommand\frenchnthstringnochar[1]
{\absolutevalueintocountzero{#1}%
 \modulointocountzero{\count0}{100} 
 \ifnum\count0>9\ifnum\count0<20 th\else\frenchstupidnthstring\fi
                                   \else\frenchstupidnthstring\fi}
\newcommand\frenchstupidnthstring
{\modulointocountzero{\count0}{10}
 \ifnum\count0=1 \hskip-0.2em re\else
 \ifnum\count0=2 me\else
 \ifnum\count0=3 rd\else 
                 th\fi\fi\fi}

\newcommand\CLAM      {{\rm CL\kern-.36em\raise.39ex\hbox{\sc a}\kern-.15emM}}

\newcommand\TEXMACS   {{\sc T\kern-.1667em\lower.5ex\hbox{E}\kern-.125emX\kern-.1em\lower.5ex\hbox{\textsc{m\kern-.05ema\kern-.125emc\kern-.05ems}}}}

\newcommand\DO             {Dort\-mund}

\newcommand\KL             {Kai\-ser\esi lau\-tern}

\newcommand\uniDO{\uni\ \DO}

\newcommand\uniDOshort{\Univ\ \DO}
\newcommand\uniKLshort{\Univ\ \KL}

\newcommand\addressuniKLshort{\FBinfshort, \uniKLshort}
\newcommand\addressuniDOshort{\FBinfshort, \uniDOshort}

\def       \email        {{\tt wirth@logic.at}}

\newcommand\LNCSvol[1]
{\mbox{LNCS}\,\ignorespaces#1}

\newcommand\academicpress{Academic Press (\elsevier)}
\newcommand\ACM{ACM}
\newcommand\acmpress{\ACM\ Press}

\newcommand\elsevier{Elsevier}

\newcommand\springerverlag{Sprin\-ger}

\newcommand\sekireportname{SEKI-Report}

\newcommand\sekireportnoaddress[2]{\sekireportname\ \mbox{SR--#1--#2}}
\newcommand\sekireport   [2]{\sekireportnoaddress{#1}{#2}, \addressuniKLshort}

\newcommand\sekiworkingpapername{SEKI-Working-Paper}

\newcommand\sekiworkingpapernoaddress[2]
{\sekiworkingpapername\ \mbox{SWP--#1--#2}}
\newcommand\sekiworkingpaper[2]{\sekiworkingpapernoaddress
{#1}{#2}, \addressuniKLshort}

\newcommand\gruenereihe[2]{Report #2/#1, \addressuniDOshort}

\newcommand\lncsconf[6]{\nth{#2}\,#1\,#3, #4, \PP{#5}{#6}, \springerverlag}

\newcommand\eighthRTAninetyseven     {\lncsconf{RTA}{ 8}{1997}{\LNCSvol{1232}}}

\newcommand\newspaperreference[5]
{\def\nameofjournalpress{#2}#1, #4 #5, #3\if?\nameofjournalpress
 \else, #2\fi}

\newcommand\dateinjournal[1]{}

\newcommand\journalreference[6]
{\def\nameofjournalpress{#2}#1\nolinebreak\hskip.2em%
 \dateinjournal{(#3) }{\mbox{\bf #4}}, \PP{#5}{#6}\if?\nameofjournalpress
 \else, #2\fi}

\newcommand\journalreferenceprintyear[6]
{\def\nameofjournalpress{#2}#1 
 {(#3) }{\bf #4}, \PP{#5}{#6}\if?\nameofjournalpress
 \else, #2\fi}

\newcommand\journalreferenceprintyearaspartofnumber[6]
{\def\nameofjournalpress{#2}#1 
 {#4/#3}, \PP{#5}{#6}\if?\nameofjournalpress
 \else, #2\fi}

\newcommand\commacm
{\journalreference{\Comm\ \ACM}\acmpress}

\newcommand\jscname
{J. Symbolic Computation}

\newcommand\jscprintyear
{\journalreferenceprintyear{\jscname}\academicpress}

\newcommand\tcsname{Theoretical \CS}
\newcommand\tcsjournal
{\journalreference\tcsname\elsevier}
\newcommand\tcsjournalprintyear
{\journalreferenceprintyear\tcsname\elsevier}

\mathcommand\ident[1]{\mathsf{#1}}
\newcommand\plussymbol  {\ident{+}}
\newcommand\minussymbol {\ident{-}}
\newcommand\dividesymbol{\ident{/}}
\newcommand\timessymbol {\ident{*}}

\newcommand\nat     {\ident{nat}}

\newcommand\set     {\ident{set}}

\newcommand\naturalssymbol{\ident{naturals}}
\newcommand\gensymsymbol{\ident{gensym}}
\mathcommand\mbpsymbol{\ident{m\hspace{-0.055em}b\hspace{-0.045em}p}}

\newcommand\csymbol     {\ident c}
\newcommand\esymbol     {\ident e}
\newcommand\fsymbol     {\ident f}
\newcommand\gsymbol     {\ident g}
\newcommand\hsymbol     {\ident h}
\newcommand\ksymbol     {\ident k}
\newcommand\psymbol     {\ident p}
\newcommand\ssymbol     {\ident s}
\newcommand\Everysymbol {\ident{Every}}
\newcommand\Permsymbol {\ident{Perm}}
\newcommand\RExistssymbol{\ident{Rexists}}
\newcommand\invertsymbol{\ident{invert}}
\newcommand\invsymbol{\ident{inv}}
\newcommand\abssymbol   {\ident{abs}}
\newcommand\cnssymbol   {\ident{cons}}
\mathcommand\cnsindexsymbol[1]{\ident{cons}_{#1}}

\newcommand\lengthsymbol{\ident{length}}
\newcommand\dlsymbol    {\ident{dl}}
\newcommand\dloncesymbol{\ident{delonce}}
\newcommand\rcsymbol    {\ident{rc}}
\newcommand\brsymbol    {\ident{br}}
\newcommand\revtailsymbol{\ident{revtail}}
\newcommand\revsymbol{\ident{rev}}
\newcommand\appendsymbol {\ident{append}}
\newcommand\zeropredicatesymbol{\ident{zerop}}
\newcommand\eqsymbol        {\ident{eq}}
\newcommand\ifthensymbol    {\mbox{\ident{If{}Then}}}
\newcommand\ifthenelsesymbol{\mbox{\ident{If{}ThenElse}}}
\mathcommand\eqindexsymbol        [1]{\eqsymbol        _{#1}}
\mathcommand\ifthenindexsymbol    [1]{\ifthensymbol    _{#1}}
\mathcommand\ifthenelseindexsymbol[1]{\ifthenelsesymbol_{#1}}
\newcommand\orsymbol    {\ident{or}}
\newcommand\andsymbol   {\ident{and}}
\newcommand\leqsymbol   {\ident{leq}}
\newcommand\lessymbol   {\ident{less}}
\newcommand\lexsymbol   {\ident{lex}}
\newcommand\acksymbol   {\ident{ack}}
\newcommand\switchsymbol{\ident{switch}}
\newcommand\swatchsymbol{\ident{swatch}}
\newcommand\diveinssymbol{\ident{div1}}
\newcommand\divzweisymbol{\ident{div2}}
\newcommand\divrestsymbol{\ident{divrest}}
\newcommand\diveinstailsymbol{\ident{div1tail}}
\newcommand\divzweitailsymbol{\ident{div2tail}}

\newcommand\turingmachinesymbol{\ident T}
\newcommand\terminatespsymbol  {\ident{terminatesp}}
\newcommand\statesymbol        {\ident{state}}
\newcommand\cmdsymbol          {\ident{cmd}}
\newcommand\nthsymbol          {\ident{nth}}
\newcommand\doublesymbol       {\ident{double}}

\newcommand\ppsymbol           {\ident{p}}
\newcommand\qpsymbol           {\ident{q}}
\newcommand\Epsymbol           {\ident{E}}
\newcommand\Ppsymbol           {\ident{P}}
\newcommand\Qpsymbol           {\ident{Q}}
\newcommand\Marriessymbol      {\ident{Marries}}
\newcommand\Lovessymbol        {\ident{Loves}}
\newcommand\StolenBysymbol     {\ident{StolenBy}}
\newcommand\Humansymbol        {\ident{Human}}
\newcommand\Evensymbol         {\ident{Even}}
\newcommand\Oddsymbol          {\ident{Odd}}
\newcommand\Primesymbol        {\ident{Prime}}
\newcommand\EveryPairsymbol   {\ident{EveryPair}}
\newcommand\Givesymbol         {\ident{Give}}
\newcommand\Fathersymbol       {\ident{Father}}
\newcommand\Elephantpsymbol    {\ident{Elephant}}
\newcommand\Flowerpsymbol    {\ident{Flower}}
\newcommand\Germanpsymbol      {\ident{German}}
\newcommand\Bicyclepsymbol     {\ident{Bicycle}}
\newcommand\Hugepsymbol        {\ident{Huge}}
\newcommand\Animalpsymbol      {\ident{Animal}}
\newcommand\Malepsymbol        {\ident{Male}}
\newcommand\Boypsymbol        {\ident{Boy}}
\newcommand\Girlpsymbol        {\ident{Girl}}
\newcommand\Femalepsymbol      {\ident{Female}}
\newcommand\Roundpsymbol       {\ident{Round}}
\newcommand\Quadrangularpsymbol{\ident{Quadrangular}}
\newcommand\Metpsymbol         {\ident{Met}}
\newcommand\Bishopsymbol       {\ident{Bishop}}
\newcommand\mindexsymbol[1]{\existsvari w{#1}}

\newcommand\nonnegpsymbol      {\ident{nonnegp}}
\newcommand\wellsymbol         {\ident{well}}
\newcommand\welltailsymbol     {\ident{welltail}}
\newcommand\varsymbol          {\ident{var}}
\newcommand\aritysymbol        {\ident{arity}}

\newcommand\whilesymbol        {\ident{while}}

\newcommand\nullsymbol         {\ident{null}}
\newcommand\hdsymbol           {\ident{hd}}
\newcommand\tlsymbol           {\ident{tl}}
\newcommand\insymbol           {\ident{in}}
\newcommand\applysymbol        {\ident{app}}
\newcommand\termsymbol         {\ident{term}}
\mathcommand\tightim{\longrightarrow}
\mathcommand\im{\ \tightim\ }
\mathcommand\rs{\:\rulesugar\:\:}
\mathcommand\rulesugar{\longleftarrow}

\mathcommand\doublepp[1]      {\doublesymbol   \beginargs{#1}\allargs}
\mathcommand\aritypp[1]      {\aritysymbol   \beginargs{#1}\allargs}
\mathcommand\lengthpp[1]      {\lengthsymbol   \beginargs{#1}\allargs}
\mathcommand\wellpp[1]      {\wellsymbol   \beginargs{#1}\allargs}
\mathcommand\welltailpp[1]      {\welltailsymbol   \beginargs{#1}\allargs}
\mathcommand\varpp[1]      {\varsymbol   \beginargs{#1}\allargs}
\mathcommand\divrestpp[2]    {\divrestsymbol\beginargs{#1}\separgs{#2}\allargs}
\mathcommand\diveinspp[2]    {\diveinssymbol\beginargs{#1}\separgs{#2}\allargs}
\mathcommand\divzweipp[3]    {\divzweisymbol\beginargs{#1}\separgs{#2}
\separgs{#3}\allargs}
\mathcommand\diveinstailpp[4]    {\diveinstailsymbol\beginargs{#1}\separgs{#2}
\separgs{#3}\separgs{#4}\allargs}
\mathcommand\divzweitailpp[6]    {\divzweitailsymbol\beginargs{#1}\separgs{#2}
\separgs{#3}\separgs{#4}\separgs{#5}\separgs{#6}\allargs}
\mathcommand\mbppp[2]         {\mbpsymbol   \beginargs{#1}\separgs{#2}\allargs}
\mathcommand\revpp[1]     
{\revsymbol\beginargs{#1}\allargs}
\mathcommand\revppi[2]     
{\revsymbol^{#1}\beginargs{#2}\allargs}
\mathcommand\revtailpp[2]     
{\revtailsymbol\beginargs{#1}\separgs{#2}\allargs}
\mathcommand\revtailppi[3]
{\revtailsymbol^{#1}\beginargs{#2}\separgs{#3}\allargs}
\mathcommand\Permpp[2]     
{\Permsymbol\beginargs{#1}\separgs{#2}\allargs}
\mathcommand\Permppi[3]
{\Permsymbol^{#1}\beginargs{#2}\separgs{#3}\allargs}
\mathcommand\appendpp[2]      
{\appendsymbol \beginargs{#1}\separgs{#2}\allargs}
\mathcommand\appendppi[3]      
{\appendsymbol^{#1}\beginargs{#2}\separgs{#3}\allargs}
\mathcommand\Everypp[2]      
{\Everysymbol \beginargs{#1}\separgs{#2}\allargs}
\mathcommand\RExistspp[1]      
{\RExistssymbol \beginargs{#1}\allargs}
\mathcommand\appendlongpp[2]      
{\appendsymbol\left(\begin{array}{@{}l@{}}{#1}\separgs\\{#2}\end{array}\right)}
\mathcommand\cnspp[2]         {\cnssymbol   \beginargs{#1}\separgs{#2}\allargs}
\mathcommand\cnsppi[3]       {\cnssymbol^{#1}\beginargs{#2}\separgs{#3}\allargs}
\mathcommand\cnsindexpp[3]
{\cnsindexsymbol{#1}\beginargs{#2}\separgs{#3}\allargs}
\mathcommand\dlpp[2]          {\dlsymbol    \beginargs{#1}\separgs{#2}\allargs}
\mathcommand\dloncepp[2]      {\dloncesymbol\beginargs{#1}\separgs{#2}\allargs}
\mathcommand\dlonceppi[3]{\dloncesymbol^{#1}\beginargs{#2}\separgs{#3}\allargs}
\mathcommand\rcpp[2]          {\rcsymbol    \beginargs{#1}\separgs{#2}\allargs}
\mathcommand\brpp[2]          {\brsymbol    \beginargs{#1}\separgs{#2}\allargs}
\mathcommand\orpp[2]          {\orsymbol    \beginargs{#1}\separgs{#2}\allargs}
\mathcommand\andpp[2]         {\andsymbol   \beginargs{#1}\separgs{#2}\allargs}
\mathcommand\shortcnspp[2]    {\csymbol     \beginargs{#1}\separgs{#2}\allargs}
\mathcommand\tightshortcnspp[2]
{\csymbol\beginargs{#1}\tightsepargs{#2}\allargs}
\mathcommand\spp[1]           {\ssymbol     \beginargs{#1}\allargs}
\mathcommand\sppiterated[2]   {\ssymbol^{#1}\beginargs{#2}\allargs}
\mathcommand\ppp[1]           {\psymbol     \beginargs{#1}\allargs}
\mathcommand\pppiterated[2]   {\psymbol^{#1}\beginargs{#2}\allargs}
\mathcommand\zeropp           {\ident 0}
\mathcommand\Julietpp         {\ident{Juliet}}
\mathcommand\Romeopp          {\ident{Romeo}}
\mathcommand\Ipp              {\ident I}
\mathcommand\onepp            {\ident1}
\mathcommand\twopp            {\ident2}
\mathcommand\threepp          {\ident3}
\mathcommand\invertpp[1]      {\invertsymbol\beginargs{#1}\allargs}
\mathcommand\invpp[1]         {\invsymbol\beginargs{#1}\allargs}
\mathcommand\abspp[1]         {\abssymbol\beginargs{#1}\allargs}
\mathcommand\naturalspp[1]    {\naturalssymbol\beginargs{#1}\allargs}
\mathcommand\gensympp[1]      {\gensymsymbol\beginargs{#1}\allargs}
\mathcommand\nilpp            {\ident{nil}}
\mathcommand\falsepp          {\ident{false}}
\mathcommand\truepp           {\ident{true}}
\mathcommand\FALSEpp          {\ident{FALSE}}
\mathcommand\TRUEpp           {\ident{TRUE}}
\mathcommand\weirdppp         {\ident{weirdp}}
\mathcommand\ambigppp         {\ident{ambigp}}
\mathcommand\zeropredicatepp[1]{\zeropredicatesymbol\beginargs{#1}\allargs}
\mathcommand\cppeins       [1]{\csymbol     \beginargs{#1}\allargs}
\mathcommand\cppzwei       [2]{\csymbol\beginargs{#1}\separgs{#2}\allargs}
\mathcommand\eppeins       [1]{\esymbol     \beginargs{#1}\allargs}
\mathcommand\fppeins       [1]{\fsymbol     \beginargs{#1}\allargs}
\mathcommand\fppeinsindex  [2]{\fsymbol_{#1}\beginargs{#2}\allargs}
\mathcommand\fppeinsiterated[2]{\fsymbol^{#1}\beginargs{#2}\allargs}
\mathcommand\gppeins       [1]{\gsymbol     \beginargs{#1}\allargs}
\mathcommand\gppzwei       [2]{\gsymbol     \beginargs{#1}\separgs{#2}\allargs}
\mathcommand\hppeins       [1]{\hsymbol     \beginargs{#1}\allargs}
\mathcommand\kppeins       [1]{\ksymbol     \beginargs{#1}\allargs}
\mathcommand\appzero          {\ident a}
\mathcommand\bppzero          {\ident b}
\mathcommand\cppzero          {\ident c}
\mathcommand\dppzero          {\ident d}
\mathcommand\eppzero          {\ident e}
\mathcommand\eqindexpp[3]{\eqindexsymbol{#1}\beginargs{#2}\separgs{#3}\allargs}
\mathcommand\ifthenindexpp
[3]{\ifthenindexsymbol{#1}\beginargs{#2}\separgs{#3}\allargs}
\mathcommand\ifthenelseindexpp
[4]{\ifthenelseindexsymbol{#1}\beginargs{#2}\separgs{#3}\separgs{#4}\allargs}
\mathcommand\eqpp[2]{\eqsymbol\beginargs{#1}\separgs{#2}\allargs}
\mathcommand\leqpp[2]{\leqsymbol\beginargs{#1}\separgs{#2}\allargs}
\mathcommand\lespp[2]{\lessymbol\beginargs{#1}\separgs{#2}\allargs}
\mathcommand\lexpp[3]{\lexsymbol\beginargs{#1}\separgs{#2}\separgs{#3}\allargs}
\mathcommand\ackpp[2]{\acksymbol\beginargs{#1}\separgs{#2}\allargs}
\mathcommand\switchpp[1]{\switchsymbol\beginargs{#1}\allargs}
\mathcommand\swatchpp[1]{\swatchsymbol\beginargs{#1}\allargs}
\mathcommand\whilepp[2]{\whilesymbol\beginargs{#1}\separgs{#2}\allargs}
\mathcommand\nullpp[1]{\nullsymbol\beginargs{#1}\allargs}
\mathcommand\nullppiterated[2]{\nullsymbol^{#1}\beginargs{#2}\allargs}
\mathcommand\hdpp[1]{\hdsymbol\beginargs{#1}\allargs}
\mathcommand\hdppiterated[2]{\hdsymbol^{#1}\beginargs{#2}\allargs}
\mathcommand\tlpp[1]{\tlsymbol\beginargs{#1}\allargs}
\mathcommand\tlppiterated[2]{\tlsymbol^{#1}\beginargs{#2}\allargs}
\mathcommand\inpp[2]{\insymbol\beginargs{#1}\separgs{#2}\allargs}
\mathcommand\inppiterated[3]{\insymbol^{#1}\beginargs{#2}\separgs{#3}\allargs}
\mathcommand\applypp[2]{\applysymbol\beginargs{#1}\separgs{#2}\allargs}
\mathcommand\applyppiterated
[3]{\applysymbol^{#1}\beginargs{#2}\separgs{#3}\allargs}
\mathcommand\termpp[2]{\termsymbol\beginargs{#1}\separgs{#2}\allargs}
\mathcommand\setpp[1]{\set\beginargs{#1}\allargs}

\mathcommand\Tpp[6]{\turingmachinesymbol\beginargs{#1}\separgs{#2}\separgs
{#3}\separgs{#4}\separgs{#5}\separgs{#6}\allargs}
\mathcommand\Tppseven[7]{\turingmachinesymbol\beginargs{#1}\separgs{#2}\separgs
{#3}\separgs{#4}\separgs{#5}\separgs{#6}\separgs{#7}\allargs}
\mathcommand\foreverppp[6]{\ident{foreverp}\beginargs{#1}\separgs{#2}\separgs
{#3}\separgs{#4}\separgs{#5}\separgs{#6}\allargs}
\mathcommand\terminatesppp[6]{\terminatespsymbol\beginargs{#1}\separgs
{#2}\separgs{#3}\separgs{#4}\separgs{#5}\separgs{#6}\allargs}
\mathcommand\terminatespppone[1]{\terminatespsymbol \beginargs{#1}\allargs}
\mathcommand\statepp
[3]{\statesymbol\beginargs{#1}\separgs{#2}\separgs{#3}\allargs}
\mathcommand\tightstatepp
[3]{\statesymbol\beginargs{#1}\tightsepargs{#2}\tightsepargs{#3}\allargs}
\mathcommand\cmdpp
[3]{\cmdsymbol  \beginargs{#1}\separgs{#2}\separgs{#3}\allargs}
\mathcommand\tightcmdpp
[3]{\cmdsymbol  \beginargs{#1}\tightsepargs{#2}\tightsepargs{#3}\allargs}
\mathcommand\stoppp           {\ident{stop}}
\mathcommand\leftpp           {\ident{left}}
\mathcommand\rightpp          {\ident{right}}
\mathcommand\nthpp         [2]{\nthsymbol  \beginargs{#1}\separgs{#2}\allargs}
\mathcommand\pppp          [1]{\ppsymbol\beginargs{#1}            \allargs}
\mathcommand\qppp          [2]{\qpsymbol\beginargs{#1}\separgs{#2}\allargs}
\mathcommand\Eppp          [1]{\Epsymbol\beginargs{#1}            \allargs}
\mathcommand\Epppzwei      [2]{\Epsymbol\beginargs{#1}\separgs{#2}\allargs}
\mathcommand\Pppp          [1]{\Ppsymbol\beginargs{#1}            \allargs}
\mathcommand\Ppppdrei      
[3]{\Ppsymbol\beginargs{#1}\separgs{#2}\separgs{#3}\allargs}
\mathcommand\Ppppvier
[4]{\Ppsymbol\beginargs{#1}\separgs{#2}\separgs{#3}\separgs{#4}\allargs}
\mathcommand\Qppp          [2]{\Qpsymbol\beginargs{#1}\separgs{#2}\allargs}
\mathcommand\Qpppeins      [1]{\Qpsymbol\beginargs{#1}\allargs}
\mathcommand\Qpppdrei      
[3]{\Qpsymbol\beginargs{#1}\separgs{#2}\separgs{#3}\allargs}
\mathcommand\Fatherpp      [2]{\Fathersymbol\beginargs{#1}\separgs{#2}\allargs}
\mathcommand\Marriespp     [2]{\Marriessymbol\beginargs{#1}\separgs{#2}\allargs}
\mathcommand\Lovespp       [2]{\Lovessymbol\beginargs{#1}\separgs{#2}\allargs}
\mathcommand\StolenBypp    [2]
{\StolenBysymbol\beginargs{#1}\separgs{#2}\allargs}
\mathcommand\Humanpp       [1]{\Humansymbol\beginargs{#1}\allargs}
\mathcommand\Evenpp        [1]{\Evensymbol\beginargs{#1}\allargs}
\mathcommand\Evenppi       [2]{\Evensymbol^{#1}\beginargs{#2}\allargs}
\mathcommand\Oddpp         [1]{\Oddsymbol\beginargs{#1}\allargs}
\mathcommand\Primepp       [1]{\Primesymbol\beginargs{#1}\allargs}
\mathcommand\EveryPairpp  [2]{\EveryPairsymbol\beginargs{#1}\separgs
{#2}\allargs}
\mathcommand\mindexppeins  [2]{\mindexsymbol{#1}\beginargs{#2}\allargs}
\mathcommand\Givepp        [3]{\Givesymbol
\beginargs{#1}\separgs{#2}\separgs{#3}\allargs}
\mathcommand\mindexppzwei  [3]{\mindexsymbol
{#1}\beginargs{#2}\separgs{#3}\allargs}
\mathcommand\mindexppdrei  [4]{\mindexsymbol
{#1}\beginargs{#2}\separgs{#3}\separgs{#4}\allargs}

\mathcommand\nonnegppp     [1]{\nonnegpsymbol\beginargs{#1}\allargs}

\mathcommand\anonymouscsymbol{c}
\mathcommand\anonymouscindexsymbol[1]{\anonymouscsymbol_{#1}}
\mathcommand\anonymousfsymbol{f}
\mathcommand\anonymouscpp
[2]{\anonymouscsymbol\beginargs{#1}\separgs\ldots\separgs{#2}\allargs}
\mathcommand\anonymouscindexpp
[3]{\anonymouscindexsymbol{#1}\beginargs{#2}\separgs\ldots\separgs{#3}\allargs}
\mathcommand\anonymousfpp
[2]{\anonymousfsymbol\beginargs{#1}\separgs\ldots\separgs{#2}\allargs}
\mathcommand\coerceindexpp[3]{[#3]_{#1}^{#2}}

\mathcommand\Elephantppp    [1]{\Elephantpsymbol\beginargs{#1}\allargs}
\mathcommand\Flowerppp      [1]{\Flowerpsymbol  \beginargs{#1}\allargs}
\mathcommand\Bicycleppp     [1]{\Bicyclepsymbol \beginargs{#1}\allargs}
\mathcommand\Germanppp      [1]{\Germanpsymbol  \beginargs{#1}\allargs}
\mathcommand\Hugeppp        [1]{\Hugepsymbol    \beginargs{#1}\allargs}
\mathcommand\Animalppp      [1]{\Animalpsymbol  \beginargs{#1}\allargs}
\mathcommand\Maleppp        [1]{\Malepsymbol    \beginargs{#1}\allargs}
\mathcommand\Boyppp         [1]{\Boypsymbol     \beginargs{#1}\allargs}
\mathcommand\Girlppp        [1]{\Girlpsymbol    \beginargs{#1}\allargs}
\mathcommand\Femaleppp      [1]{\Femalepsymbol  \beginargs{#1}\allargs}
\mathcommand\Roundppp       [1]{\Roundpsymbol   \beginargs{#1}\allargs}
\mathcommand\Bishoppp       [1]{\Bishopsymbol   \beginargs{#1}\allargs}
\mathcommand\Quadrangularppp[1]{\Quadrangularpsymbol  \beginargs{#1}\allargs}
\mathcommand\Metppp[2]{\Metpsymbol     \beginargs{#1}\separgs{#2}\allargs}

\newcommand\bound     {{\rm bound}}
\newcommand\free      {{\rm free}}

\mathcommand\Vtripleindex[3]{\V\!_{{#1},\,{#2},\,{#3}}}
\mathcommand\Vdoubleindex[2]{\V\!_{{#1},\,{#2}}}
\mathcommand\Vsingleindex[1]{\V\!_{{#1}}}

\mathcommand\Erel[1]{\Gammaoffont\!_{#1}}
\mathcommand\Urel[1]{\Deltaoffont_{#1}}

\mathcommand\theRprimefromstrongtoweak{
  \inparenthesesinlinetight{
     \domres\id{\Vwall\cup\Vsome\setminus\RAN\varsigma}
     \nottight{\nottight\uplus}
     \reverserelation\varsigma
  }
  \nottight{\circ}
  \ranres
    {\transclosureinline R}
    {\Vwall\cup\Vsome\setminus\RAN\varsigma}
  \nottight{\nottight{\nottight{\uplus}}}
  \Vsome\tighttimes\Vsall
}

\mathcommand\deltaminus{\delta^-}
\mathcommand\deltaplus{\delta^+}
\mathcommand\deltaplusplus{\delta^{+^+}}
\mathcommand\deltastar{\delta^*}
\mathcommand\deltastarstar{\delta^{*^*}}

\mathcommand\Vall     {\Vsingleindex\indexdelta         }
\mathcommand\Vwall    {\Vsingleindex\indexdeltaminu     }
\mathcommand\Vsall    {\Vsingleindex\indexdeltaplus     }
\mathcommand\Vgsome   {\Vsingleindex\indexgammaplus     }
\mathcommand\Vsome    {\Vsingleindex\indexgamma         }
\mathcommand\Vfree    {\Vsingleindex\indexfree          }
\mathcommand\Vbound   {\Vsingleindex\indexbound         }
\mathcommand\Vsomesall{\Vsingleindex\indexgammadeltaplus}

\mathapplycommand\VARall      {\VARsingleindex\indexdelta         }
\mathapplycommand\VARwall     {\VARsingleindex\indexdeltaminu     }
\mathapplycommand\VARsall     {\VARsingleindex\indexdeltaplus     }
\mathapplycommand\VARgsome    {\VARsingleindex\indexgammaplus     }
\mathapplycommand\VARsome     {\VARsingleindex\indexgamma         }
\mathapplycommand\VARfree     {\VARsingleindex\indexfree          }
\mathapplycommand\VARbound    {\VARsingleindex\indexbound         }
\mathapplycommand\VARsomesall {\VARsingleindex\indexgammadeltaplus}
\mathcommand\displayVARsall[1]{\VARsingleindex\indexdeltaplus
\!\!\!\:\left(\begin{array}{@{}c@{}}#1\end{array}\right)}

\mathcommand\rigidvari     [2]{#1_{#2}^\indexgammadeltaplus}
\mathcommand\existsvari    [2]{#1_{#2}^\indexgamma    }
\mathcommand\forallvari    [2]{#1_{#2}^\indexdelta    }
\mathcommand\freevari      [2]{#1_{#2}^\indexfree     }
\mathcommand\wforallvari   [2]{#1_{#2}^\indexdeltaminu}
\mathcommand\sforallvari   [2]{#1_{#2}^\indexdeltaplus}
\mathcommand\gexistsvari   [2]{#1_{#2}^\indexgammaplus}
\mathcommand\boundvari     [2]{#1_{#2}}
\mathcommand\vari          [2]{#1_{#2}}
\mathcommand\wforallvarilow[2]{#1_{#2}^
{\raisebox{-.82ex}{\math\indexdeltaminu}}}

\newcommand\indexhelper[1]{{\scriptscriptstyle#1\:\!\!}}
\newcommand\indexdeltaplus
{\indexhelper{\delta^{\raisebox{-.17ex}{\fvesf\hskip-0.14em +}}}}
\newcommand\indexdeltaminu
{\indexhelper{\delta^{\mbox{\fvesf\hskip-0.14em\rule[.2ex]{.7em}{.15ex}}}}}
\newcommand\indexgammaplus
{\indexhelper{\gamma^{\mbox{\fvesf\hskip-0.14em +}}}}
\newcommand\indexgammadeltaplus
{\indexhelper{\gamma\delta^{\raisebox{-.17ex}{\fvesf\hskip-0.14em +}}}}

\newcommand\indexdelta{\indexhelper\delta}
\newcommand\indexgamma{\indexhelper\gamma}
\newcommand\indexfree
{{\scriptscriptstyle\free}}
\newcommand\indexbound
{{\scriptscriptstyle\bound}}

\newcommand\Wellfsymb{\ident{Wellf}}
\mathapplycommand\Wellfpp{\Wellfsymb}

\mathcommand\asfrs{\mbox{\asffont{if~}}}
\newcommand\asffont[1]{{\tt #1}}
\renewcommand\ident[1]{{\iden #1}}
\newcommand\iden{\rm}

\mathcommand\parblock
[1]{\asffont(\begin{array}[t]{@{}l@{}}#1\asffont)\end{array}}

\newcommand\import
[2]{\\\headroom\ident{#1}\\\asffont\{\\\begin{tabular}{l@{}}#2\end
{tabular}\\\asffont\}}

\mathcommand\declarevars[1]
{{\begin{array}[t]{l@{\ :\ }l}#1\end{array}}}

\mathcommand\declarefunctions
[1]{{\begin{array}[t]{@{}l@{\ :\ }l@{}l@{}l}#1\end{array}}}
\mathcommand\sepsort{\ }
\mathcommand\as{\;\ \mbox{\asffont{->}}\;\ }
\newcommand\placeholder{\asffont\_}

\mathcommand\definefunction
[1]{{\begin{array}[t]{@{}l l@{\ =\ }l l}#1\end{array}}}
\mathcommand\definelongfunction
[1]{\begin{array}[t]{@{}l l l l l}#1\end{array}}

\mathcommand\definegoal
[1]{{\begin{array}[t]{@{}l l}#1\end{array}}}

\mathcommand\qualifyname[2]{#1\mbox{\asffont<}#2\mbox{\asffont>}}

\newcommand\applyparsymb[2]{\ident{APPLY#1to#2}}
\mathcommand\applyparpp
[4]{\applyparsymb{#1}{#2}\beginargs#3\separgs#4\allargs}
\newcommand\applypardreisymb[3]{\mbox{\ident{APPLY[#1X#2]to#3}}}
\mathcommand\applypardreipp
[6]{\applypardreisymb{#1}{#2}{#3}\beginargs#4\separgs#5\separgs#6\allargs}

\newcommand\mktablelinesymb{\ident{mktableline}}
\newcommand\mktablelinepp[1]{\mktablelinesymb\beginargs#1\allargs}

\newcommand\appendsymb{\ident{append}}
\renewcommand\appendpp[2]{\appendsymb\beginargs#1\separgs#2\allargs}

\newcommand\insertsymb{\ident{insert}}
\mathcommand\insertpp[2]{\insertsymb\beginargs#1\separgs#2\allargs}
\newcommand\insertatsymb{\ident{insertat}}
\mathcommand\insertatpp
[3]{\insertatsymb\beginargs#1\separgs#2\separgs#3\allargs}
\newcommand\iteratesymb{\ident{iterate}}
\mathcommand\iteratepp
[3]{\iteratesymb\beginargs#1\separgs#2\separgs#3\allargs}
\newcommand\mapcarsymb{\ident{mapcar}}
\mathcommand\mapcarpp
[3]{\mapcarsymb\beginargs#1\separgs#2\separgs#3\allargs}
\newcommand\insertatextendsymb{\ident{insertatextend}}
\mathcommand\insertatextendpp
[3]{\insertatextendsymb\beginargs#1\separgs#2\separgs#3\allargs}

\mathcommand\insertatextendvierpp
[4]{\insertatextendsymb\beginargs#1\separgs#2\separgs#3\separgs#4\allargs}
\newcommand\hdinsertatsymb{\ident{insertat}}
\mathcommand\hdinsertatpp
[4]{\hdinsertatsymb\beginargs#1\separgs#2\separgs#3\separgs#4\allargs}
\newcommand\hdinsertatliftersymb{\ident{insertatlifter}}
\mathcommand\hdinsertatlifterpp
[5]{\hdinsertatliftersymb\beginargs#1\separgs#2\separgs#3\separgs#4\separgs
#5\allargs}
\newcommand\hdinsertatextendsymb{\ident{insertatextend}}
\mathcommand\hdinsertatextendpp
[4]{\hdinsertatextendsymb\beginargs#1\separgs#2\separgs#3\separgs#4\allargs}
\newcommand\EMBEDsymb{\ident{EMBED}}
\mathcommand\EMBEDpp[1]{\EMBEDsymb\beginargs#1\allargs}
\newcommand\tlsymb{\ident{tl}}
\remathcommand\tlpp[1]{\tlsymb\beginargs#1\allargs}
\newcommand\hdsymb{\ident{hd}}
\remathcommand\hdpp[1]{\hdsymb\beginargs#1\allargs}
\newcommand\domsymb{\ident{dom}}
\mathcommand\dompp[1]{\domsymb\beginargs#1\allargs}
\newcommand\ransymb{\ident{ran}}
\mathcommand\ranpp[1]{\ransymb\beginargs#1\allargs}
\newcommand\isfunctionsymb{\ident{isfunction}}
\mathcommand\isfunctionpp[1]{\isfunctionsymb\beginargs#1\allargs}

\newcommand\domdlsymb{\ident{domdl}}
\mathcommand\domdlpp[2]{\domdlsymb\beginargs#1\separgs#2\allargs}
\newcommand\dlsymb{\ident{dl}}
\remathcommand\dlpp[2]{\dlsymb\beginargs#1\separgs#2\allargs}
\mathcommand\mbpsymb{\in}
\remathcommand\mbppp[2]{#1\:\mbpsymb\:#2}
\mathcommand\mbpppinpar[2]{\inparenthesesinlinetight{#1\:\mbpsymb\:#2}}
\mathcommand\concsymb{\ident{::}}
\mathcommand\concpp[2]{#1\:\concsymb\:#2}
\mathcommand\concppinpar[2]{\inparenthesesinlinetight{#1\:\ident{::}\:#2}}
\newcommand\rcsymb{\ident{rc}}
\remathcommand\rcpp[2]{\rcsymb\beginargs{#1}\separgs{#2}\allargs}
\newcommand\brsymb{\ident{br}}
\remathcommand\brpp[2]{\brsymb\beginargs{#1}\separgs{#2}\allargs}
\remathcommand\andpp[2]{\andsymb\beginargs{#1}\separgs{#2}\allargs}
\mathcommand\andpplong
[2]{\begin{array}[b]{@{}l@{}}\andsymb\beginargs\\\ \ #1\separgs\\\ \ 
#2\allargs\end{array}}
\newcommand\andsymb{\ident{and}}
\mathcommand\setminussymb{\setminus}
\mathcommand\setminuspp[2]{#1\:\setminussymb\:#2}
\mathcommand\unionsymb{\cup}
\mathcommand\unionpp[2]{#1\:\unionsymb\:#2}
\mathcommand\unionpplong
[2]{\begin{array}[b]{@{}l@{}}\ \ #1\\\unionsymb\\\ \ #2\end{array}}
\mathcommand\intersectionsymb{\cap}
\mathcommand\intersectionpp[2]{#1\:\intersectionsymb\:#2}
\mathcommand\intersectionppinpar
[2]{\inparenthesesinlinetight{#1\:\intersectionsymb\:#2}}
\mathcommand\subseteqsymb{\subseteq}
\mathcommand\subseteqpp[2]{#1\:\subseteqsymb\:#2}
\mathcommand\subseteqppinpar
[2]{\inparenthesesinlinetight{#1\:\subseteqsymb\:#2}}
\newcommand\isprefixsymb{\ident{isprefix}}
\mathcommand\isprefixpp[1]{\isprefixsymb\beginargs#1\allargs}
\mathcommand\isprefixtwopp[2]{\isprefixsymb\beginargs#1\separgs#2\allargs}
\newcommand\isproperprefixsymb{\ident{isproperprefix}}
\mathcommand\isproperprefixpp[1]{\isproperprefixsymb\beginargs#1\allargs}
\mathcommand\isproperprefixtwopp
[2]{\isproperprefixsymb\beginargs#1\separgs#2\allargs}
\newcommand\choosesymb{\ident{choose}}
\mathcommand\choosepp[1]{\choosesymb\beginargs{#1}\allargs}
\newcommand\chooserestsymb{\ident{chooserest}}
\mathcommand\chooserestpp[1]{\chooserestsymb\beginargs{#1}\allargs}
\newcommand\equalsymb{\ident{equal}}
\mathcommand\equalpp[2]{\equalsymb\beginargs{#1}\separgs{#2}\allargs}
\newcommand\applysymb{\ident{apply}}
\remathcommand\applypp[2]{\applysymb\beginargs#1\separgs#2\allargs}
\mathcommand\applyppdrei
[3]{\applysymb\beginargs#1\separgs#2\separgs#3\allargs}
\newcommand\applyopsymb{\ident{apply}}
\mathcommand\applyoppp
[3]{\applyopsymb\beginargs#1\separgs#2\separgs#3\allargs}
\newcommand\APPLYsymb{\ident{APPLY}}
\mathcommand\APPLYpp[2]{\APPLYsymb\beginargs#1\separgs#2\allargs}
\newcommand\APPLYOPsymb{\ident{APPLY}}
\mathcommand\APPLYOPpp
[3]{\APPLYOPsymb\beginargs#1\separgs#2\separgs#3\allargs}
\newcommand\APPLYeinssymb{\ident{APPLY1}}
\mathcommand\APPLYeinspp[2]{\APPLYeinssymb\beginargs#1\separgs#2\allargs}
\newcommand\forallsymb{\ident{forall}}
\mathcommand\forallpp[2]{\forallsymb\beginargs#1\separgs#2\allargs}
\newcommand\existssymb{\ident{exists}}
\mathcommand\existspp[2]{\existssymb\beginargs#1\separgs#2\allargs}
\newcommand\filtersymb{\ident{filter}}
\mathcommand\filterpp[2]{\filtersymb\beginargs#1\separgs#2\allargs}
\newcommand\dualfiltersymb{\ident{dualfilter}}
\mathcommand\dualfilterpp[2]{\dualfiltersymb\beginargs#1\separgs#2\allargs}
\newcommand\replaceaddrlinksymb{\ident{replaceaddrlink}}
\mathcommand\replaceaddrlinkpp
[2]{\replaceaddrlinksymb\beginargs#1\separgs#2\allargs}
\newcommand\replaceaddrspecsymb{\ident{replaceaddrspec}}
\mathcommand\replaceaddrspecpp
[2]{\replaceaddrspecsymb\beginargs#1\separgs#2\allargs}
\newcommand\replacesymb{\ident{mapch}}
\mathcommand\replacepp[2]{\replacesymb\beginargs#1\separgs#2\allargs}
\mathcommand\replacepplong[2]{\begin{array}[b]{@{}l@{}}\replacesymb\beginargs\\\ \ #1\separgs\\\ \ #2\allargs\end{array}}
\newcommand\mkpairsymb{\ident{mkpair}}
\mathcommand\mkpairpp[2]{\mkpairsymb\beginargs#1\separgs#2\allargs}
\newcommand\mklinkctsymb{\ident{mklinkct}}
\mathcommand\mklinkctpp[4]{\mklinkctsymb\beginargs#1\separgs#2\separgs#3\separgs#4\allargs}
\newcommand\mksitesymb{\ident{mksite}}
\mathcommand\mksitepp[2]{\mksitesymb\beginargs#1\separgs#2\allargs}
\newcommand\lengthsymb{\ident{length}}
\remathcommand\lengthpp[1]{\lengthsymb\beginargs#1\allargs}
\newcommand\applyleftsymb{\ident{applyleft}}
\mathcommand\applyleftpp[1]{\applyleftsymb\beginargs#1\allargs}
\newcommand\applyrightsymb{\ident{applyright}}
\mathcommand\applyrightpp[1]{\applyrightsymb\beginargs#1\allargs}
\newcommand\cardsymb{\ident{card}}
\mathcommand\cardpp[1]{\cardsymb\beginargs{#1}\allargs}
\newcommand\adtoldsymb{\ident{addr2ld}}
\mathcommand\adtoldpp[2]{\adtoldsymb\beginargs#1\separgs#2\allargs}
\newcommand\leftsymb{\ident{left}}
\remathcommand\leftpp[1]{\leftsymb\beginargs{#1}\allargs}
\newcommand\rightsymb{\ident{right}}
\remathcommand\rightpp[1]{\rightsymb\beginargs{#1}\allargs}
\newcommand\addrquestsymb{\ident{addr?}}
\mathcommand\addrquestpp[1]{\addrquestsymb\beginargs#1\allargs}
\newcommand\multiplicityquestsymb{\ident{multiplicity?}}
\mathcommand\multiplicityquestpp
[1]{\multiplicityquestsymb\beginargs#1\allargs}

\newcommand\charquestsymb{\ident{char?}}
\mathcommand\charquestpp[1]{\charquestsymb\beginargs#1\allargs}
\newcommand\chartoldsymb{\ident{char2ld}}
\mathcommand\chartoldpp[1]{\chartoldsymb\beginargs#1\allargs}
\newcommand\droplevelsymb{\ident{droplevel}}
\mathcommand\droplevelpp[1]{\droplevelsymb\beginargs{#1}\allargs}
\newcommand\droplocalitysymb{\ident{droplocality}}
\mathcommand\droplocalitypp[1]{\droplocalitysymb\beginargs#1\allargs}
\newcommand\sonssymb{\ident{sons}}
\mathcommand\sonspp[1]{\sonssymb\beginargs#1\allargs}
\newcommand\rootsymb{\ident{root}}
\mathcommand\rootpp[1]{\rootsymb\beginargs#1\allargs}
\newcommand\isproperpagesymb{\ident{isproperpage}}
\mathcommand\isproperpagepp[1]{\isproperpagesymb\beginargs#1\allargs}
\newcommand\mktreesymb{\ident{mktree}}
\mathcommand\mktreepp[2]{\mktreesymb\beginargs#1\separgs#2\allargs}
\newcommand\urtreesymb{\ident{urtree}}
\mathcommand\urtreepp[1]{\urtreesymb\beginargs#1\allargs}
\newcommand\headlinesymb{\ident{headline}}
\mathcommand\headlinepp[1]{\headlinesymb\beginargs#1\allargs}
\newcommand\attquestsymb{\ident{att?}}
\mathcommand\attquestpp[1]{\attquestsymb\beginargs#1\allargs}
\newcommand\okquestsymb{\ident{ok?}}
\mathcommand\okquestpp[1]{\okquestsymb\beginargs#1\allargs}
\newcommand\hdquestsymb{\ident{hd?}}
\mathcommand\hdquestpp[1]{\hdquestsymb\beginargs#1\allargs}
\newcommand\ldquestsymb{\ident{ld?}}
\mathcommand\ldquestpp[1]{\ldquestsymb\beginargs#1\allargs}
\newcommand\typequestsymb{\ident{type?}}
\mathcommand\typequestpp[1]{\typequestsymb\beginargs#1\allargs}
\newcommand\hdquestlistsymb{\ident{hd?list}}
\mathcommand\hdquestlistpp[1]{\hdquestlistsymb\beginargs#1\allargs}
\newcommand\attquestlistsymb{\ident{att?list}}
\mathcommand\attquestlistpp[1]{\attquestlistsymb\beginargs#1\allargs}
\newcommand\hdtoldsymb{\ident{hd2ld}}
\mathcommand\hdtoldpp[1]{\hdtoldsymb\beginargs#1\allargs}
\newcommand\locatesymb{\ident{locate}}
\mathcommand\locatepp[2]{\locatesymb\beginargs#1\separgs#2\allargs}
\newcommand\anchlocquestsymb{\ident{anchloc?}}
\mathcommand\anchlocquestpp[1]{\anchlocquestsymb\beginargs#1\allargs}
\newcommand\anchloctabquestsymb{\ident{anchloctab?}}
\mathcommand\anchloctabquestpp[1]{\anchloctabquestsymb\beginargs#1\allargs}
\newcommand\sourcequestsymb{\ident{source?}}
\mathcommand\sourcequestpp[1]{\sourcequestsymb\beginargs#1\allargs}
\newcommand\anchnamequestsymb{\ident{anchname?}}
\mathcommand\anchnamequestpp[1]{\anchnamequestsymb\beginargs#1\allargs}
\newcommand\anchproptabquestsymb{\ident{anchproptab?}}
\mathcommand\anchproptabquestpp[1]{\anchproptabquestsymb\beginargs#1\allargs}
\newcommand\linkquestsymb{\ident{link?}}
\mathcommand\linkquestpp[1]{\linkquestsymb\beginargs#1\allargs}
\newcommand\targetquestsymb{\ident{target?}}
\mathcommand\targetquestpp[1]{\targetquestsymb\beginargs#1\allargs}
\newcommand\uriquestsymb{\ident{uri?}}
\mathcommand\uriquestpp[1]{\uriquestsymb\beginargs#1\allargs}
\newcommand\uritohdsymb{\ident{uri2hd}}
\mathcommand\uritohdpp[1]{\uritohdsymb\beginargs#1\allargs}
\newcommand\addrtohdsymb{\ident{addr2hd}}
\mathcommand\addrtohdpp[1]{\addrtohdsymb\beginargs#1\allargs}
\newcommand\accesssymb{\ident{access}}
\mathcommand\accesspp[1]{\accesssymb\beginargs#1\allargs}
\newcommand\moquestsymb{\ident{mo?}}
\mathcommand\moquestpp[1]{\moquestsymb\beginargs#1\allargs}
\newcommand\emptyquestsymb{\ident{null}}
\mathcommand\emptyquestpp[1]{\emptyquestsymb\beginargs#1\allargs}
\newcommand\nthsymb{\ident{nth}}
\remathcommand\nthpp[2]{\nthsymb\beginargs#1\separgs#2\allargs}
\newcommand\pnthsymb{\ident{pnth}}
\mathcommand\pnthpp[2]{\pnthsymb\beginargs#1\separgs#2\allargs}
\newcommand\haspnthsymb{\ident{haspnth}}
\mathcommand\haspnthpp[2]{\haspnthsymb\beginargs#1\separgs#2\allargs}
\newcommand\haslocationsymb{\ident{haslocation}}
\mathcommand\haslocationpp[2]{\haslocationsymb\beginargs#1\separgs#2\allargs}
\newcommand\mklinksymb{\ident{mklink}}
\mathcommand\mklinkpp[4]
{\mklinksymb\beginargs#1\separgs#2\separgs#3\separgs#4\allargs}
\newcommand\mkldsymb{\ident{mkld}}
\mathcommand\mkldpp[3]{\mkldsymb\beginargs#1\separgs#2\separgs#3\allargs}
\newcommand\mkspecsymb{\ident{mkspec}}
\mathcommand\mkspecpp[2]
{\mkspecsymb\beginargs#1\separgs#2\allargs}
\newcommand\mkanchpropsymb{\ident{mkanchprop}}
\mathcommand\mkanchproppp[2]
{\mkanchpropsymb\beginargs#1\separgs#2\allargs}
\newcommand\mkanchsymb{\ident{mkanch}}
\mathcommand\mkanchpp
[4]{\mkanchsymb\beginargs#1\separgs#2\separgs#3\separgs#4\allargs}
\newcommand\ldtohdsymb{\ident{ld2hd}}
\mathcommand\ldtohdpp
[5]{\ldtohdsymb\beginargs#1\separgs#2\separgs#3\separgs#4\separgs#5\allargs}
\newcommand\mkhdsymb{\ident{mkhd}}
\mathcommand\mkhdpp
[6]{\mkhdsymb\beginargs#1\separgs#2\separgs#3\separgs#4\separgs#5\separgs
#6\allargs}
\mathcommand\mkhdpplong[6]
{\begin{array}[b]{@{}l@{}}\mkhdsymb\beginargs\\\ \ #1\separgs\\\ \ #2\separgs\\\ \ #3\separgs\\\ \ #4\separgs\\\ \ #5\separgs\\\ \ #6\allargs\end{array}}
\newcommand\nicemotohdsymb{\ident{mkhd}}
\mathcommand\nicemotohdpp
[4]{\nicemotohdsymb\beginargs#1\separgs#2\separgs#3\separgs#4\allargs}
\newcommand\motohdsymb{\ident{mo2hd}}
\mathcommand\motohdpp
[5]{\motohdsymb\beginargs#1\separgs#2\separgs#3\separgs#4\separgs#5\allargs}
\newcommand\ssymb{\ident s}
\remathcommand\spp[1]{\ssymb\beginargs#1\allargs}
\newcommand\notargetsymb{\ident{notarget}}
\mathcommand\notargetpp[1]{\notargetsymb\beginargs#1\allargs}
\newcommand\nosourcesymb{\ident{nosource}}
\mathcommand\nosourcepp[1]{\nosourcesymb\beginargs#1\allargs}
\newcommand\allocatesymb{\ident{allocate}}
\mathcommand\allocatepp[1]{\allocatesymb\beginargs#1\allargs}
\newcommand\hasaddrsymb{\ident{hasaddr}}
\mathcommand\hasaddrpp[1]{\hasaddrsymb\beginargs#1\allargs}
\newcommand\dimensionsymb{\ident{dimension}}
\mathcommand\dimensionpp[1]{\dimensionsymb\beginargs#1\allargs}

\mathcommand\beginargs{(}
\mathcommand\allargs  {)}
\mathcommand\separgs  {,\,}
\mathcommand\tightsepargs{,}

\mathcommand\minusppnoparentheses  [2]{{#1}\,\minussymbol\,{#2}}
\mathcommand\tightminusppnoparentheses  [2]{{#1}\minussymbol{#2}}
\mathcommand\divideppnoparentheses [2]{{#1}\,\dividesymbol\,{#2}}
\mathcommand\plusppnoparentheses   [2]{{#1}\,\plussymbol \,{#2}}
\mathcommand\plusppnoparenthesesi  [3]{{#2}\,\plussymbol^{#1}\,{#3}}
\mathcommand\tightplusppnoparentheses   [2]{{#1}\plussymbol{#2}}
\mathcommand\timesppnoparentheses  [2]{{#1}\,\timessymbol\,{#2}}
\mathcommand\undppnoparentheses    [2]{{#1}\und            {#2}}
\mathcommand\oderppnoparentheses   [2]{{#1}\oder           {#2}}
\mathcommand\impliesppnoparentheses[2]{{#1}\implies        {#2}}
\mathcommand\leqinfixppnoparentheses[2]{{#1}\,\tight\leq\,{#2}}
\mathcommand\geqinfixppnoparentheses[2]{{#1}\,\tight\geq\,{#2}}
\mathcommand\dividepp [2]{(\divideppnoparentheses {#1}{#2})}
\mathcommand\minuspp  [2]{(\minusppnoparentheses  {#1}{#2})}
\mathcommand\pluspp   [2]{(\plusppnoparentheses   {#1}{#2})}
\mathcommand\timespp  [2]{(\timesppnoparentheses  {#1}{#2})}
\mathcommand\undpp    [2]{(\undppnoparentheses    {#1}{#2})}
\mathcommand\oderpp   [2]{(\oderppnoparentheses   {#1}{#2})}
\mathcommand\impliespp[2]{(\impliesppnoparentheses{#1}{#2})}

\newcommand\myemptyset{\ensuremath{\mbox{\{\}}}}

\renewcommand\mycomment[1]{{\tt ---~{#1}~---}}

\newcommand\projections{\mycomment {Observer~Functions}}
\newcommand\edfunctions{\mycomment {Editing~Functions}}

\newcommand\sortsect
[1]{\\\SORTKEYWORD             \ {$\it\begin{array}[t]{l}#1\end{array}$}}

\newcommand\vissortsect
[1]{\\\VISSORTKEYWORD\\{$\it\begin{array}[t]{l}#1\end{array}$}}

\newcommand\conssect
[1]{\\\CONSTRUCTORKEYWORD\\{$\it\begin{array}[t]{l}#1\end{array}$}}

\newcommand\NONCONSTRUCTORKEYWORD{{\tt funs}}
\newcommand\funsect
[1]{\noindent\NONCONSTRUCTORKEYWORD
\\\math{\it\begin{array}[t]{l}#1\end{array}}}

\newcommand\deffunsect
[1]{\\\DEFUNSKEYWORD\\{$\it\begin{array}[t]{l}#1\end{array}$}}

\newcommand\varsect
[1]{\\\VARIABLEKEYWORD         \ {$\it\begin{array}[t]{l}#1\end{array}$}}

\newcommand\AXIOMKEYWORD{{\tt axioms}}
\newcommand\axiomsect[1]{\noindent\AXIOMKEYWORD#1}

\newcommand\phantomaxiomsect[1]{\\\math{\it\begin{array}[t]{l}#1\end{array}}}

\newcommand\commentsect
[1]{
\fbox
{
\begin{tabular}[t]{p{\textwidthminustwocm}}#1\end{tabular}}}

\newcommand\comfundef[3]
{{\yestop\noindent
\begin{tabular}{|p{\textwidthminushalfcm}|}\hline
\math{
\headroom#1\footroom}\\\hline\begin
{minipage}{\textwidthminushalfcm}{
\headroom
#2\footroom}\end{minipage}\\\hline\end{tabular}}
\\[1ex]\math{\begin{array}[t]{@{}l@{}}#3\end{array}}}

\mathcommand\simplefundef[1]{{\begin{array}{@{}l@{\,}ll@{}}#1\end{array}}}
  
\newcommand\vardecl  [2]{#1\stopq#2}

\newcommand\fundecl  [3]{#1\stopq#2\rightarrow#3}

\newcommand\mident[1]{{\miden #1}}
\newcommand\miden{\rm}

\newcommand\setsort{\mident{set}}
\newcommand\setof    [1]{\mbox{\setsort(#1)}}

\newcommand\listof   [1]{\mbox{\mident list(#1)}}

\newcommand\mapof    [2]{\mbox{\mident function(#1,#2)}}

\newcommand\treeof  [1]{\mbox{\mident tree(#1)}}

\newcommand\identsorts[2]{\mident{#1=#2}}
\newcommand\renamesort[2]{#1$\/\mapsto$#2}

\newcommand\andspec{\mident{\tt and}}
\newcommand\thenspec{\mident{\tt then}}

\newcommand\PPeq{=}

\newcommand\ENTRY{ENTRY}
\newcommand\entrysort{\mident{entry}}
\newcommand\entrysortwitharg[1]{\mident{entry#1}}

\newcommand\BOOL{BOOL}
\newcommand\boolsort{\mident{bool}}

\newcommand\true{\mident{true}}
\newcommand\false{\mident{false}}

\newcommand\NAT{NAT}
\newcommand\natsort{\mident{nat}}
\newcommand\succn{\mident{s}}
\newcommand\succpp[1]{\succn\beginargs{#1}\allargs}

\newcommand\LIST{LIST}
\newcommand\LISTPAIR{LISTPAIR}
\newcommand\listsort{\mident{list}}

\newcommand\myemptylist{\mident{[]}}

\newcommand\inslistpp[2]{#1::#2}

\newcommand\appendlistpp[2]{#1\mident{@}#2}
\newcommand\maxi{\mident{max}}
\mathcommand\maxipp[2]{\maxi\beginargs{#1}\separgs{#2}\allargs}
\newcommand\rep{\mident{repeat}}
\mathcommand\reppp[2]{\rep\beginargs{#1}\separgs{#2}\allargs}
\newcommand\isproperprefix{\mident{is\_proper\_prefix}}
\mathcommand\VMisproperprefixpp[1]{\isproperprefix\beginargs{#1}\allargs}
\mathcommand\VMisproperprefixtwopp
[2]{\isproperprefix\beginargs{#1}\separgs{#2}\allargs}

\newcommand\SET{SET}
\newcommand\MAPSET{MAPSET}
\newcommand\SETof[1]{\SET[#1]}

\newcommand\inserts{\mident{insert}}
\mathcommand\VMinsertpp[2]{\inserts\beginargs{#1}\separgs{#2}\allargs}
\newcommand\remsymb{\mident{rem}}
\mathcommand\rempp[2]{\remsymb\beginargs{#1}\separgs{#2}\allargs}
\newcommand\removes{\mident{dl}}
\mathcommand\removepp[2]{\removes\beginargs{#1}\separgs{#2}\allargs}
\newcommand\listmap{\mident{map}}
\mathcommand\listmappp[2]{\listmap\beginargs{#1}\separgs{#2}\allargs}
\newcommand\listpairmap{\mident{map}}
\mathcommand\listpairmappp[3]{\listpairmap\beginargs{#1}\separgs{#2}\separgs{#3}\allargs}
\newcommand\listpairmapdefault{\mident{map\_default}}
\mathcommand\listpairmapdefaultpp[5]{\listpairmapdefault\beginargs{#1}\separgs{#2}\separgs{#3}\separgs{#4}\separgs{#5}\allargs}
\newcommand\mapset{\mident{map\_set}}
\mathcommand\mapsetpp[2]{\mapset\beginargs{#1}\separgs{#2}\allargs}
\mathcommand\mapsetpplong
[2]{{\begin{array}[b]{@{}l@{}l@{}}\mapset\beginargs
&#1\separgs\\&#2\allargs\end{array}}}
\newcommand\mappset{\mident{map}}
\mathcommand\mappsetpp[3]{\mappset\beginargs{#1}\separgs{#2}\separgs{#3}\allargs}

\mathcommand\VMcardpp[1]{|{#1}|}

\newcommand\TREE{TREE}
\newcommand\treesort{\mident{tree}}
\newcommand\mktree{\mident{Mktree}}
\mathcommand\VMmktreepp[2]{\mktree\beginargs{#1}\separgs{#2}\allargs}

\newcommand\MAP{FUNCTION}
\newcommand\mapsort{\mident{function}}
\newcommand\MAPof[1]{\MAP[#1]}
\newcommand\domainsort{\mident{domain}}
\newcommand\rangesort{\mident{range}}
\newcommand\concatmap{\mident{map\_range}}
\newcommand\concatmappp[2]{\concatmap\beginargs#2\separgs#1\allargs}
\newcommand\domainS{\mident{dom}}
\mathcommand\domainSpp[1]{\domainS\beginargs{#1}\allargs}
\newcommand\rangeS{\mident{ran}}
\mathcommand\rangeSpp[1]{\rangeS\beginargs{#1}\allargs}
\newcommand\composemap{\mident{union}}
\mathcommand\composemappp[2]{\composemap\beginargs#1\separgs#2\allargs}

\newcommand\upd{\mident{upd}}
\mathcommand\updpp[3]{\upd\beginargs{#1}\separgs{#2}\separgs{#3}\allargs}
\mathcommand\updpplong
[3]{{\begin{array}[b]{@{}l@{}l@{}}\upd\beginargs
&#1\separgs\\&#2\separgs\\&#3\allargs\end{array}}}
\newcommand\revapplysymb{\mident{rev\_apply}}
\mathcommand\revapplypp[2]{\revapplysymb\beginargs#1\separgs#2\allargs}

\newcommand\CHAR{CHAR}

\newcommand\STRING{STRING}

\newcommand\BASISP{DOCUMENT\_P}
\newcommand\basisparamsort{\mident{document}}
\newcommand\locationparamsort{\mident{location}}

\newcommand\ADDRP{ADDR\_P}
\newcommand\addrparamsort{\mident{addr}}

\newcommand\ATTj[1]{ATT\_#1}
\newcommand\attjsort[1]{\mident{att\_#1}}
\newcommand\concat{\mident{concat}}
\mathcommand\concatpp[2]{\concat\beginargs{#1}\separgs{#2}\allargs}
\newcommand\removeatt{\mident{remove}}
\mathcommand\removeattpp[2]{\removeatt\beginargs{#1}\separgs{#2}\allargs}
\newcommand\emptyatt{\mident{[]_{Att}}}

\newcommand\URI{URI}
\newcommand\urisort{\mident{uri}}

\newcommand\ADDR[1]{\mident{#1\_ADDR}}

\newcommand\addrisort[1]{\mident{#1\_addr}}

\newcommand\structitypesort[1]{\mident{{#1}\_struct}}
\newcommand\locisort[1]{\mident{{#1}\_location}}
\newcommand\hdtable{\mident{Table}}
\newcommand\hdtableline{\mident{Tableline}}
\newcommand\hdlist{\mident{Page\_list}}
\newcommand\hdtext{\mident{Text}}
\newcommand\hdfootnote{\mident{Footnote}}
\newcommand\hdheadline{\mident{Headline}}
\newcommand\hdparagraph{\mident{Paragraph}}
\newcommand\hdpage{\mident{Minipage}}

\newcommand\hdcopyright{\mident{Copyright}}
\newcommand\hdlinebreak{\mident{Br}}
\newcommand\hdhframeset{\mident{Horiz\_frameset}}
\newcommand\hdvframeset{\mident{Vert\_frameset}}
\newcommand\hdaframeset{\mident{Alt\_frameset}}
\newcommand\hdsitemap{\mident{sitemap}}

\newcommand\SYMBOL[1]{#1\_SYMBOLS}
\newcommand\symbolisort[1]{\mident{#1\_symbols}}

\newcommand\ANCHORNAME{ANCHOR\_ID}
\newcommand\anchornamesort{\mident{anchor\_id}}

\newcommand\ANCHORCLASS{ANCHOR}

\newcommand\anchorsort{\mident{anchor}}
\newcommand\anchortypesort{\mident{\anchorsort\_type}}

\newcommand\mkanchor{\mident{Mkanchor}}
\mathcommand\mkanchorpp[3]{\mkanchor\beginargs{#1}\separgs{#2}\separgs{#3}\allargs}
\mathcommand\mkanchorpplong
[3]{{\begin{array}[b]{@{}l@{}l@{}}\mkanchor\beginargs&#1
\separgs\\&#2
\separgs\\&#3\allargs\end{array}}}
\newcommand\labelpp{\mident{Label}}
\newcommand\sourcepp{\mident{Source}}
\newcommand\targetpp{\mident{Target}}

\newcommand\locSanchor{\mident{get\_location}}
\mathcommand\locSanchorpp[1]{\locSanchor\beginargs{#1}\allargs}
\newcommand\typeSanchor{\mident{get\_type}}
\mathcommand\typeSanchorpp[1]{\typeSanchor\beginargs{#1}\allargs}
\newcommand\attSanchor{\mident{get\_att}}
\mathcommand\attSanchorpp[1]{\attSanchor\beginargs{#1}\allargs}

\newcommand\maxSanchor{\mident{suptype}}
\mathcommand\maxSanchorpp[2]{\maxSanchor\beginargs{#1}\separgs{#2}\allargs}

\newcommand\chanchortype{\mident{ch\_type}}
\mathcommand\chanchortypepp[2]{\chanchortype\beginargs{#1}\separgs{#2}\allargs}
\newcommand\chlocation{\mident{ch\_location}}
\mathcommand\chlocationpp[2]{\chlocation\beginargs{#1}\separgs{#2}\allargs}
\newcommand\addattributeanchor{\mident{add\_attribute}}
\mathcommand\addattributeanchorpp[2]{\addattributeanchor\beginargs{#1}\separgs{#2}\allargs}
\newcommand\delattributeanchor{\mident{del\_attribute}}
\mathcommand\delattributeanchorpp[2]{\delattributeanchor\beginargs{#1}\separgs{#2}\allargs}
\newcommand\sinklocation{\mident{sinkloc}}
\mathcommand\sinklocationmappp[1]{\sinklocation\beginargs{#1}\allargs}
\mathcommand\sinklocationpp[2]{\sinklocation\beginargs{#1}\separgs{#2}\allargs}

\newcommand\LINKCLASS{LINK}

\newcommand\linksort{\mident{link}}
\newcommand\linktypesort{\mident{\linksort\_type}}
\newcommand\showsort{\mident{show}}
\newcommand\actuatesort{\mident{actuate}}
\newcommand\specifiersort{\mident{specifier}}

\newcommand\mkspecifier{\mident{Mkspecifier}}
\mathcommand\mkspecifierpp[2]{\mkspecifier\beginargs{#1}\separgs{#2}\allargs}
\newcommand\mklink{\mident{Mklink}}
\mathcommand\VMmklinkpp[4]{\mklink\beginargs{#1}\separgs{#2}\separgs{#3}\separgs{#4}\allargs}
\newcommand\embedpp{\mident{Embed}}
\newcommand\VMreplacepp{\mident{Replace}}
\newcommand\VMnewpp{\mident{New\_window}}
\newcommand\userpp{\mident{User}}
\newcommand\autopp{\mident{Auto}}
\newcommand\bipp{\mident{Bi}}
\newcommand\mkuni{\mident{Uni}}
\mathcommand\mkunipp[2]{\mkuni\beginargs{#1}\separgs{#2}\allargs}

\newcommand\attSlink{\mident{get\_att}}
\mathcommand\attSlinkpp[1]{\attSlink\beginargs{#1}\allargs}
\newcommand\typeSlink{\mident{get\_type}}
\mathcommand\typeSlinkpp[1]{\typeSlink\beginargs{#1}\allargs}
\newcommand\specifierSlink{\mident{get\_specifier}}
\mathcommand\specifierSlinkpp[1]{\specifierSlink\beginargs{#1}\allargs}
\newcommand\sourceSlink{\mident{get\_source}}
\mathcommand\sourceSlinkpp[1]{\sourceSlink\beginargs{#1}\allargs}
\newcommand\targetSlink{\mident{get\_target}}
\mathcommand\targetSlinkpp[1]{\targetSlink\beginargs{#1}\allargs}
\newcommand\addSlink{\mident{get\_uri}}
\mathcommand\addSlinkpp[1]{\addSlink\beginargs{#1}\allargs}
\newcommand\anchSlink{\mident{get\_id}}
\mathcommand\anchSlinkpp[1]{\anchSlink\beginargs{#1}\allargs}

\newcommand\chaddrspec{\mident{ch\_uri}}
\mathcommand\chaddrspecpp[2]{\chaddrspec\beginargs{#1}\separgs{#2}\allargs}
\newcommand\chname{\mident{ch\_id}}
\mathcommand\chnamepp[2]{\chname\beginargs{#1}\separgs{#2}\allargs}
\newcommand\inserttarget{\mident{insert\_target}}
\mathcommand\inserttargetpp[2]{\inserttarget\beginargs{#1}\separgs{#2}\allargs}
\newcommand\deletetarget{\mident{delete\_target}}
\mathcommand\deletetargetpp[2]{\deletetarget\beginargs{#1}\separgs{#2}\allargs}
\newcommand\insertsource{\mident{insert\_source}}
\mathcommand\insertsourcepp[2]{\insertsource\beginargs{#1}\separgs{#2}\allargs}
\newcommand\deletesource{\mident{delete\_source}}
\mathcommand\deletesourcepp[2]{\deletesource\beginargs{#1}\separgs{#2}\allargs}
\newcommand\chlinktype{\mident{ch\_type}}
\mathcommand\chlinktypepp[2]{\chlinktype\beginargs{#1}\separgs{#2}\allargs}
\newcommand\addattributelink{\mident{add\_attribute}}
\mathcommand\addattributelinkpp[2]{\addattributelink\beginargs{#1}\separgs{#2}\allargs}
\newcommand\delattributelink{\mident{del\_attribute}}
\mathcommand\delattributelinkpp[2]{\delattributelink\beginargs{#1}\separgs{#2}\allargs}
\newcommand\replaceaddrspec{\mident{replace\_uri\_sp}}
\mathcommand\VMreplaceaddrspecpp[3]{\replaceaddrspec\beginargs{#1}\separgs{#2}\separgs{#3}\allargs}
\mathcommand\VMreplaceaddrspecmappp[2]{\replaceaddrspec\beginargs{#1}\separgs{#2}\allargs}
\newcommand\combinelink{\mident{combine\_link}}
\mathcommand\combinelinkpp[5]{\combinelink\beginargs{#1}\separgs{#2}\separgs{#3}\separgs{#4}\separgs{#5}\allargs}
\newcommand\replaceaddrlink{\mident{replace\_uri\_li}}
\mathcommand\VMreplaceaddrlinkpp[3]{\replaceaddrlink\beginargs{#1}\separgs{#2}\separgs{#3}\allargs}

\newcommand\HDCLASS{HD}

\newcommand\hdsort{\mident{hd}}

\newcommand\mkhd{\mident{Mkhd}}
\mathcommand\VMmkhdpp[5]{\mkhd\beginargs{#1}\separgs{#2}\separgs{#3}\separgs{#4}\separgs{#5}\allargs}

\mathcommand\VMmkhdpplong
[5]{{\begin{array}[b]{@{}l@{}l@{}}\mkhd\beginargs&#1
\separgs\\&#2
\separgs\\&#3
\separgs\\&#4
\separgs\\&#5\allargs\end{array}}}

\newcommand\addShd{\mident{get\_addr}}
\mathcommand\addShdpp[1]{\addShd\beginargs{#1}\allargs}
\newcommand\attShd{\mident{get\_att}}
\mathcommand\attShdpp[1]{\attShd\beginargs{#1}\allargs}
\mathcommand\ldShd{\mident{|\hspace*{-1.6pt}|\_|\hspace*{-1.6pt}|}}
\mathcommand\ldShdpp[1]{|\hspace*{-1.6pt}|{#1}|\hspace*{-1.6pt}|}
\newcommand\linkShd{\mident{get\_link}}
\mathcommand\linkShdpp[1]{\linkShd\beginargs{#1}\allargs}
\newcommand\anchornameShd{\mident{get\_anchor\_id}}
\mathcommand\anchornameShdpp[2]{\anchornameShd\beginargs{#1}\separgs{#2}\allargs}
\newcommand\anchorShd{\mident{get\_anchor}}
\mathcommand\anchorShdpp[2]{\anchorShd\beginargs{#1}\separgs{#2}\allargs}
\newcommand\anchorsShd{\mident{get\_anchors}}
\mathcommand\anchorsShdpp[1]{\anchorsShd\beginargs{#1}\allargs}

\newcommand\chaddr{\mident{ch\_addr}}
\mathcommand\chaddrpp[2]{\chaddr\beginargs{#1}\separgs{#2}\allargs}
\newcommand\addanchor{\mident{add\_anchor}}
\mathcommand\addanchorpp[3]{\addanchor\beginargs{#1}\separgs{#2}\separgs{#3}\allargs}
\newcommand\delanchor{\mident{del\_anchor}}
\mathcommand\delanchorpp[2]{\delanchor\beginargs{#1}\separgs{#2}\allargs}
\newcommand\addlink{\mident{add\_link}}
\mathcommand\VMaddlinkpp[2]{\addlink\beginargs{#1}\separgs{#2}\allargs}
\newcommand\dellink{\mident{del\_link}}
\mathcommand\dellinkpp[2]{\dellink\beginargs{#1}\separgs{#2}\allargs}
\newcommand\addattributehd{\mident{add\_attribute}}
\mathcommand\addattributehdpp[2]{\addattributehd\beginargs{#1}\separgs{#2}\allargs}
\newcommand\delattributehd{\mident{del\_attribute}}
\mathcommand\delattributehdpp[2]{\delattributehd\beginargs{#1}\separgs{#2}\allargs}
\newcommand\combineanchor{\mident{combine\_anchor}}
\mathcommand\combineanchorpp[3]{\combineanchor\beginargs{#1}\separgs{#2}\separgs{#3}\allargs}

\newcommand\integrate{\mident{embed}}
\mathcommand\integratepp[1]{\integrate\beginargs{#1}\allargs}

\newcommand\MO{MO}
\newcommand\mosort{\mident{\mident{mo}}}
\newcommand\mkmo{\mident{Mkmo}}
\mathcommand\mkmopp[2]{\mkmo\beginargs{#1}\separgs{#2}\allargs}

\newcommand\mtlocS{\mident{empty?}}
\mathcommand\mtlocSpp[1]{\mtlocS\beginargs{#1}\allargs}

\newcommand\embedlinkokS{\mident{embed\_link\_ok?}}
\mathcommand\embedlinkokSpp[2]{\embedlinkokS\beginargs{#1}\separgs{#2}\allargs}

\newcommand\LDONE{\mident{PAGE}}

\newcommand\ldonesort{\mident{page}}

\newcommand\emptypage{\mident{Emptypage}}
\newcommand\basic{\mident{Basic}}
\newcommand\symb{\mident{Symbol}}
\newcommand\mtpage{\mident{[\hspace*{-1.6pt}[]\hspace*{-1.6pt}]}}
\newcommand\mtpagepp[1]{\mtpage}
\newcommand\impsymbol{\mident{''\_''}}
\mathcommand\impsymbolpp[1]{''{#1}''}
\newcommand\imphdlow{\mident{[\hspace*{-1.6pt}[\_]\hspace*{-1.6pt}]}}
\mathcommand\imphdlowpp[1]{[\hspace*{-1.6pt}[{#1}]\hspace*{-1.6pt}]}
\newcommand\mkld{\mident{Mkpage}}
\mathcommand\VMmkldpp[3]{\mkld\beginargs{#1}\separgs{#2}\separgs{#3}\allargs}

\newcommand\structSld{\mident{get\_struct}}
\mathcommand\structSldpp[1]{\structSld\beginargs{#1}\allargs}
\newcommand\pagesSld{\mident{get\_pages}}
\mathcommand\pagesSldpp[1]{\pagesSld\beginargs{#1}\allargs}
\newcommand\attSld{\mident{get\_att}}
\mathcommand\attSldpp[1]{\attSld\beginargs{#1}\allargs}

\newcommand\includelinkokldoneS{\mident{embed\_link\_\ldonesort\_ok?}}
\mathcommand\includelinkokldoneSpp[2]{\includelinkokldoneS\beginargs{#1}\separgs{#2}\allargs}
\newcommand\hasnthS{\mident{has\_pnth?}}
\mathcommand\hasnthSpp[2]{\hasnthS\beginargs{#1}\separgs{#2}\allargs}
\newcommand\haslocationS{\mident{has\_location?}}
\mathcommand\haslocationSpp[2]{\haslocationS\beginargs{#1}\separgs{#2}\allargs}
\newcommand\atomicS{\mident{atomic?}}
\mathcommand\atomicSpp[1]{\atomicS\beginargs{#1}\allargs}

\newcommand\pnth{\mident{pnth}}
\mathcommand\VMpnthpp[2]{\pnth\beginargs{#1}\separgs{#2}\allargs}
\newcommand\changestruct{\mident{ch\_struct}}
\mathcommand\changestructpp[2]{\changestruct\beginargs{#1}\separgs{#2}\allargs}
\newcommand\addattributeld{\mident{add\_attribute}}
\mathcommand\addattributeldpp[2]{\addattributeld\beginargs{#1}\separgs{#2}\allargs}
\newcommand\delattributeld{\mident{del\_attribute}}
\mathcommand\delattributeldpp[2]{\delattributeld\beginargs{#1}\separgs{#2}\allargs}
\newcommand\insertatld{\mident{insert\_at}}
\mathcommand\insertatldpp[3]{\insertatld\beginargs{#1}\separgs{#2}\separgs{#3}\allargs}
\newcommand\insertatlde{\mident{place\_at}}
\mathcommand\insertatldepp[3]{\insertatlde\beginargs{#1}\separgs{#2}\separgs{#3}\allargs}
\newcommand\insertatldee{\mident{place\_at\_help}}
\mathcommand\insertatldeepp[5]{\insertatldee\beginargs{#1}\separgs{#2}\separgs{#3}\separgs{#4}\separgs{#5}\allargs}
\newcommand\mktable{\mident{mktable}}
\mathcommand\VMmktablepp[2]{\mktable\beginargs{#1}\separgs{#2}\allargs}
\newcommand\mktableline{\mident{mktableline}}
\mathcommand\VMmktablelinepp[1]{\mktableline\beginargs{#1}\allargs}
\newcommand\mklist{\mident{mklist}}
\mathcommand\mklistpp[1]{\mklist\beginargs{#1}\allargs}
\newcommand\locate{\mident{locate}}
\mathcommand\VMlocatepp[2]{\locate\beginargs{#1}\separgs{#2}\allargs}
\newcommand\dimensionpagesymb{\mident{page\_dimension}}
\newcommand\dimensionpagelistsymb{\mident{page\_list\_dimension}}
\mathcommand\dimensionpagepp[1]{\dimensionpagesymb\beginargs{#1}\allargs}
\mathcommand\dimensionpagelistpp
[1]{\dimensionpagelistsymb\beginargs{#1}\allargs}

\newcommand\HDONE{\mident{HMD}}
\newcommand\hdonesort{\mident{hmd}}

\newcommand\insertathmd{\mident{insert\_at}}
\mathcommand\insertathmdpp[4]{\insertathmd\beginargs{#1}\separgs{#2}\separgs{#3}\separgs{#4}\allargs}
\newcommand\insertathmde{\mident{place\_at}}
\mathcommand\insertathmdepp[4]{\insertathmde\beginargs{#1}\separgs{#2}\separgs{#3}\separgs{#4}\allargs}

\newcommand\LDTWO{\mident{CHAPTER}}

\newcommand\ldtwosort{\mident{chapter}}

\newcommand\HDTWO{FSD}
\newcommand\hdtwosort{\mident{fsd}}

\newcommand\includelinkokldtwoS{\mident{include\_link\_\ldtwosort\_ok?}}
\mathcommand\includelinkokldtwoSpp[2]{\includelinkokldtwoS\beginargs{#1}\separgs{#2}\allargs}

\mathcommand\insertatfsdpp[3]{\insertathd\beginargs{#1}\separgs{#2}\separgs{#3}\allargs}

\mathcommand\insertatfsdepp[3]{\insertathde\beginargs{#1}\separgs{#2}\separgs{#3}\allargs}

\newcommand\LDTHREE{\mident{BOOK}}

\newcommand\ldthreesort{\mident{book}}

\newcommand\HDTHREE{SITE}
\newcommand\hdthreesort{\mident{site}}

\newcommand\includelinkokldthreeS{\mident{include\_link\_\ldthreesort\_ok?}}
\mathcommand\includelinkokldthreeSpp[2]{\includelinkokldthreeS\beginargs{#1}\separgs{#2}\allargs}

\mathcommand\insertatsitepp[3]{\insertathd\beginargs{#1}\separgs{#2}\separgs{#3}\allargs}

\mathcommand\insertatsiteepp[3]{\insertathde\beginargs{#1}\separgs{#2}\separgs{#3}\allargs}

\newcommand\Fsd{Frameset-Document}

\newcommand\Hd{Hyperdocument}
\newcommand\hd{hyperdocument}

\newcommand\Hmd{Hypermedia-Document}
\newcommand\hmd{hypermedia-document}

\newcommand\Mo{Media-Object}
\newcommand\mo{media-object}

\newcommand\Site{Site}

\newcommand\paper{paper}
\newcommand\tightemph[1]{{\em #1}\/} 

\setshortbibstyle

\begin{document}
\setcounter{page}{-99}
{\flushbottom
\thispagestyle{empty}

\noindent
\begin{minipage}{\textwidth}
\mbox{}

\vspace{9.0em}
\begin{center}
{\LARGE\bf
An Algebraic Dexter-Based 
\\Hypertext Reference Model
\\\mbox{}
}
{
\\\mbox{}
\\\mbox{}
\\
\mattickname, \wirthname
\\\mbox{}
\\\mbox{}
\\
{\tt volker.mattick@cs.tu-dortmund.de, 
\\\email
\\
http://ls1-www.cs.uni-dortmund.de/cms/mattick.html
\\\www}
\\\mbox{}
\\\mbox{}
\\\mbox{}
\\
}
{\small
Research Report 719/1999
\\
{\tt http://www.ags.uni-sb.de/~cp/p/gr719}
\\
\mbox{}\\
November~6, 1999
\\
\mbox{}
\\
\mbox{}\\
\uniDO
\\
Fakult\"at f\"ur Informatik
\\
44227 \DO
\\
Germany
\\
}

\end{center}
\end{minipage}

\vspace{\fill}
{\small

\noindent
{\bf Abstract:}
\renewcommand{\baselinestretch}{0.8}
We present the first formal algebraic specification of
a hypertext reference model.
It is based on the well-known Dexter Hypertext Reference Model
and includes modifications with respect to the development 
of hypertext since the \WWW\
came up. 
Our hypertext model was developed as a product model 
with the aim to automatically support the design process
and is extended to a model of hypertext-systems in order 
to be able to describe the state transitions in this process.
While the specification should be easy to read for 
non-experts in algebraic specification,
it guarantees a unique understanding and 
enables a close connection to logic-based development
and verification.
\par}


}\raggedbottom

\setcounter{tocdepth}{2}
\pagestyle{empty}
\tableofcontents\nopagebreak
\vfill\vfill
 

\pagestyle{myheadings}%

\flushbottom
\pagenumbering{arabic}
\setcounter{page}{1}
\section{Introduction}

The number of hypertext applications is growing. 
What started as an idea of \bushname\ 
more than half a century ago (\cf\ \cite{bus45}) 
has now become one of the most rapidly growing fields in software engineering.
The reason for this rapid 
development is the World-Wide Web \nolinebreak(\tightemph\WWW). 

Nearly all of the web sites used nowadays are hypermedia applications
and only a few are mere hypertexts.
In this \paper\ we will refer the term\emph{hypermedia} to a 
 combination of\emph{hypertext} and\emph{multimedia}, 
 as suggested \eg\ in \cite{hbr94}.  
If the textual or multimedial nature is not relevant, 
we will speak of\emph{hyperdocuments}. 
As hypermedia is an open approach, 
there are infinitely many different types of \mo s in principle.
In a closed reference model, these different types of \mo s can only be
modeled with an abstract interface. Therefore, it seems to be justified
to speak of a\emph{hypertext} reference model even for models of 
hypermedia like the one we are going to specify in this \paper.

Our hypertext reference model is\emph{Dexter-based} because 
it deviates from the Dexter Hypertext Reference Model (\cf\ \cite{hal90}) 
only in some aspects that had to be corrected
in order to be compatible with the \WWW.
A detailed comparison with the Dexter model, however, is not subject
of this \paper.

The hypertext model (\cf\ \sectref{datatype}) 
was developed as a product model 
with the aim to support the design of the product 
``hyperdocument'' automatically. 
It is extended to a model of hypertext-systems (\cf\ \sectref{extdatatype}) 
in order to describe the state transitions of the design-process.  

To our knowledge, our hypertext reference model is the
first\footnote{Note that we do not consider Z to be a formal algebraic
specification language.}\emph{formal algebraic} modeling approach for
hypertexts, hypermedia, or hypertext-systems. 
Algebraic specification came up in the seventies 
based on concepts of universal algebra and abstract datatypes.
Due to the technical complexity of the subject, it is still an
area of ongoing research on the one hand. 
On the other hand,
there is still a gap between what practice demands and 
what theory delivers.
One motivation for our work
is to make this gap a little smaller
and we hope that our specification is quite readable 
for non-experts in algebraic specification.
Due to its origin, algebraic specification is
superior to other specification
formalisms in its clear relation to logic and semantics
that guarantees a unique understanding and 
enables a close connection to logic-based development
and verification.




\subsection{Product Models for Hyperdocuments}

In the domain of hyperdocuments there are three fundamental different kinds
of product models 
(\cf\ \cite[\p\,221\,\ff]{low99}).  Programming language
based, information-centered and screen-based models.  The programming
language based approach, which applies any general purpose programming
language starting from scratch, was used in former days due to the
lack of any other sophisticated models, and has nearly no importance
in the presence.

For a long time the\emph{information-centered model} has dominated.   
The most popular product model for hyperdocuments, 
the ``Dexter Hypertext Reference Model'' \cite{hal90}, 
is information-centered.
Dexter or one of its modifications, \eg\ \cite{gro94} or \cite{oss95b},
describe the structure of a hyperdocument, divided into its logical structure,
its linkage, and its style. 
A hyperdocument can import components from a 
``within-component layer'' 
via an anchor mechanism and specify how the 
document should be
presented in a ``presentation specification''. 

Similar ideas are presented in an object-oriented style in the 
so-called ``Tower Model'', \cf\ \cite{bra92}.
Additionally a hierarchization is added. It is described that
components could include other components. 
But no restrictions on how to compose hyperdocuments are mentioned.
Thus, 
you can produce a lot of components not used in any actual hypermedia system. 

In both models 
there is no possibility to describe strategies how to navigate 
through a set of hyperdocuments. But this is a design goal of increasing importance
in the rapidly growing world of hypermedia. 
The Dexter-based reference model
for adaptive hypermedia\emph{(AHAM)} 
(\cf\ \cite{bra98}) describes first steps towards this direction.

Even the wide-spread Hypertext Markup Language (\tightemph{HTML}) has 
obviously its roots in the information-centered paradigm, 
even though many designers use it in an other way,
namely as a screen-based design language. 
``Screen-based'' means that the focus is not the logical
structure of the document, enriched with some display attributes, but the display of the
document itself. 
With the upcoming of the \WWW\ 
and the WYSIWYG-editors the\emph{screen-based model} became 
more important in hyperdocument design, 
because sometimes it is easier to think in terms of
the produced view on the screen. 
As far as we know, there are only two models for this approach:
The Document Object Model (\tightemph{DOM}, \cf\ \cite{w3c98b}) 
and the Document Presentation Language (P Language) 
of\emph{THOT} (\cite{qui97}). 
The goal of DOM is to define an application programming interface for XML and
HTML. 
Thus it is limited to the features used in that languages. 
The P Language of THOT,
used by the W3C-test-bed client browser Amaya (\cite{gue98}), 
is more general, but lacks device-independence; \ie\ 
the presentation only describes a function of the structure of the documents 
and the image that would be produced on an idealized device. 

\subsection{Semantics for Hyperdocument Models}
All hyperdocument models have in common that no explicit semantics is given. 
Some information-centered models try to treat the structural
part of a hyperdocument as a data type and assign a semantics, 
but no semantics for the attributes is given. 
\Eg, DOM reduces the DOM-semantics to the semantics of HTML, 
but up to now there is no unique semantics for HTML, 
but only device- and browser-dependent semantics. 

But there are two widely accepted 
device-independent description formalisms for documents: 
The postscript- and the PDF-format (\cite{pdf96}). 
Postscript is very mighty but lacks the hyperlinks. 
Hence, we will use PDF as a screen-based model for hyperdocuments.

\begin{figure}
\centerline{\epsfxsize\textwidth\epsfysize 11.0cm\epsfbox{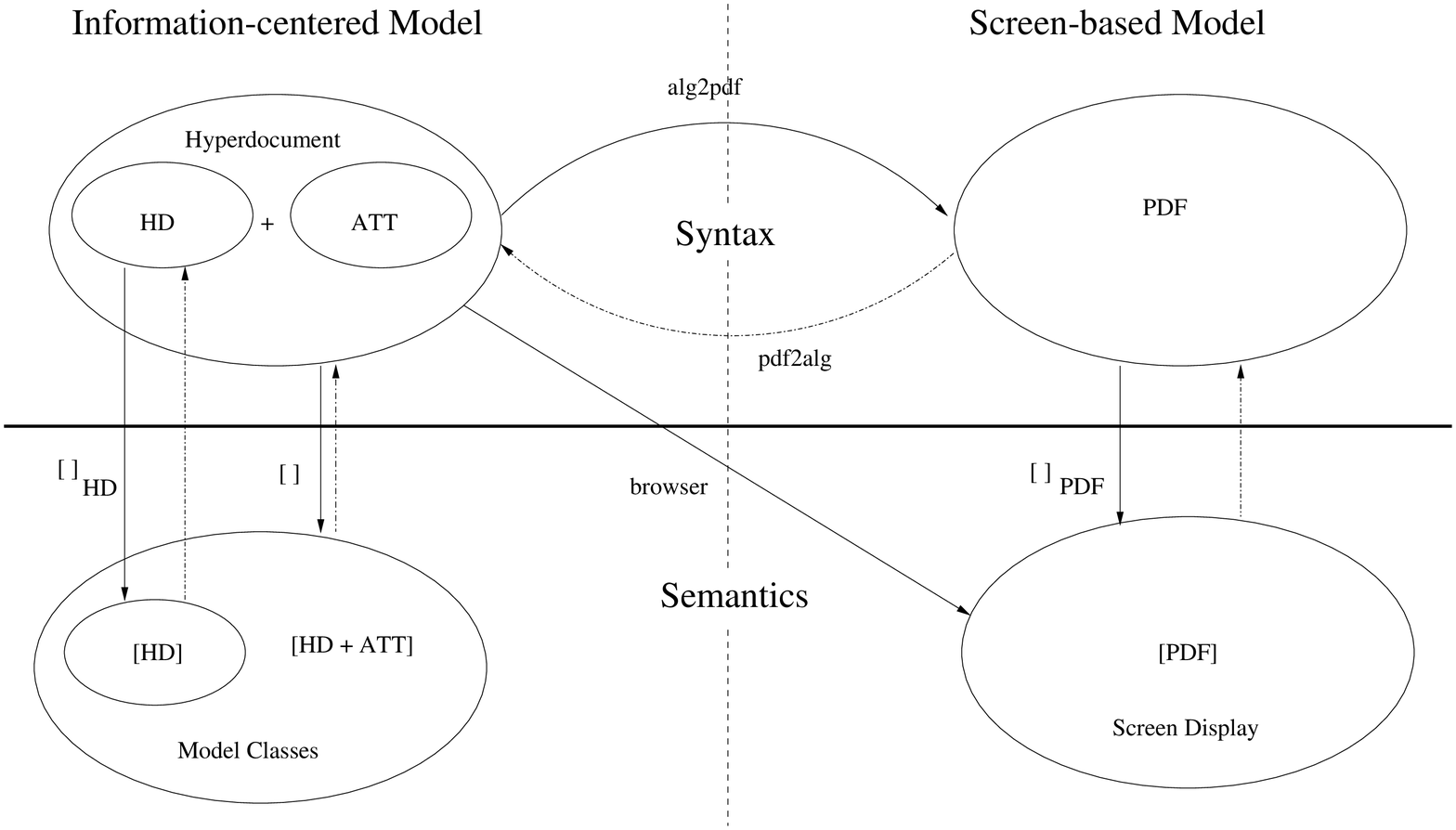}}
\caption{Browser\label{overview}}
\end{figure}

Both kinds of models, 
the screen-based and the information-centered, 
have in common that they abstract from the contents to be displayed. 
In practice the gap between both is bridged by a user agent, 
often called\emph{browser}, \cf\ Fig.~\vref{overview}. 
A browser is a mapping between the syntax of the information-centered model 
of hyperdocuments and the semantics of the screen-based model.
It should be equal to the concatenation of the translation (alg2pdf) from the
algebraic signature of \hd s into the language of PDF and 
a display mapping ($[\,]_{\rm PDF}$) 
assigning the semantics to the screen-based PDF-model. 
Thus, the semantics of an information-centered description of \hd s 
is defined in terms of 
the semantics of a well-known description language for documents.
Up to now it is an enormous problem both for browser developers and for 
designers that there is no unique meaning for a \hd, 
but only meanings together with particular browsers and output devices. 
Note that the lower left corner of Fig.\vref{overview} denotes some
model class providing the algebraic or logic semantics of the 
information-centered model.

Another problem with the existing models is 
that they do not reflect the actual state of hypertext technology.
Two of the three information-centered models mentioned above come from 
the ``pre-\WWW'' times.

Therefore we will not formalize the models as
they are, but use their crucial ideas, add some new ones coming up with 
the \WWW\ and structure a document in analogy to classical linear text. 
We will describe all this in an algebraic specification language
(\cf\ \cite{pad99}) enriched with a modularity concept, \cf\ \asfplus\ 
(\cite{lundewirth})
or cTLA (\cite{mes95}). 

This paper deals mainly with the upper left part of Figure~\vref{overview} and
the relation to its neighbors. The formalization of the screen-based model as
well as the formal description of the browser defining mapping will be 
left to another paper. 
In \sectref{datatype} we start with the formal description of the 
general hyperdocument data-structure and identify different
hierarchy levels, similar to the levels in linear texts. 
In \sectref{extdatatype} the extension to a model for hypertext-systems
is presented. 
\clearpage
\section{The Information-Centered Model for Hypermedia}
\label{datatype}

\begin{quote}
``It is essential to have a solid understanding of the kinds of 
information present during a design process before the design process 
itself can be studied.'' 
\cite{sal96}.
\end{quote}
The\emph{product model} (sometimes called ``object model'') is a formal
representation of exactly the above-mentioned kinds of information.
For our description formalism we choose the constructor-based algebraic
approach (\cf\ \cite{pad99}, \cite{kwspec}). 
In section \ref{algapproach} we describe 
how that formalism can be transferred to our domain.
In section \ref{prodmod} we develop our product model for hypermedia
documents and compare it with existing reference models, like
the Dexter Model \cite{hal90} or the Tower Model \cite{bra92}, and
with certain standards as, \eg, 
those used by the World Wide Web Consortium (W3C)
for defining XML \cite{w3c98d}. 

\subsection{Algebraic Specifications for Describing Product 
Models}\label{algapproach}

In classical first-order algebraic specifications, the world 
is represented with the help of a signature. 
A\emph{signature} \math{\sig=\pair\sigfunsym\sigarity}
consists of an (enumerable) set of function symbols 
\nolinebreak\sigfunsym\
and a (computable) arity function \FUNDEF\sigarity\sigfunsym\N,
saying that each function symbol \math{f\in\sigfunsym} takes \app\sigarity f
arguments. 
A corresponding \sig-algebra (or \sig-structure) consists
of a single homogeneous universe (or carrier) and,  
for each function symbol in \sigfunsym,
a total function on this universe.

Heterogeneous, however, is the world we have to model.\footnote
{For a more detailed discussion \cf\ \cite{len92}.}
We have at least three different sorts 
of objects: {\em anchors} (\cf\ \sectref{anchordomain}), {\em links} 
(\ref{linkdomain}) and {\em documents} (\ref{hddomain}). 
Therefore, an adequate structural representation
should contain different universes for different sorts.
This leads us to the following refinement of the notion of 
a signature.

A\emph{many-sorted signature} \math{\sig=\trip\sigsorts\sigfunsym\sigarity}
consists of a (finite) set of sorts \sigsorts,
an (enumerable) set of function symbols \sigfunsym\
and a (computable) arity function 
\FUNDEF\sigarity\sigfunsym{\transclosureinline\sigsorts},
saying that each function symbol \math{f\in\sigfunsym} 
with \bigmath{\app\sigarity f= s_1\ldots s_n s'}
takes \math n arguments of the sorts \math{s_1,\ldots,s_n}
and produces a term of sort \math{s'}.
A corresponding \sig-algebra \algebra\ 
consists
of a separate universe \math{\algebra_s} for each sort \math{s\in\sigsorts}
and, for each function symbol \math{f\in\sigfunsym} with 
\bigmath{\app\sigarity f= s_1\ldots s_n s'
,}
a total function 
\FUNDEF
  {f^\algebra}
  {\algebra_{s_0}\tighttimes\ldots\tighttimes\algebra_{s_n}}
  {\algebra_{s'}}.

Typically, certain function symbols are called ``constructors''
because they construct the data domains 
(or domains of discourse) of an algebra.
More precisely, the constructor (ground) terms\footnote
{\Ie\
 the well-sorted terms built-up solely from constructor function symbols.}
are used for designating the data items of an algebra\emph{completely} 
and\emph{uniquely};
the popular catchwords being\emph{no junk} and\emph{no confusion}, \resp.
\Eg, zero `0' and successor `\ssymb' may construct the sort of 
natural numbers `\nat',
`nil' and `cons' the lists, `\true' and `\false' the Boolean sort, \etc.
For the sort `\nat' of natural numbers, 
each data item of the sort `\nat' is to be denoted
by some constructor term of the sort `\nat' (no junk),
and two different constructor terms of the sort `\nat' describe two
different data objects (no confusion). 
Note that the latter is special for constructor
terms: \Eg, for a non-constructor function symbol `\plussymbol', the terms
\bigmath{\plusppnoparentheses{\spp 0}0,}
\bigmath{\plusppnoparentheses 0{\spp 0},} and \bigmath{\spp 0}
may well denote the same data object, but only the last one
is a constructor term.

Since we are strongly convinced that
the notion of a ``constructor function symbol''
must be based on the signature only
(and not on the axioms of a specification),
this leads us to the following refinement of the notion of 
a many-sorted signature.

\begin{sloppy}
\math{\sig'}
is a\emph{subsignature}
of \math{\sig} \udiff\
\math{\sig'} and \sig\
are many-sorted signatures and,
for \linebreak
\bigmath{
  \trip{\sigsorts'}{\sigfunsym'}{\sigarity'}
  :=
  \sig'
}
and
\bigmath{
  \trip{\sigsorts}{\sigfunsym}{\sigarity}
  :=
  \sig
,}
we have
\bigmath{
  \sigsorts'\tightsubseteq\sigsorts
,}
\bigmath{
  \sigfunsym'\tightsubseteq\sigfunsym
,}
and
\bigmath{
  \sigarity'\tightsubseteq\sigarity
.}
The\emph{\math{\sig'}-reduct} of a \sig-algebra \algebra\
consists only of the universes for the sorts of
\math{\sigsorts'} and of the functions for the symbols in 
\math{\sigfunsym'}.
Formally, when a \sig-algebra \algebra\ is seen as a total function 
with domain \math{\sigsorts\tightuplus\sigfunsym},\footnote
{We use `\tightuplus' for the disjoint union of classes.}
the \math{\sig'}-reduct can be seen as the restriction of \algebra\ 
to the domain \math{\sigsorts'\tightuplus\sigfunsym'}, 
which we generally denote in the form 
\domres\algebra{\sigsorts'\tightuplus\sigfunsym'}.
For a subset \math{\consfunsym\subseteq\sigfunsym} 
we denote with \math{\sig^\consfunsym} the subsignature
\trip\sigsorts\consfunsym{\domres\sigarity\consfunsym} of \sig.\footnote
{Note that \domres\alpha\consfunsym\ 
 denotes the restriction of the function \math\alpha\ 
 to the domain \consfunsym.}
If the function \balgebra\
that differs from the \sig-algebra \algebra\
only in that the universe of
each sort \math{s\in\sigsorts} contains only the values of the \consfunsym
-terms of the sort \nolinebreak\math s 
under the evaluation function of \algebra,
is a \sig-algebra again, then we call \balgebra\
the\emph{\math{\consfunsym}-generated subalgebra} of \algebra.
We call \consfunsym\ a set of\emph{constructors for \sig} \udiff\ 
\bigmath{\consfunsym\subseteq\sigfunsym} 
and the signature \math{\sig^\consfunsym} is\emph{sensible} 
(or ``inhabited''), \ie, for each \math{s\in\sigsorts},
there is at least one constructor ground term of sort \math s.
\begin{definition}[Data Reduct]\\
If \consfunsym\ is a set of constructors for \sig,
then, for each \sig-algebra \algebra,
the \consfunsym-generated subalgebra of the 
\math{\sig^\consfunsym}-reduct of \algebra\ 
is a \math{\sig^\consfunsym}-algebra, which is called 
the\emph{\consfunsym-data reduct of \algebra.}
\end{definition}
A\emph{constructor-based specification}
\bigmath{{\rm spec} = \trip\sig\consfunsym\Ax} 
is composed of a 
set of constructors \nolinebreak\consfunsym\ of the signature \sig\
and of a set \Ax\ of axioms (over \sig).
\begin{definition}[Data Model]\label{definition data model}
\\
Let \bigmath{{\rm spec} = \trip\sig\consfunsym\Ax} 
be a constructor-based specification.
\\
\algebra\
is a\emph{data model} of `{\rm spec}' \udiff\ 
\Ax\ is valid in the \sig-algebra \nolinebreak\algebra\
and the \consfunsym-data reduct of \nolinebreak\algebra\ is 
isomorphic to the term algebra over \math{\sig^\consfunsym}.\footnote
{\Ie\
 isomorphic to the initial \math{\sig^\consfunsym}-algebra.}
\end{definition}
Note that the latter is just a formal way to express the catchword
``no confusion'' from above.
The catchword ``no junk'' can formally be realized by
variables ranging only over the constructor ground terms 
or the \consfunsym-data reduct of \algebra.
For technical details \cf\ \cite{kwspec}.
\end{sloppy}

Let \math{\deffunsym:=\sigfunsym\tightsetminus\consfunsym} 
denote the set of\emph{non-constructor} (or\emph{defined}) function symbols.
Note that by \defiref{definition data model}, the data reduct of 
data models of a consistent specification `{\rm spec}' 
is uniquely defined (up to isomorphism)
as the constructor ground term algebra.
Data models for `{\rm spec}' may differ, however, in the way
partially specified functions from \deffunsym\
behave in the unspecified cases.
\Eg, 
suppose that the operator `\minussymbol' is partially specified on `\nat' 
by the two equations 
\bigmath{\minusppnoparentheses x 0=x}
and
\bigmath{
  \minusppnoparentheses{\spp x}{\spp y}
  =
  \minusppnoparentheses x y
.}
In this case, data models may differ on the evaluation of
the term \minusppnoparentheses 0{\spp 0}, which may evaluate
to different values of the \consfunsym-data reduct or even to
different ``junk'' or ``error'' values.
Note that in this way we can model partial functions with total algebras.

This possibility to model partiality is also the reason why
we prefer characteristic functions (\ie\ 
functions of Boolean sort)
to predicates: the result of the application of a characteristic function 
can be true, false or possibility 
neither true nor false (undefined, unspecified).
With predicates we do not have the latter possibility.
\pagebreak

\yestop
\noindent
The constructor ground terms of the sorts of some subset 
\math{\sigsorts_{\rm P}\subseteq\sigsorts} will be used to 
describe the fixed unchanging parts of a product.
The constructor ground terms of the 
remaining sorts in \math{\sigsorts\tightsetminus\sigsorts_{\rm P}}
statically describe the dynamic states of the product
without its dynamic behavior. 
The dynamic functions from \deffunsym\
will change the static description of the product \wrt\ 
the constructor ground terms of these sorts.
As we do not have final algebra domains
or state sorts in our application by now, 
we have not treated these subjects explicitly here.

It is useful to further classify the function symbols from \deffunsym.
\Eg, functions that inspect a data item may be called ``selectors'' or
``observers'', functions that manipulate may be called ``mutators'' or
``editors'', \etc.
More important here is the classification of a function symbol
as belonging to the product of the design process;
contrary to functions for the design process itself,
auxiliary functions for the implementation, \etc.
Thus, let \math{{\Bbb P}\subseteq\deffunsym} be a set of\emph{product
function symbols}.
\quin\sig\consfunsym{\sigsorts_{\rm P}}{\Bbb P}\Ax\
is a\emph{product specification} \udiff\
\trip\sig\consfunsym\Ax\ is a constructor-based specification,
\math{\sigsorts_{\rm P}\subseteq\sigsorts} is non-empty and
\bigmath{{\Bbb P}\tightsubseteq\deffunsym,}
for 
\bigmath{
  \trip{\sigsorts}{\sigfunsym}{\sigarity}
  :=
  \sig
}
and
\bigmath{
  \deffunsym:=\sigfunsym\tightsetminus\consfunsym
.}
\begin{definition}[Product Model]\label{defprodmod}
\\
Let \math{\sig=\trip\sigsorts\sigfunsym\sigarity} be a many-sorted signature.
\\
Let \bigmath{{\rm spec}=\quin\sig\consfunsym{\sigsorts_{\rm P}}{\Bbb P}\Ax} 
be a product specification.
\\
Let \math C be the set of those function symbols \math{c\in\consfunsym} whose
argument and result sorts in \app\sigarity c do all belong to 
\math{\sigsorts_{\rm P}}.
\\
A\emph{structural product model} of `{\rm spec}' 
is the \trip{\sigsorts_{\rm P}}C{\domres\sigarity C}-reduct of the
\consfunsym-data reduct of
a data model of \trip\sig\consfunsym\Ax.
\\
A\emph{behavioral product model} of `{\rm spec}' is the 
\math{\sig^{\consfunsym\tightcup{\Bbb P}}}-reduct of 
a data model of \trip\sig\consfunsym\Ax.
\end{definition}
Note that by this definition, a structural product model 
of a consistent specification is uniquely defined (up to isomorphism).
Behavioral product models, however, may differ in the way
partially specified functions from \math{\Bbb P} behave 
in the unspecified cases.

The present situation of our application is not very complicated
because at first only the structural product model is of interest.
Moreover, since \bigmath{\sigsorts_{\rm P}\tightequal\sigsorts,}
the \trip{\sigsorts_{\rm P}}C{\domres\sigarity C}-reduct of the
\consfunsym-data reduct is the \consfunsym-data reduct itself.
Therefore, the whole universe of discourse, 
namely all possible descriptions of products,
can and will be represented by constructor ground terms.
To simplify the description of the structural product model we use some predefined
data types, like 
`\tightemph\natsort ' and
`\tightemph\boolsort', some of them generic, like 
`\tightemph\setsort ',
`\tightemph\mapsort ',
`\tightemph\listsort', and 
`\tightemph\treesort'. For the 
understanding of the product model, it suffices to assume that 
these data types do what their mathematical counterparts do. For 
a deeper understanding a detailed description can be found in \cite{pad99}. 
For the presentation of our specification we 
use the fairly intuitively readable style from \cite{pad99}.\footnote
{This style is constantly improved. Thus, there can be little differences in 
 the notation, which should not disturb the understanding of the 
 presented specifications.}
The only further remark that may be necessary here is the way
the structured specification is meant to be put together:
The union of two specifications is the element-wise non-disjoint union
of sort symbols, function symbols, 
arity functions, constructors symbols, and axioms. 
When parameters of a specification are bound to some actual name of 
a specification, we take the union of both specifications and
replace the parameter with the actual name everywhere.
Although this approach is not perfect,\footnote
{\Eg, the approach is error-prone and does not provide any
proper modularization, \ie\ the specification can only be checked or
properly understood as a whole.} 
we have chosen it for its simplicity, power and conciseness.\vfill\pagebreak

\subsection{The Object under Consideration: Hyperdocuments\label{prodmod}}
In the domain of hyperdocuments there are three fundamental different kinds
of product models (\cf\ \cite[\p\,221\,\ff]{low99}):\emph
{Programming language based},\emph
{information-centered} and\emph
{screen-based} models. 
The programming language based approach, which applies any general
purpose programming language starting from scratch, was used in former 
days due to the lack of any other sophisticated models, and has nearly 
no importance in the presence.

\begin{sloppypar}
For a long time the\emph{information-centered model} has dominated.   
The most popular model for hyperdocuments, 
the ``Dexter Hypertext Reference Model'' \cite{hal90}, 
is information-centered.
Dexter or one of its modifications, \eg\ \cite{gro94} or \cite{oss95b},
describe the structure of a hyperdocument, divided into its logical structure,
its linkage, and its style. A hyperdocument can import components from a 
 ``within-component layer'' via an anchor mechanism and specify how the 
document should be presented by a  ``presentation specification''. 
\end{sloppypar}

Similar ideas are presented in an object-oriented style in the 
 ``Tower Model'', \cf\ \cite{bra92}.
Additionally a hierarchy is added. It is described that
components could include other components. But there are not mentioned any 
restrictions how to compose hyperdocuments. So you can produce a lot of 
components not used in any actual hypermedia system. 
In both models 
there is no possibility to describe strategies how to navigate 
through a set of hyperdocuments. But this is a design goal of increasing 
importance in the rapidly growing world of hypermedia. 
The Dexter-based reference model
for adaptive hypermedia\emph{(AHAM)} 
(\cf\ \cite{bra98}) describes first steps toward this direction.

Even the wide-spread Hypertext Markup Language (\tightemph{HTML}) has 
obviously its roots in the information-centered paradigm, 
even though many designers use it in another way,
namely as a screen-based design language. 
``Screen-based'' means that the focus is not the logical
structure of the document, enriched with some display attributes, 
but the display of the document itself. 

With the upcoming of the \WWW\ 
and the WYSIWYG-editors the\emph{screen-based model} became 
more important in hyperdocument design, 
because sometimes it is easier to think in terms of
the produced view on the screen. 

A simple and common characterization of our object under
consideration is:

\begin{definition}[Informal Description of a Hyperdocument\label{definfhd}]\\
A\emph{hyperdocument} is a {\em basis document}, sometimes called {\em lineardocument}, 
consisting of a fixed set of basic contents, organized 
according to a {\em media structure}, 
enriched with a pointer concept, called {\em anchors}, to access a specific 
content inside the document, and a reference concept, 
called\emph{hyperlinks}, to access another document by 
its {\em address}. If the only medium in a hyperdocument is text, 
then we speak of a\emph{hypertext document}, 
or else of a\emph{hypermedia document}.\\
Moreover, device independence is often formulated
as a hypermedia requirement. This is only possible if 
you disjoin the structural description and the {\em presentation 
attributes}, as \eg\ in HTML or \TeX.
\vfill\pagebreak
\end{definition}
In the following sections, we will examine how the five crucial
elements of a hyperdocument,
\noindent
\begin{itemize}
\item the basis document (\ref{basisdomain}), 
\item the set of anchors (\ref{anchordomain}), 
\item the set of links (\ref{linkdomain}),
\item the presentation attributes\footnote{Because of the fact that 
presentation attributes are meaningful only in connection 
with a screen-based-model, we leave them undefined at the
moment. They will be added later.} and 
\item the addresses (\ref{addressdomain})  
\end{itemize}
can be specified.

\subsection{Basis Documents\label{basisdomain}}

According to the Dexter Model the structure of the basis is not known. 
It is only assumed, that each basis element has a fixed set of properties
(which can be observed by some special observer functions,
which are not part of the product model) and a particular structure, which 
can be accessed by the anchor-mechanism via a {\em location}. 
Accordingly we model basis documents as a parameter.

\noindent\fbox{\parbox{\textwidthminustwomm}{
\begin{definition}[Parameter ``Basis Document'']\label{defbasisp}\\\\
\BASISP[\basisparamsort,\locationparamsort]~=
\sortsect{
  \basisparamsort
\\\locationparamsort
}\end{definition}}}

\noindent
Note that in the boxes like the one above we do not present the full
specification (\cf\ \sectref{specp}) but only an essential part of it that 
should be easy to understand.

\subsection{Anchors\label{anchordomain}}

Originally a hyperdocument used to have no layout at all. It was seen only as
an arbitrary collection of atomic basis elements. 
The\emph{anchor} was the only
possibility to get access to one of these atoms. It had a name and
a method which could be interpreted by the underlying database.
When hypertext evolved, more complex construction mechanisms came up
and the need to control the layout became more important. The anchor-method
depended no longer only on the data base but also on the 
document structure. This method to access an element at a given 
position is a bit confusingly named {\em location}. We
adopt that name, because it is used in most hypermedia models.

In contrast to Dexter, our anchors are enriched with an anchor-type.
So you may not only mark a special element, but you also mark it as 
a possible start-point (\tightemph{source}) or end-point 
(\tightemph{target}) of a link or both (\tightemph{label}). 
Note that in our specification the
anchor-types are part of anchors and therefore\emph{local} to the
hyperdocuments. In Dexter this feature
is included into the\emph{global} specifier-mechanism of hyperlinks, however 
(\cf\ \sectref{linkdomain}).
In pre-WWW times, where hypertext was usually a non-distributed system,
this made no difference. But in a distributed system like the \WWW\
it becomes important that the anchor types can be found without searching
the whole \WWW\ and must therefore be stored local to the document
they are related to.

Considering all these facts and adding the attributes, as discussed
previously, we come to the following specification for anchors: 

\noindent\fbox{\parbox{\textwidthminustwomm}{
\begin{definition}[Structural Product Model ``Anchor'']\label{defanchor}\\\\
\ANCHORCLASS[\locationparamsort]~=~\BASISP[\basisparamsort,\locationparamsort]~\andspec~{\ATTj \ANCHORCLASS}~\thenspec
\vissortsect{
  \anchortypesort
\\\identsorts\anchorsort {\anchorsort(\locationparamsort)}
}
\conssect{
  \sourcepp,\targetpp,\labelpp:~\anchortypesort
\\\fundecl{\mkanchor}{\locationparamsort \times \anchortypesort \times \attjsort\anchorsort}\anchorsort
}\end{definition}}}

\subsection{Hyperlinks}\label{linkdomain}

A {\em hyperlink} (or\emph{link} for short) is a reference from a fixed
set of contents (\tightemph{source}) to a fixed set of contents 
(\tightemph{target}).
Each of these sets of contents are described by a set of {\em specifiers}.
Our model differs from the Dexter Model insofar as no links to
links are possible. But our view is compatible to most other hypermedia
models. A specifier consists of a global 
address of sort `\urisort' and a local 
name of sort `\anchornamesort'.
`\urisort' is the abbreviation for ``Unified Resource Identifier''\footnote
{The well-known URLs in the {\WWW} are a subset of URIs.} 
\cite{ber98}.
The anchor-name is to be mapped to an anchor of the hyperdocument under 
the global address. This mapping is not global but part of the hyperdocument.
In the Dexter Model, 
specifiers have also a direction. We split this direction 
into the {\anchortypesort} and the {\linktypesort}. Hence we get
uni- and bi-directional links. 

Moreover links are classified according
their intended behavior. This idea goes back to \cite{eng84}, 
where {\em jump-} and {\em include-}links were introduced. 
Often the term ``jump-link'' is used
synonymous with link at all. It denotes that kind of link where the system
is waiting for a user action (\eg\ a mouse-click) 
and then the old source-document
is replaced by the new target-document. 
The term ``include-link'' denotes a class of links 
which are to be automatically evaluated and presented inside a previously
defined location.
These ``traditional'' kinds of links do not suffice since systems work with
multiple windows. A third kind of link is necessary, namely one that
can open new windows to present the target-document and leave the 
source-document untouched in its old place. 

This kind of presentational 
behavior is represented in the\emph{show-type}, as we will call
it according to \cite{w3c98e}. 
Links of show-type `\embedpp' embed their target into the context of 
their source. Links of show-type `\VMreplacepp' replace the hyperdocument 
of their source with the hyperdocument of their target. Finally, links 
of show-type  `\VMnewpp' open a new window with the document of their target.

The second distinction is whether a user interaction is required or not.
This is represented by the\emph{actuate-type}, as we will call it
according to \cite{w3c98e}. 
Links of actuate-type `\userpp' are followed upon user interaction.
Links of actuate-type `\autopp' are followed automatically.

If we combine all the named possibilities we get twelve different types
of links. But, what sense makes \eg\ a bi-directional link of show-type
`\embedpp'? Or a bi-directional link of actuate-type `\autopp'?
We think that the only meaningful bi-directional links are of show-type 
`\VMreplacepp' and of actuate-type `\userpp'.
Therefore, 
uni-directional links (`\mkunipp * *') are modeled with two parameters
(show-type, actuate-type), but no arguments are given to the bi-directional
links (`\bipp').

The previously mentioned jump-link has the type 
\mkunipp{\VMreplacepp}{\userpp} and the include-link \mkunipp{\embedpp}{\autopp}.

\noindent\fbox{\parbox{\textwidthminustwomm}{
\begin{definition}[Structural Product Model ``Links'']\label{deflink}\\\\
\LINKCLASS~=~\ANCHORNAME~\andspec~{\URI}~\andspec~{\ATTj LINK}~\andspec~\SETof{\renamesort\entrysort {\specifiersort}}~\thenspec
\vissortsect{
  \linktypesort
\\\showsort
\\\actuatesort
\\\specifiersort
\\\linksort
}
\\\conssect{
  \embedpp, \VMreplacepp, \VMnewpp:~\showsort
\\\userpp, \autopp:~\actuatesort
\\\fundecl{\mkuni}{\showsort \times \actuatesort}\linktypesort
\\\bipp:~\linktypesort
\\\fundecl{\mkspecifier}{\urisort \times \anchornamesort}\specifiersort
\\\fundecl{\mklink}{\setof\specifiersort \times \setof\specifiersort \times \linktypesort \times \attjsort\linksort}\linksort
}\end{definition}}}

\noindent
The generic abstract data type
`\set' in \defiref{deflink} is assumed to be predefined,
\cf\ \p\,\pageref{signature of SET} for its signature.

\vfill\pagebreak

\subsection{Addresses\label{addressdomain}}

In order to be referenced, each hyperdocument must have an address. 
In general, this address space
is described by the already described sort `\urisort'.
But we will allow to define special 
address subspaces for local addresses
where the type of a hyperdocument can be inferred 
from the type of its address. 
Thus, we have a second parameter. 

\noindent\fbox{\parbox{\textwidthminustwomm}{
\begin{definition}[Parameter ``Addresses'']\label{defaddr}\\\\
\ADDRP[\addrparamsort]~=
\sortsect{
  \addrparamsort
}\end{definition}}}

\vfill

\subsection{Hyperdocuments}\label{hddomain}

We have now modeled all parts of our product, but as often, 
the product is more than the sum of its parts. 
It is not very convenient to access specific
parts of the basis via a possibly cryptic location-description. That is the
reason why hyperlinks deal only with anchor-names, instead of their values. 
Therefore each anchor, if it is used in a document, must be combined with 
a name. We model this by using a function, 
thereby ensuring that no anchor name 
can be used twice inside the same document. 
We get a product model for a class of \hd s that vary in the
underlying documents and the address space. 
These open parameters 
will be instantiated in the following section. 

\noindent\fbox{\parbox{\textwidthminustwomm}{
\begin{definition}[Structural Product Model ``Hyperdocuments'']\label{defhd}\\
\begin{tabbing}
\HDCLASS[\basisparamsort,\locationparamsort,\addrparamsort]~=~ \= \HDCLASS~ \= \hspace*{-6mm} \=\kill
\HDCLASS[\basisparamsort,\locationparamsort,\addrparamsort]~= \>  
   \BASISP[\basisparamsort,\locationparamsort]~\andspec\\
\> \ADDRP[\addrparamsort]~\andspec\\
\> \ANCHORCLASS[\locationparamsort]~\andspec\\
\> \LINKCLASS~\andspec\\
\> {\ATTj \HDCLASS}~\andspec\\
\> \MAPof{\renamesort\domainsort {\anchornamesort},\renamesort\rangesort {\anchorsort}}~\andspec\\
\> \SETof{\renamesort\entrysort {\linksort}}
\end{tabbing}
\thenspec
\vissortsect{
  \identsorts\hdsort {\mident{\hdsort(\basisparamsort,\locationparamsort,\addrparamsort)}}
}
\conssect{
  \fundecl{\mkhd}{\basisparamsort \times \mapof{\anchornamesort}{\anchorsort} \times \setof\linksort \times \attjsort\hdsort \times \addrparamsort}\hdsort
}\end{definition}}}

\vfill\pagebreak

\subsection{The Hierarchy of Hyperdocuments}

Most hypermedia models end here with the definition of \hd s. 
Some of these models give no further information about the 
structuring of \hd s at all, others define new kinds of objects, \eg\ 
views. We suggest another approach, based on the classical
organization of texts. They are structured by a
hierarchy of at least three levels, shown in the left column 
of Table\,\ref{tab1}.

\begin{table}[h t]
\begin{center}
\begin{tabular}{|l|l|l}
  \cline{1-2}
    {\it Linear Text} 
   &\Hd
  \\\cline{1-2}&\\[-4.5mm]\cline{1-2}
   {\it Book} 
   &\Site
   &(Section \ref{hdonemod})
  \\\cline{1-2}
   {\it Chapter}\/\footnotemark
   &\Fsd
   &(Section \ref{hdtwomod})
  \\\cline{1-2}
   {\it Page}
   &\Hmd 
   &(Section \ref{hdthreemod})
  \\\cline{1-2}
  \end{tabular}
\caption{The Levels of a Document\label{tab1}}
\end{center}
\end{table}\footnotetext{\tightemph{Wall news sheet} may be intuitionally 
closer to ``frameset document'' because it describes a multi-dimensional
combination of pages.}

The only basic element of a linear text is the character.
Together with the media-structures like paragraphs, tables or lists, 
they build the structured basis for documents. Arranging these 
structured elements sequentially leads to a page. 
Now you have the possibility to combine pages
into a document of a higher level.
We believe that this hierarchy is a good strategy to organize
{\hd s} as well, because these levels can also be found, 
when you examine the most popular application for
hyperdocuments, the \WWW, and the wide spread Hypertext Markup Language 
(\cite{w3c98a}) or some of its relatives out of the 
SGML-family\footnote{SGML is the Structured Generalized Markup 
Language (ISO-Norm 8779)}. The right column of Table\,\ref{tab1}
shows the hypermedial
counterpart in terms of the most prominent hypertext application,
the \WWW.

Thus, we will define three typical levels of hierarchy for a hyperdocument. 
These levels belong to the ``storage layer'' in the Dexter Model,
\cf\ Table\vref{tablecd}. 
The {\mo s} belong to the ``within-component layer'' 
of the Dexter-Model.  This is not the focus of our work and it will 
not be viewed in detail.

\begin{table}[h]
\begin{center}
\begin{tabular}{|l|l|}
  \hline
   \bf Dexter Model
  &\bf Our Product Model
  \\\hline\hline Run-time Layer
  &{ ---}
\\\hline
  {\it Presentation Specifications}
& \it Attributes
\\\hline
  { Storage Layer}
& {\begin{tabular}{|l|}
  \multicolumn{1}{c}{}\\[-2mm]
  \hline
    \Site
  \\\hline
    \Fsd
  \\\hline
    \Hmd
  \\\hline\multicolumn{1}{c}{}\\[-2mm]
  \end{tabular}
}
\\\hline
  {\it Anchoring}
& \it Anchor
\\\hline
  { Within-Component Layer}
& {\begin{tabular}{|l|}
  \multicolumn{1}{c}{}\\[-2mm]
  \hline
   \Mo
\\\hline\multicolumn{1}{c}{}\\[-2mm]
\end{tabular}
}
\\\hline
\end{tabular}
\end{center}
\caption{Comparison with the Dexter Model (Interfaces in italics)\label{tablecd}}
\end{table}

\vfill\pagebreak

\subsubsection{Media-Objects}

\begin{sloppypar}
Media-objects are not {\hd s}. They only provide the interface to the
Within-Component-Layer in the Dexter Model.  
As hypermedia is an open approach, there 
are infinitely many different types of \mo s in principle.
\end{sloppypar}

Our interface to media-objects is quite simple because we are not
interested in modeling their internal behavior. The only thing we require
is that they have some unified resource identifier of sort `\urisort'
and a set of anchor identifiers to which links may refer.
Thus, a media-object basically introduces a legal set of specifiers
referring to it.

\noindent\fbox{\parbox{\textwidthminustwomm}{
\begin{definition}[Structural Product Model ``Media-Objects'']
\label{defmo}\\\\
\MO~=~\URI~\andspec~\ANCHORNAME~\andspec
~\SET[\renamesort\entrysort\anchornamesort]~\thenspec
\vissortsect{
  \mosort=\apptotuple\mosort{\pair\urisort\anchornamesort}
}
\conssect{\fundecl{\mkmo}{\urisort\times\setof\anchornamesort}\mosort
}\end{definition}}}

\vfill\vfill

\subsubsection{Pages and Hypermedia-Documents}\label{hdonemod}

Pages are at the lowest level in the hierarchy. 
As mentioned before the basic contents,
represented by the media-objects, is hierarchically structured. 
Some models (\cf\ \eg\ \cite{dob96a}) 
introduce a sub-document relation for this purpose, 
which only describes which document is part of another. 
The way in that they are related is left to the
presentation attributes. 
This strategy is adequate to examine the navigational structure of a document,
but it is not sufficient to describe ``real-world`` {\hd s}.
We believe that presentation attributes must be reserved for simple lay-out
purposes only, 
and that a change of presentation attributes must not change the document
in a fundamental way. 
\Eg, if you re-arrange a table into a linear
list, you change the information. 
Of course, the distinction between structural 
elements and lay-out attributes is not sharp in general. 
To avoid a discussion about
this topic here, we pragmatically follow the HTML-definitions. 
Note that our product model
allows both, 
a description solely with the predefined 
structural elements or solely with presentation
attributes of an unstructured text. 
We think that our proposed mix of both is the best way, but the model
does not enforce this. 

Pages are simple linear texts, 
with a fixed set of logical structuring elements,
such as paragraphs, lists or tables. Of course, one can imagine more functions
than we define here, but we tried to model the minimal necessary set of functions.

Besides the basic elements, we introduce a set of level-dependent 
symbols, which are simply characters on the first level. 
We differentiate them for practical reasons. 
Generally, symbols differ from basic elements
in that they do not have an individual address, but are immediately handled 
by the browser.
\vfill\pagebreak

\noindent\fbox{\parbox{\textwidthminustwomm}{
\begin{definition}[Structural Product Model ``Page'']\label{defldone}\\\\
\LDONE~=~\MO~\andspec~\SYMBOL{\LDONE}~\andspec~{\ATTj \LDONE}~\andspec\\
\phantom{\LDONE~=~}\TREE[\renamesort\entrysort \structitypesort\ldonesort]~\andspec\\
\phantom{\LDONE~=~}\LIST[\renamesort\entrysort \ldonesort]~\andspec~\LIST[\renamesort\entrysort \natsort]~\thenspec
\vissortsect{ 
  \ldonesort
\\\structitypesort\ldonesort
\\\identsorts{\locisort{\ldonesort}} {\mident{\listsort(\natsort)}}
}
\conssect{
  \basic,\symb,\emptypage,\hdlist,\hdtable,\hdtableline,\hdheadline,\hdpage,\hdtext,\\
  \hspace{5pt}\hdlinebreak,\hdfootnote,\hdparagraph,\hdcopyright~:~\structitypesort\ldonesort
\\\mtpage~:~\ldonesort
\\\fundecl{\imphdlow}{\mosort}\ldonesort
\\\fundecl{\impsymbol}{\symbolisort{\ldonesort}}\ldonesort
\\\fundecl{\mkld}{{\structitypesort\ldonesort} \times \listof{\ldonesort} \times \attjsort\ldonesort}\ldonesort
}
\end{definition}
}}

\yestop
\noindent
To construct a {\hd} of our first level we now only have to 
combine our product models
for page and the address space and instantiate the parameters 
`\basisparamsort', `\locationparamsort', and `\addrparamsort'.

\noindent\fbox{\parbox{\textwidthminustwomm}{
\begin{definition}[Structural Product Model ``Hypermedia-Documents'']
\label{defhdone}\\
\begin{tabbing}
\HDONE~=~ \= \= \HDCLASS \= \kill
\HDONE~=~ \> \LDONE~\andspec~\ADDR{\HDONE}~\andspec~\\
\>\> \HDCLASS [{\renamesort\basisparamsort {\LDONE}.\ldonesort},\\
\>\>\> {~\renamesort\locationparamsort {\LDONE}.\locisort{\ldonesort}},\\
\>\>\> {~\renamesort\addrparamsort {\ADDR{\HDONE}.\addrisort{\hdonesort}}}]
~\thenspec
\vissortsect{ 
  \hdonesort=
  \apptotuple\hdsort{\trip
    {\LDONE.\ldonesort}
    {\LDONE.\locisort\ldonesort}
    {\ADDR\HDONE.\addrisort\hdonesort}}}
\end{tabbing}\end{definition}}}
\vfill\pagebreak

\subsubsection{Chapters and Frameset Documents}\label{hdtwomod}

\yestop\noindent\tightemph
{The following specifications are essentially
incomplete and have to be completed in the
future!!!}

\yestop\noindent
At the second level, our basic elements are the structured {\hd s} (\defiref{defhdone}).
From this point of view, 
the name ``lineardocument'', mentioned previously, is not 
quite right. 
Though it is organized without links on the
discussed level (and hence ``linear''), 
its basic documents might obviously be 
hyperdocuments already.
The symbols at this level are geometrical forms, such as lines, 
rectangles or bars.

\noindent\fbox{\parbox{\textwidthminustwomm}{
\begin{definition}[Structural Product Model ``Chapter'']\label{defldtwo}\\\\
\LDTWO~=~\HDONE~\andspec~\SYMBOL{\LDTWO}~\andspec~{\ATTj \LDTWO}~\andspec\\
\phantom{\LDTWO~=~}\TREE[\renamesort\entrysort \structitypesort\ldtwosort]~\andspec\\
\phantom{\LDTWO~=~}\LIST[\renamesort\entrysort \ldtwosort]~\andspec~\LIST[\renamesort\entrysort \natsort]~\thenspec
\vissortsect{ 
  \ldtwosort
\\\structitypesort\ldtwosort
\\\identsorts{\locisort{\hdtwosort}} {\mident{\listsort(\natsort)}}
}
\conssect{
  \hdhframeset,\hdvframeset,\hdaframeset~:~\structitypesort\hdtwosort
}\end{definition}}}

\yestop\noindent
Analogous to the previous section, we must instantiate the parameters.

\noindent\fbox{\parbox{\textwidthminustwomm}{
\begin{definition}[Structural Product Model ``Frameset Document'']
\label{defhdtwo}
\begin{tabbing}
\HDTWO~=~ \= \= \HDCLASS \= \kill
\HDTWO~=~ \> \LDTWO~\andspec~\ADDR{\HDTWO}~\andspec\\
\>\> \HDCLASS [{\renamesort\basisparamsort {\LDTWO}.\ldtwosort},\\
\>\>\> {~\renamesort\locationparamsort {\LDTWO}.\locisort{\ldtwosort}},\\
\>\>\> {~\renamesort\addrparamsort {\ADDR\HDTWO.\addrisort{\hdtwosort}}}]
\end{tabbing}
\thenspec
\vissortsect{ 
  \hdtwosort=
  \apptotuple\hdsort{\trip
    {\LDTWO.\ldtwosort}
    {\LDTWO.\locisort\ldtwosort}
    {\ADDR\HDTWO.\addrisort\hdtwosort}}}
\end{definition}}}
\vfill\pagebreak

\subsubsection{Books and Sites}\label{hdthreemod}

\yestop\noindent\tightemph
{The following specifications are essentially
incomplete and have to be completed in the
future!!!}

\yestop\noindent
The third level is the aggregation of chapters to a book. A book consists
of ``hyperchapters''.

\noindent\fbox{\parbox{\textwidthminustwomm}{
\begin{definition}[Structural Product Model ``Book'']\label{defldthree}\\\\
\LDTHREE~=~\HDTWO~\andspec~\SYMBOL{\LDTHREE}~\andspec~{\ATTj \LDTHREE}~\andspec\\
\phantom{\LDTHREE~=~}\TREE[\renamesort\entrysort \structitypesort\ldthreesort]~\andspec\\
\phantom{\LDTHREE~=~}\LIST[\renamesort\entrysort \ldthreesort]~\andspec~\LIST[\renamesort\entrysort \natsort]~\thenspec
\vissortsect{ 
  \ldthreesort
\\\structitypesort\ldthreesort
\\\identsorts{\locisort{\ldthreesort}} {\mident{\listsort(\natsort)}}
}
\conssect{
  \hdsitemap~:~\structitypesort\ldthreesort
}\end{definition}}}

\noindent\fbox{\parbox{\textwidthminustwomm}{
\begin{definition}[Structural Product Model ``Site'']\label{defhdthree}
\begin{tabbing}
\HDTHREE~=~ \= \= \HDCLASS \= \kill
\HDTHREE~=~ \> \LDTHREE~\andspec~\ADDR{\HDTHREE}~\andspec\\
\>\> \HDCLASS[{\renamesort\basisparamsort {\LDTHREE}.\ldthreesort},\\
\>\>\> {~\renamesort\locationparamsort {\LDTHREE}.\locisort{\ldthreesort}},\\
\>\>\> {~\renamesort\addrparamsort{\ADDR\HDTHREE.\addrisort{\hdthreesort}}}]
\end{tabbing}
\thenspec
\vissortsect{ 
  \hdthreesort=
  \apptotuple\hdsort{\trip
    {\LDTHREE.\ldthreesort}
    {\LDTHREE.\locisort\ldthreesort}
    {\ADDR\HDTHREE.\addrisort\hdthreesort}}}
\end{definition}}}



\section{Extending the Product Model}\label{extdatatype}

In \sectref{datatype} we introduced an algebraic Dexter-based product model 
for hyperdocuments. We now extend this model with observer
and editing functions to an algebraic model for {\em hyperdocument systems}. 
By ``hyperdocument system'' we mean, as suggested  \eg\ by 
\cite{low99}, functions of tools used by a developer to create and modify a
hyperdocument. 
{\em Observer functions} supply information about the objects, \eg\ 
which elements a document contain. 
{\em Editing functions} can modify a concrete object, 
but of course not the domain. 
The remaining functions are merely {\em auxiliary functions}. 
They are not discussed in detail, but documented in the appendix.

In the constructor-based algebraic approach the set of functions is divided
into a set of {\em constructors} (\cf\ \sectref{algapproach}) and a set of 
{\em non-constructors} or {\em defined functions}. 
Defined Functions are defined via axioms on the basis of the constructors. 
Observer functions and editing functions are both represented by defined functions.

In our domain we have parameter specifications (\tightemph{document}),
object-classes (\tightemph{anchor} and\emph{hyperdocument}), and 
concrete objects ({\em link}, {\em page}, {\em hypermedia document}, 
{\em chapter}, {\em frame}, {\em book}, and\emph{site}). 
For each of these
we will explain at first the observer functions (\sectref{obsfunct})
and then the editing functions (\sectref{edfunct}). 

\subsection{Observer Functions}\label{obsfunct}

Objects are represented by tuples, build up with the help of the 
constructors. Observer functions are characterized by their ability to 
extract information out of these tuples. 
Historically they are sometimes called {\em destructors}, because
they can deconstruct objects. 
As the term ``destructor'' has already been used with so many connotations
and it is not clear whether it includes the Boolean functions,
we prefer the term ``observer functions'' here. 

\yestop\noindent
The\emph{observer functions}
include the following two special cases:
\begin{description}\item[Boolean functions] 
will be marked with a question mark `?' at the end of their names.
\item[Projections] extract exactly one
component of a composite object. 
Names of projections will be prefixed with `get\_'. 
\end{description}

\noindent
Observer functions must not be mixed up with 
{\em display functions}. 
Even though both help the user or
developer to observe an object, 
the latter transforms the logical description into
a `physical' and visible description, in our case a notation that
can be displayed by a user agent or browser. Display functions
are much more sophisticated in their algebraic representation
and a part of our ongoing work.

\subsubsection{Document}
The parameter specification
for {\em documents} has only one Boolean function, namely `{\embedlinkokS}'.
It tests whether an embed link can be positioned at a
given location in the document. All other observer and editing
functions belong to the documents on the corresponding level.
\pagebreak

\subsubsection{Page}
At the first level are the {\em pages}. 
A page is either an empty page,
some media object of lower level, 
some page symbol of the corresponding level, 
or a triple constructed by `\mkld' (\cf\ \sectref{hdonemod})
from a structure name (`\structitypesort\ldonesort'), 
a list of pages (`\listof\ldonesort'), 
and some attributes (`\attjsort\ldonesort').

\noindent\fbox{\parbox\textwidthminustwomm{
\begin{definition}[Observer Functions ``Page'']\label{obpage}
\deffunsect{
  \fundecl{\atomicS}{\ldonesort}\boolsort
\\\fundecl{\hasnthS}{\natsort\times\listof\ldonesort}\boolsort
\\\fundecl{\haslocationS}{\locisort{\ldonesort}\times\ldonesort}\boolsort
\\\fundecl{\includelinkokldoneS}{\locisort\ldonesort\times\ldonesort}\boolsort
\\\fundecl{\structSld}{\ldonesort}\treeof{\structitypesort\ldonesort}
\\\fundecl\pagesSld\ldonesort{\listof\ldonesort}
\\\fundecl{\attSld}{\ldonesort}\attjsort\ldonesort
\\\fundecl{\locate}{\locisort\ldonesort\times\ldonesort}\ldonesort
\\\fundecl\dimensionpagesymb{\ldonesort}\listof\natsort
}\end{definition}}}

A page is called\emph{atomic} (`\atomicS')
\uiff\ it is empty, a media object, or a symbol. 

`{\haslocationS}' is a partially defined boolean
function, which tests whether a location occurs 
in a page. 
The empty location means the whole page 
and therefore it exists in every page. 

`{\includelinkokldoneS}' returns
`\true' if a given location exists in the page and the document 
located there is an empty page. If the location does not exist,
it returns `\false'.

As a page is a nested
structure, the adequate result of the observer function 
`{\structSld}' is the tree of structures in the page under consideration. 

Similarly the result of `\pagesSld' is the list of all pages that a 
given page includes on top level. 

`{\attSld}' returns merely the top level
attributes of the page. 

`\locate' returns the sub-page located at a given position in a
given page. 

`\dimensionpagesymb' returns the list of natural numbers of the sizes of 
the page in all its dimensions. \Eg, a two dimensional 
table with \math m lines and a maximum of
\math n columns in one of these lines has a dimension of \math{(m,n)}.
This means that the smallest two dimensional cube around it will
have hight \math m and breadth \math n.
A three dimensional table with dimension \math{(m,n,p)} will fill
a cube of depth \math p.
If the objects are not atomic, the element-wise maximum of its dimensions
will be appended at the end of the dimension list of the table.
Generally speaking, a page object represented as an \mkld-node tree
of depth \nolinebreak\math d has the dimension \math{(n_1,\ldots,n_d)} where
\math{n_i} is the maximum number of children of a node at depth \math i.
\vfill\pagebreak

\subsubsection{Anchor}

An {\em anchor} is a triple constructed by `\mkanchor' 
(\cf\ \sectref{anchordomain}) 
from a location (`\locationparamsort'), a type (`\anchortypesort'), and some attributes (`\attjsort\anchorsort').

\noindent\fbox{\parbox\textwidthminustwomm{
\begin{definition}[Observer Functions ``Anchor'']\label{obanchor}
\deffunsect{
  \fundecl{\locSanchor}{\anchorsort}\locationparamsort
\\\fundecl{\typeSanchor}{\anchorsort}\anchortypesort
\\\fundecl{\attSanchor}{\anchorsort}\attjsort\anchorsort
\\\fundecl{\maxSanchor}{\anchorsort\times\anchorsort}\anchortypesort}
\end{definition}
}}

We need a projection for each component, called `\locSanchor', 
`\typeSanchor' and `\attSanchor'.

The last observer function, `\maxSanchor',
returns the supremal type according to 
`\math{\forall x\stopq x\leq\labelpp}'
 because an anchor of type `\labelpp' can serve both as
 source and as target, while 
 the types `\sourcepp' and `\targetpp' are incomparable.
\vfill

\subsubsection{Link}
A {\em (hyper) link} is a quadruple 
constructed by `\mklink' (\cf\ \sectref{linkdomain})
of a two sets of specifiers denoting the source and 
target (`\setof\specifiersort'),
a type (`\linktypesort'),
and some attributes (`\attjsort\linksort').
Specifiers again are pairs consisting of a global address 
(`\urisort') and local name (`\anchornamesort'). 

\noindent\fbox{\parbox\textwidthminustwomm{
\begin{definition}[Observer Functions ``Link'']\label{oblink}
\deffunsect{
  \fundecl{\addSlink}{\specifiersort}\urisort
\\\fundecl{\anchSlink}{\specifiersort}\anchornamesort
\\\fundecl{\sourceSlink}{\linksort}\setof\specifiersort
\\\fundecl{\targetSlink}{\linksort}\setof\specifiersort
\\\fundecl{\typeSlink}{\linksort}\linktypesort 
\\\fundecl{\attSlink}{\linksort}\attjsort\linksort
\\\fundecl{\specifierSlink}{\linksort}\setof\specifiersort}
\end{definition}}}

We need projections, `\addSlink', `\anchSlink' for the specifiers
and `\sourceSlink', `\targetSlink', `\typeSlink' and `\attSlink' for
the links. 

`\specifierSlink' returns the set of all specifiers in the source and the
target of a link.
\vfill\pagebreak

\subsubsection{Hyperdocument}
A {\em hyperdocument} is a quintuple constructed by `\mkhd'
(\cf\ \sectref{hddomain})
from a document (`\basisparamsort'),
a function mapping anchor names to anchors 
(`\mapof\anchornamesort\anchorsort'),
a set of links (`\setof\linksort'),
some attributes (`\attjsort\hdsort'),
and an address (`\addrparamsort').

\noindent\fbox{\parbox\textwidthminustwomm{
\begin{definition}[Observer Functions ``Hyperdocument'']\label{obhd}
\deffunsect{
  \fundecl{\ldShd}{\hdsort}\basisparamsort
\\\fundecl{\anchorsShd}{\hdsort}\mapof\anchornamesort\anchorsort
\\\fundecl{\linkShd}{\hdsort}\setof\linksort
\\\fundecl{\attShd}{\hdsort}\attjsort\hdsort
\\\fundecl{\addShd}{\hdsort}\addrparamsort
\\\fundecl{\anchornameShd}{\anchorsort\times\mapof\anchornamesort\anchorsort}\setof\anchornamesort
\\\fundecl{\anchorShd}{\anchornamesort\times\mapof\anchornamesort\anchorsort}\anchorsort}
\end{definition}}}

Of course we get five projections,
namely {$\ldShd$}, {\anchorsShd}, {\linkShd}, {\attShd} and {\addShd}.   
The first one extracts the (linear) document from the hyperdocument.
Because this function will be used very often, we
use the short notation `\ldShd' instead of the name `get\_document'.

`\anchorShd',  returns the anchor 
to a given anchor name. 

`\anchornameShd' returns the set of all 
anchor names referring to a given anchor.

\subsection{Editing Functions}\label{edfunct}

The\emph{editing functions} are the most interesting functions for the user. 
With the help of these functions a hyperdocument can be designed and modified.

\subsubsection{Page}\label{section page in editing functions}
We will start with the functions for working with {\em pages}.

\noindent\fbox{\parbox\textwidthminustwomm{
\begin{definition}[Editing Functions ``Page'']\label{edpage}
\deffunsect{
  \fundecl{\changestruct}{\structitypesort\ldonesort\times\ldonesort}\ldonesort
\\\fundecl{\insertatld}{\ldonesort\times\locisort\ldonesort\times\ldonesort}\ldonesort
\\\fundecl{\insertatlde}{\ldonesort\times \locisort{\ldonesort}\times\ldonesort}\ldonesort
\\\fundecl{\addattributeld}{\attjsort \ldonesort\times\ldonesort}\ldonesort
\\\fundecl{\delattributeld}{\attjsort \ldonesort\times\ldonesort}\ldonesort
\\\fundecl{\mktable}{\natsort\times\natsort}\ldonesort
\\\fundecl{\mklist}{\natsort}\ldonesort
}\end{definition}}}\pagebreak

`\changestruct' is a kind of converter function. The components of
the page are left untouched, but arranged in another structure.

Inserting one page into another at a special place is the most important
editing action a designer might need. We give two different functions
to do that: `\insertatld' and `\insertatlde'. Both replace a part
of an existing page, residing at a given location, with a new page.
A location is represented by a node position.
`\insertatld' moreover extends the page with sufficiently many child 
nodes, if this location does not yet exist.
The type of these child nodes may depend on the parent node.
\Eg, if the parent node is a table then the child nodes will be of
type table-line. If no special knowledge is given, the child nodes will
be simply of type empty page.

`\addattributeld' and `\delattributeld' add or remove attributes \resp.
Editing functions for attributes exist for every object and 
are not mentioned in the further sections anymore.

A special kind of editing functions are `\mklist' and `\mktable'.
They are syntactic sugar for very often used construction
mechanisms. `\mklist' produces a list with a given number of items, 
containing an empty page in every item.
`\mktable' produces a $m \times n$-table, containing an empty page
in every cell. 

\subsubsection{Anchor}
Anchor has only editing function that change the values of
the location (`\chlocation') or the type (`\chanchortype') \resp. 

\noindent\fbox{\parbox\textwidthminustwomm{
\begin{definition}[Editing Functions ``Anchor'']\label{edanchor}
\deffunsect{
  \fundecl{\chlocation}{\locationparamsort\times\anchorsort}\anchorsort
\\\fundecl{\chanchortype}{\anchortypesort\times\anchorsort}\anchorsort 
\\\fundecl{\addattributeanchor}{\attjsort \anchorsort\times\anchorsort}\anchorsort
\\\fundecl{\delattributeanchor}{\attjsort \anchorsort\times\anchorsort}\anchorsort
}\end{definition}}}

\subsubsection{Link}
According to the construction of links, we have editing
functions for specifiers and for links, which are very simple functions 
for changing the value of a component.

\noindent\fbox{\parbox\textwidthminustwomm{
\begin{definition}[Editing Functions ``Specifier'']\label{edspec}
\deffunsect{
  \fundecl{\chaddrspec}{\urisort\times\specifiersort}\specifiersort
\\\fundecl{\chname}{\anchornamesort\times\specifiersort}\specifiersort}
\end{definition}}}

\noindent\fbox{\parbox\textwidthminustwomm{
\begin{definition}[Editing Functions ``Link'']\label{edlink}
\deffunsect{
  \fundecl{\insertsource}{\setof\specifiersort\times\linksort}\linksort
\\\fundecl{\deletesource}{\setof\specifiersort\times\linksort}\linksort
\\\fundecl{\inserttarget}{\setof\specifiersort\times\linksort}\linksort
\\\fundecl{\deletetarget}{\setof\specifiersort\times\linksort}\linksort
\\\fundecl{\chlinktype}{\linktypesort\times\linksort}\linksort
\\\fundecl{\addattributelink}{{\attjsort \linksort}\times\linksort}\linksort
\\\fundecl{\delattributelink}{{\attjsort \linksort}\times\linksort}\linksort
}\end{definition}}}

\subsubsection{Hyperdocument}
At the first glimpse, things seem to be as easy with hyperdocuments 
as with the other objects. 
For the most functions, `\delanchor',
`\dellink', `\addattributehd', `\delattributehd' and `\chaddr',
this is true. 
But `\addanchor' and `\addlink' are much more 
sophisticated in their details.
 
\noindent\fbox{\parbox\textwidthminustwomm{
\begin{definition}[Editing Functions ``Hyperdocument'']\label{edhd}
\deffunsect{
  \fundecl{\addanchor}{\anchornamesort\times\anchorsort\times\hdsort}\hdsort
\\\fundecl{\delanchor}{\anchornamesort\times\hdsort}\hdsort
\\\fundecl{\addlink}{\linksort\times\hdsort}\hdsort
\\\fundecl{\dellink}{\linksort\times\hdsort}\hdsort
\\\fundecl{\addattributehd}{\attjsort \hdsort\times\hdsort}\hdsort
\\\fundecl{\delattributehd}{\attjsort \hdsort\times\hdsort}\hdsort
\\\fundecl{\chaddr}{\addrparamsort\times\hdsort} \hdsort
}\end{definition}}}

`\addanchor' produces a hyperdocument after a given anchor with 
given name
has been added to the anchors of the original hyperdocument, provided that
an anchor with this name does not exist before. 
If an anchor with this name does exist in the original document at the same
location it is updated to an anchor with supremal type and attributes.
Otherwise the function is not defined.

\begin{sloppypar}
`\addlink' is the most complex editing function because we
must consider several different cases in that the addition of a link 
can be accepted.
A link of the type `\mkunipp\VMreplacepp *' or `\mkunipp\VMnewpp *' 
may be added when its source contains a specifier 
that refers to an anchor in the the given hyperdocument of 
type `\sourcepp' or `\labelpp'.
For a link of the type `\mkunipp\embedpp\userpp' we additionally require
that this anchor  must point to a location that may carry an embed link. 
For a link of the type `\mkunipp\embedpp\autopp' we additionally require
that the link has exactly one target.
Finally, a link of the type `\bipp' may be added when its source contains 
a specifier that refers to an anchor in
the the given hyperdocument of type `\labelpp'.
\end{sloppypar}

\subsubsection{Hypermedia Document}

The hyperdocument at level\,1 is called\emph{hypermedia document}.
It is an instantiation of the \hd\
object-class and therefore it includes all functions given there. 
Besides that, it provides the two insertion functions
`\insertathmde' and `\insertathmd'. 

\noindent\fbox{\parbox\textwidthminustwomm{
\begin{definition}[Editing Functions ``Hypermedia Document'']\label{edhmd}
\deffunsect{
  \fundecl{\insertathmde}{\hdonesort\times\locisort{\ldonesort}\times\hdonesort\times\addrisort{\hdonesort}}\hdonesort
\\\fundecl{\insertathmd}{\hdonesort\times\locisort{\ldonesort}\times\hdonesort\times\addrisort{\hdonesort}}\hdonesort
}\end{definition}}}

`\insertathmd'
replaces the part of a given hyperdocument, located
at a fixed existing location, with a new hyperdocument. 
The replacement is only possible when the names 
of the anchors in the two hyperdocuments are disjoint
and the replaced part does not carry any anchors.
The result gets the address given
in the last argument of the function
and all links referring to any of two input hyperdocuments are 
changed in order to refer to the the resulting hyperdocument.

`\insertathmde' has the same result as `\insertathmd'
provided that the location actually exists in the
given hyperdocument. 
Otherwise, it generates this location
just as `\insertatlde' from ``Page'', 
\cf\ \sectref{section page in editing functions}.

\section{Conclusion and Outlook}
\label{conclusion}

To our knowledge we have presented the
first\footnote{Note that we do not consider Z to be a formal algebraic
specification language.}\emph{formal algebraic} 
hypertext reference model.
It guarantees a unique understanding and 
enables a close connection to logic-based development
and verification.
With the exception of some deviations in order to be compatible with the \WWW\
it follows the Dexter Hypertext Reference Model (\cf\ \cite{hal90})
and could be seen as an updated formally algebraic version of it.
Additionally, three different levels of hyperdocuments, namely
hypermedia documents, frameset documents, and sites are introduced
---
although the specification of the latter two is still essentially incomplete
and has to be completed in future work.

The hypertext model (\cf\ \sectref{datatype}) 
was developed as a product model 
with the aim to support the design of the product 
``hyperdocument'' automatically. 
It is extended to a model of hypertext-systems (\cf\ \sectref{extdatatype}) 
in order to describe the state transitions of the design-process.  
The whole specification is in the appendix and a prototypical implementation
in ML will be found under {\tt http://www.ags.uni-sb.de/~cp/ml/come.html}.

In this \paper\ we have algebraically specified 
the information-centered model and the
interfaces to the screen-based model.
Before we can start the formalization of the screen-based model, 
we need to study the numerous existing, non-formalized, screen-based
approaches. 
Up to now the favorite idea is to use PDF as a reference model. 
The mapping between
the formalized information-centered model and 
the formalized screen-based model will then provide an
abstract kind of reference user agent (browser), \cf\ Fig.~\vref{overview}. 

\cleardoublepage

\begin{appendix}
\section{The Algebraic Specification}
\label{specp}

\yestop
\subsection{Basic Specifications}

\yestop\noindent
The specifications for \BOOL\ 
(for the Boolean functions), \NAT, \CHAR, \STRING, \TREE, 
\LIST, \LISTPAIR, \SET, \MAPSET, and \MAP\ 
are assumed to be given, but we will present some of their signatures below.

The maximum operator {\maxipp n{n'}} must be defined in the module `\NAT'.
The standard boolean function {\VMisproperprefixtwopp l{l'}} and 
the functions {\reppp n x} 
(which returns a list containing \math x \math n-times),
and \listmappp f l must be defined in the module `\LIST'.

\yestop\yestop\yestop
\noindent
The following parameter specification provides only one single sort.
Note, however, that for any specification we tacitly assume 
the inclusion of the module `\BOOL' 
and the existence of an equality and an inequality predicate which 
exclude each other and are total on objects
described by constructor ground terms (data objects).
 
\yestop\yestop\yestop
\noindent
\ENTRY
\sortsect{
  \entrysort
}

\yestop\yestop\yestop\noindent
Since \SET\ is so fundamental, we present its signature here.\label
{signature of SET}

\yestop\yestop\yestop\noindent
\SET~=~\ENTRY~\andspec~\NAT~\thenspec
\sortsect{
  \setsort=\app\setsort\entrysort
}
\\\funsect{
\\\commentsect
{`\myemptyset' is the empty set.}
\\\fundecl\myemptyset{}\setsort
\\\\\commentsect
{`\emptyquestsymb' test whether a set is empty.}
\\\fundecl\emptyquestsymb\setsort\boolsort
\\\\\commentsect
{Is first argument contained in the second argument?}
\\\fundecl{\mbppp\placeholder\placeholder}{\entrysort\times\setsort}\boolsort
\\\\\commentsect
{`\VMcardpp\placeholder' returns the cardinality (\ie\ the number of elements)
 of a set.}
\\\fundecl{\VMcardpp\placeholder}\setsort\natsort
\\\\\commentsect
{`\inserts' inserts its first argument 
 as an element into its second argument.}
\\\fundecl\inserts{\entrysort\times\setsort}\setsort
\\}\pagebreak\\\phantomaxiomsect{\commentsect
{`\removes' 
 deletes its first argument as an element from its second argument.}
\\\fundecl\removes{\entrysort\times\setsort}\setsort
\\\\\commentsect
{`\unionpp\placeholder\placeholder' returns the union of its arguments.}
\\\fundecl{\unionpp\placeholder\placeholder}{\setsort\times\setsort}\setsort
\\\\\commentsect
{`\intersectionpp\placeholder\placeholder' returns the intersection 
of its arguments.}
\\\fundecl{\intersectionpp\placeholder\placeholder}
    {\setsort\times\setsort}\setsort
\\\\\commentsect
{`\existssymb' tests whether its second argument contains an element
 satisfying its first argument.}
\\\fundecl\existssymb{(\entrysort\rightarrow\boolsort)\times\setsort}\boolsort
}

\yestop\yestop\yestop\yestop\yestop\yestop\noindent
\MAPSET\ will be use to map sets to sets. Note that it cannot be a part
of \SET\ because it needs two sort parameters (one for the domain and
one for the range of the mapping function) instead of one.

\yestop\yestop\noindent
\MAPSET~=~\SET[\renamesort\entrysort{\entrysort1}]~\andspec
~\SET[\renamesort\entrysort{\entrysort2}]~\thenspec
\\\funsect{
\\\commentsect
{`\mapset' replaces all elements of its second argument by their values
 under its first argument.}
\\\fundecl
  \mapset
  {(\entrysort1\rightarrow\entrysort2)\times\app\setsort{\entrysort1}}
  {\app\setsort{\entrysort2}}
}

\yestop\yestop\yestop\yestop\yestop\yestop\noindent
\LISTPAIR\label{where is listpair} 
provides operations on pairs of lists
and is similar to the Standard ML Basis Library module of the same name,
but we need the following non-standard function:

\yestop\yestop\noindent
\LISTPAIR~=~\LIST[\renamesort\entrysort{\entrysortwitharg{D1}}]~\andspec
\\\phantom{\LISTPAIR~=}~\LIST[\renamesort\entrysort{\entrysortwitharg{D2}}]~\andspec
\\\phantom{\LISTPAIR~=}~\LIST[\renamesort\entrysort{\entrysortwitharg R}]~\thenspec
\\\funsect{
\\\commentsect
{`\listpairmapdefault' 
 maps two input lists (fourth and fifth argument) into a new list by 
 applying a binary function (third argument). In case one of the
 input lists is shorter than the other, default values (first and
 second argument) are appended to the shorter list.
}
\\\begin{array}{@{}l l@{}}
  \listpairmapdefault\stopq
  &\entrysortwitharg{D1}
   \times
  \\&\entrysortwitharg{D2}
   \times
  \\&(\entrysortwitharg{D1}\times\entrysortwitharg{D2}\rightarrow\entrysortwitharg R)
   \times
  \\&\app\listsort{\entrysortwitharg{D1}}
   \times
  \\&\app\listsort{\entrysortwitharg{D2}}
  \\&\rightarrow\app\listsort{\entrysortwitharg R}
\end{array}}

\pagebreak

\yestop\yestop\noindent
Since \MAP\ is non-standard, we present its signature here.

\yestop\yestop\noindent
\MAP~=~\SET[\renamesort\entrysort\domainsort]~\andspec
~\SET[\renamesort\entrysort\rangesort]~\thenspec
\sortsect{
  \mapsort=\apptotuple\mapsort{\pair\domainsort\rangesort}
}
\\\funsect{
\\\commentsect
{\mident{empty\_function} is the function with empty domain.}
\\\fundecl{\mident{empty\_function}}{}\mapsort
\\\\\commentsect
{`\upd' returns its third argument but with its second argument being 
the new value of its first argument. UPDate.}
\\\fundecl\upd{\domainsort\times\rangesort\times\mapsort}\mapsort
\\\\\commentsect
{`\applysymb' applies its first argument to its second argument and 
is undefined if the second argument is not in the domain of the 
first argument.}
\\\fundecl\applysymb{\mapsort\times\domainsort}\rangesort
\\\\\commentsect
{`\remsymb' returns its second argument but now undefined for its 
first argument. REMove from domain.}
\\\fundecl\remsymb{\domainsort\times\mapsort}\mapsort
\\\\\commentsect
{DOMain of a function.}
\\\fundecl\domsymb\mapsort{\setof\domainsort}
\\\\\commentsect
{RANge of a function.}
\\\fundecl\ransymb\mapsort{\setof\rangesort}
\\\\\commentsect
{`\revapplysymb' applies the reverse relation of first argument to the 
singleton set containing its second argument. REVerse-APPLY. }
\\\fundecl\revapplysymb{\mapsort\times\rangesort}{\setof\domainsort}
\\\\\commentsect
{`\composemap' unites its first argument with its second argument 
in such a way that first argument wins in case of conflicts.}
\\\fundecl\composemap{\mapsort\times\mapsort}\mapsort
\\\\\commentsect
{`\concatmap' replaces the range elements of its second argument with their
 values under its first argument.}
\\\fundecl\concatmap{(\rangesort\rightarrow\rangesort)\times\mapsort}\mapsort
}

\vfill\pagebreak
\subsection{Parameter Specifications}

\yestop\noindent
The specifications for \URI, \ADDR\HDONE, \ADDR\HDTWO, \ADDR\HDTHREE, 
\ANCHORNAME,
as well as for \SYMBOL\HDONE, \SYMBOL\HDTWO, \SYMBOL\HDTHREE\
and  \ATTj\HDONE, \ATTj\HDTWO, \ATTj\HDTHREE\
are left open and are subject of future work. 

\yestop\yestop\yestop
\noindent
\BASISP\ 
below is merely a parameter specification. 
Intuitively you would expect a rudimentary structure here 
characterizing the genre ``document''. 
For the first level, the\emph{pages}, this structure 
is obvious, for  the second level, the\emph{frames}, 
it seems to be very similar. 
For the third level, the\emph{sites}, it is far from clear, however, 
whether this modeling is actually adequate. 
We therefore have chosen a parameter specification
to ensure sufficient flexibility.

\yestop\yestop\yestop
\noindent
\BASISP~=~\ENTRY[{\renamesort\entrysort {\basisparamsort}}]~\andspec\\
\phantom{\BASISP~=~}\ENTRY[{\renamesort\entrysort {\locationparamsort}}]~\thenspec
\sortsect{
  \basisparamsort
\\\locationparamsort
}
\\\funsect{
\\\commentsect
{\embedlinkokSpp l b tests whether an embed link can be positioned at
location \math l in document \math b.
}\\
  \fundecl{\embedlinkokS}{\locationparamsort\times\basisparamsort}\boolsort
}

\yestop
\yestop
\yestop
\noindent
The following parameter specification provides us with a sort `\addrparamsort'
of addresses for local storage of hyperdocuments.

\yestop
\yestop
\yestop
\noindent
\ADDRP~=~\ENTRY[{\renamesort\entrysort {\addrparamsort}}]

\vfill\pagebreak

\subsection{Anchors}
\label{anchorclass}

\ANCHORCLASS[\locationparamsort]~=~\BASISP[\basisparamsort,\locationparamsort]~\andspec~{\ATTj \ANCHORCLASS}~\thenspec
\vissortsect{
  \anchortypesort
\\\identsorts\anchorsort {\anchorsort(\locationparamsort)}
}
\conssect{
  \sourcepp,\targetpp,\labelpp~:~\anchortypesort
\\\fundecl{\mkanchor}{\locationparamsort\times\anchortypesort\times\attjsort\anchorsort}\anchorsort
}
\deffunsect{
  \projections
\\\fundecl{\locSanchor}{\anchorsort}\locationparamsort
\\\fundecl{\typeSanchor}{\anchorsort}\anchortypesort
\\\fundecl{\attSanchor}{\anchorsort}\attjsort\anchorsort
\\\fundecl{\maxSanchor}{\anchorsort\times\anchorsort}\anchortypesort
\\\edfunctions
\\\fundecl{\chlocation}{\locationparamsort\times\anchorsort}\anchorsort
\\\fundecl{\chanchortype}{\anchortypesort\times\anchorsort}\anchorsort 
\\\fundecl{\addattributeanchor}{\attjsort \anchorsort\times\anchorsort}\anchorsort
\\\fundecl{\delattributeanchor}{\attjsort \anchorsort\times\anchorsort}\anchorsort
}
\varsect{
  \vardecl{o,o'}{\locationparamsort}
\\\vardecl{t,t'}{\anchortypesort}
\\\vardecl{att,att'}{\attjsort\anchorsort}
\\\vardecl{c,c'}{\anchorsort}
}
\\\axiomsect{\phantomaxiomsect{
  \projections
\\\locSanchorpp {\mkanchorpp o t{att}} \PPeq o
\\\typeSanchorpp{\mkanchorpp o t{att}} \PPeq t
\\\attSanchorpp {\mkanchorpp o t{att}} \PPeq att}

\comfundef
{\maxSanchorpp c{c'}}
{Returns the supremal type according to 
`\math{\forall x\stopq x\leq\labelpp}'
 because `\labelpp' can serve both as
 source and as target, while `\sourcepp' and `\targetpp' are incomparable.}
{\simplefundef
{\maxSanchorpp{\mkanchorpp o\labelpp{att}}{c'}
&=\labelpp
&
\\\maxSanchorpp{c}{\mkanchorpp{o'}\labelpp{att'}}
&=\labelpp
&
\\\maxSanchorpp{\mkanchorpp o t{att}}{\mkanchorpp{o'}{t'}{att'}}
&=\labelpp
&\Longleftarrow t\boldunequal t'
\\\maxSanchorpp{\mkanchorpp o t{att}}{\mkanchorpp{o'}{t'}{att'}}
&=t
&\Longleftarrow t\boldequal t'
}}

\yestop
\noindent
\phantomaxiomsect{
\edfunctions
\\\chlocationpp{o'}{\mkanchorpp o t{att}} \PPeq {\mkanchorpp{o'}t{att}}  
\\\chanchortypepp{t'}{\mkanchorpp o t{att}}\PPeq {\mkanchorpp o{t'}{att}}   
\\\addattributeanchorpp{att'}{\mkanchorpp o{t}{att}} 
\PPeq \mkanchorpp o t{\concatpp{att'}{att}}
\\\delattributeanchorpp{att'}{\mkanchorpp o t{att}} 
\PPeq \mkanchorpp o t{\removeattpp{att'}{att}}
 
}}

\newpage

\subsection{Links}
\label{linkclass}

\LINKCLASS~=~\ANCHORNAME~\andspec~{\URI}~\andspec~{\ATTj LINK}~\andspec
\\\phantom{\LINKCLASS~=}~\MAPSET
[\renamesort{\entrysort1}\specifiersort,
 \renamesort{\entrysort2}\specifiersort]~\thenspec
\vissortsect{
  \linktypesort
\\\showsort
\\\actuatesort
\\\specifiersort
\\\linksort
}
\\\conssect{
\\\commentsect
{Links of show-type `\embedpp' embed their target into the context of 
 their source.
 Links of show-type `\VMreplacepp' replace the hyperdocument of their source
 with the hyperdocument of their target. Finally, links of show-type 
 `\VMnewpp'
 open a new window with the document of their target.}
\\\embedpp, \VMreplacepp, \VMnewpp~:~\showsort
\\\\\commentsect
{Links of actuate-type `\userpp' are followed upon user interaction.
 Links of actuate-type `\autopp' are followed automatically.}
\\\userpp, \autopp~:~\actuatesort
\\\\\commentsect
{Links may be uni-directional (`\mkunipp * *') or bi-directional (`\bipp').
 Since bi-directional links are always of show-type `\VMreplacepp'
 and of actuate-type `\userpp', no arguments are given to `\bipp'.}
\\\fundecl{\mkuni}{\showsort\times\actuatesort}\linktypesort
\\\bipp~:~\linktypesort
\\\\\commentsect
{A specifier consists of a global address of sort `\urisort'
 and a local name of sort `\anchornamesort' that is to be mapped to an anchor
 by the hyperdocument under the global address.}
\\\fundecl{\mkspecifier}{\urisort\times\anchornamesort}\specifiersort
\\\fundecl{\mklink}{\setof\specifiersort\times\setof\specifiersort\times\linktypesort\times\attjsort\linksort}\linksort
}
\vfill\pagebreak

\noindent
\deffunsect{
\projections
\\\fundecl{\addSlink}{\specifiersort}\urisort
\\\fundecl{\anchSlink}{\specifiersort}\anchornamesort
\\\fundecl{\sourceSlink}{\linksort}\setof\specifiersort
\\\fundecl{\targetSlink}{\linksort}\setof\specifiersort
\\\fundecl{\specifierSlink}{\linksort}\setof\specifiersort
\\\fundecl{\typeSlink}{\linksort}\linktypesort 
\\\fundecl{\attSlink}{\linksort}\attjsort\linksort
\\\mycomment{Editing~Functions~for~Specifier}
\\\fundecl{\chaddrspec}{\urisort\times\specifiersort}\specifiersort
\\\fundecl{\chname}{\anchornamesort\times\specifiersort}\specifiersort
\\\fundecl{\replaceaddrspec}{\urisort\times\urisort\times\specifiersort}\specifiersort
\\\mycomment{Editing~Functions~for~Link}
\\\fundecl{\insertsource}{\setof\specifiersort\times\linksort}\linksort
\\\fundecl{\deletesource}{\setof\specifiersort\times\linksort}\linksort
\\\fundecl{\inserttarget}{\setof\specifiersort\times\linksort}\linksort
\\\fundecl{\deletetarget}{\setof\specifiersort\times\linksort}\linksort
\\\fundecl{\chlinktype}{\linktypesort\times\linksort}\linksort
\\\fundecl{\addattributelink}{{\attjsort \linksort}\times\linksort}\linksort
\\\fundecl{\delattributelink}{{\attjsort \linksort}\times\linksort}\linksort
\\\fundecl{\replaceaddrlink}{\urisort\times\urisort\times\linksort}\linksort
}
\varsect{
  \vardecl{S,S',S'',S'''}{\setof\specifiersort}
\\\vardecl{s,s'}{\specifiersort}
\\\vardecl{l,l'}{\linksort}
\\\vardecl{L}{\setof\linksort}
\\\vardecl{t,t'}{\linktypesort}
\\\vardecl{n,n'}{\anchornamesort}
\\\vardecl{att,att'}{\attjsort\linksort}
\\\vardecl{a,a',a''}{\urisort}
}\vfill\pagebreak\axiomsect{\phantomaxiomsect{
\projections
\\\addSlinkpp{\mkspecifierpp{a}{n}} \PPeq a
\\\anchSlinkpp{\mkspecifierpp{a}{n}} \PPeq n
\\\sourceSlinkpp{\VMmklinkpp S{S'}t{att}} \PPeq S
\\\targetSlinkpp{\VMmklinkpp S{S'}t{att}} \PPeq S'
\\\specifierSlinkpp{\VMmklinkpp S{S'}t{att}} \PPeq\unionpp S{S'}
\\\typeSlinkpp{\VMmklinkpp S{S'}t{att}} \PPeq t
\\\attSlinkpp{\VMmklinkpp S{S'}t{att}} \PPeq att
\\\mycomment{Editing~Functions~for~Specifier}
\\\chaddrspecpp{a'}{\mkspecifierpp{a}{n}} \PPeq \mkspecifierpp{a'}n
\\\chnamepp{n'}{\mkspecifierpp{a}{n}} \PPeq \mkspecifierpp{a}{n'}
\\\simplefundef
{\VMreplaceaddrspecpp{a'}{a''}{\mkspecifierpp a n}
&\PPeq\mkspecifierpp{a''}n
&\Longleftarrow~a'\boldequal a
\\\VMreplaceaddrspecpp{a'}{a''}{\mkspecifierpp a n} 
&\PPeq\mkspecifierpp a n
&\Longleftarrow~a'\boldunequal a}
\\\mycomment{Editing~Functions~for~Link}
\\\insertsourcepp s{\VMmklinkpp S{S'}t{att}} 
\PPeq  \VMmklinkpp{\insertpp s S}{S'}t{att}
\\\deletesourcepp s{\VMmklinkpp S{S'}t{att}} 
\PPeq  \VMmklinkpp{\removepp s S}{S'}t{att}
\\\inserttargetpp s{\VMmklinkpp S{S'}t{att}} 
\PPeq  \VMmklinkpp S{\insertpp s{S'}}t{att}
\\\deletetargetpp s{\VMmklinkpp S{S'}t{att}} 
\PPeq  \VMmklinkpp S{\removepp s{S'}}t{att}
\\\chlinktypepp{t'}{\VMmklinkpp S{S'}t{att}} 
\PPeq  \VMmklinkpp S{S'}{t'}{att}
\\\addattributelinkpp{att'}{\VMmklinkpp S{S'}t{att}} 
\PPeq \VMmklinkpp S{S'}t{\concatpp{att'}{att}}
\\\delattributelinkpp{att'}{\VMmklinkpp S{S'}t{att}} 
\PPeq \VMmklinkpp S{S'}t{\removeattpp{att'}{att}}
}
 
\comfundef
{\VMreplaceaddrlinkpp{a'}a l=l'}
{Replaces any reference to the URI \math{a'} in the specifiers of the link
\math l with the URI \math a.
\\Note that we can use 
`\replaceaddrspec' as a binary function in the definition 
because we consider all functions to be curried and argument tupling
just to be syntactic sugar.
\\Finally, note that `\mapset' is from \MAPSET
[\renamesort{\entrysort1}\specifiersort,
 \renamesort{\entrysort2}\specifiersort].}
{\VMreplaceaddrlinkpp{a'}a{\VMmklinkpp S{S'}t{att}} \PPeq
\\~\VMmklinkpp
   {\mapsetpp{\VMreplaceaddrspecmappp{a'}a}S}
   {\mapsetpp{\VMreplaceaddrspecmappp{a'}a}{S'}}
   t
   {att}
}}

\newpage

\subsection{Hyperdocuments}
\label{hdclass}

\begin{tabbing}
\HDCLASS[\basisparamsort,\locationparamsort,\addrparamsort]~=~ \= \HDCLASS~ \= \hspace*{-6mm} \=\kill
\HDCLASS[\basisparamsort,\locationparamsort,\addrparamsort]~= \>  
   \BASISP[\basisparamsort,\locationparamsort]~\andspec\\
\> \ADDRP[\addrparamsort]~\andspec\\
\> \ANCHORCLASS[\locationparamsort]~\andspec\\
\> \LINKCLASS~\andspec\\
\> {\ATTj \HDCLASS}~\andspec\\
\> \MAPof{\renamesort\domainsort\anchornamesort,
          \renamesort\rangesort\anchorsort}~\andspec\\
\> \SETof{\renamesort\entrysort\linksort}
\end{tabbing}
\thenspec
\vissortsect{
  \identsorts\hdsort {\mident{\hdsort(\basisparamsort,\locationparamsort,\addrparamsort)}}
}
\conssect{
  \fundecl{\mkhd}{\basisparamsort\times\mapof{\anchornamesort}{\anchorsort}\times\setof\linksort\times\attjsort\hdsort\times\addrparamsort}\hdsort
}
\deffunsect{
  \projections
\\\fundecl{\ldShd}{\hdsort}\basisparamsort
\\\fundecl{\anchorsShd}{\hdsort}\mapof\anchornamesort\anchorsort
\\\fundecl{\linkShd}{\hdsort}\setof\linksort
\\\fundecl{\attShd}{\hdsort}\attjsort\hdsort
\\\fundecl{\addShd}{\hdsort}\addrparamsort
\\\fundecl{\anchorShd}{\anchornamesort\times\mapof\anchornamesort\anchorsort}\anchorsort
\\\fundecl{\anchornameShd}{\anchorsort\times\mapof\anchornamesort\anchorsort}\setof\anchornamesort
\\\edfunctions
\\\fundecl{\addanchor}{\anchornamesort\times\anchorsort\times\hdsort}\hdsort
\\\fundecl{\delanchor}{\anchornamesort\times\hdsort}\hdsort
\\\fundecl{\addlink}{\linksort\times\hdsort}\hdsort
\\\fundecl{\dellink}{\linksort\times\hdsort}\hdsort
\\\fundecl{\addattributehd}{\attjsort \hdsort\times\hdsort}\hdsort
\\\fundecl{\delattributehd}{\attjsort \hdsort\times\hdsort}\hdsort
\\\fundecl{\chaddr}{\addrparamsort\times\hdsort} \hdsort
\\\mycomment{Converter~Functions}
\\\fundecl{\integrate}{\addrparamsort}\urisort
}
\varsect{
  \vardecl{d,d'}{\basisparamsort}
\\\vardecl{L,L'}{\setof\linksort}
\\\vardecl{l}{\linksort}
\\\vardecl{act}{\actuatesort}
\\\vardecl{sp,sp'}{\specifiersort}
\\\vardecl{A,A'}{\mapof\anchornamesort\anchorsort}
\\\vardecl{c,c'}{\anchorsort}
\\\vardecl{a,a',a''}{\addrparamsort}
\\\vardecl{att,att'}{\attjsort\hdsort}
\\\vardecl{n}{\anchornamesort}
}
\vfill\pagebreak\axiomsect{\phantomaxiomsect{
  \projections
\\\ldShdpp{\VMmkhdpp{d}{A}{L}{att}{a}} \PPeq d
\\\anchorsShdpp{\VMmkhdpp{d}{A}{L}{att}{a}} \PPeq A
\\\linkShdpp{\VMmkhdpp{d}{A}{L}{att}{a}} \PPeq L
\\\attShdpp{\VMmkhdpp{d}{A}{L}{att}{a}} \PPeq att
\\\addShdpp{\VMmkhdpp{d}{A}{L}{att}{a}} \PPeq a}

\comfundef
{\anchorShdpp n A}
{Returns the anchor referred to by the name \math n
 by calling the function `\applysymb' from \MAP.}
{\simplefundef
{\anchorShdpp n A
&=\applypp A n
}}

\comfundef
{\anchornameShdpp c A}
{Returns the set of all names referring to the anchor \math c
 by calling the function `\revapplysymb' from \MAP.}
{\simplefundef
{\anchornameShdpp c A
&= \revapplypp A c
}}}

\yestop
\noindent
\edfunctions

\comfundef
{\addanchorpp{n}{c} h = h'}
{\math{h'} is the hyperdocument after the anchor \math c with name $n$ 
 has been added to the anchors of hyperdocument \math h, 
 provided that an anchor with this name does not exist in \math h before. 
 If an anchor with name \math n does exist in h at the same location as anchor
 \math c, then 
 \math{h'} is updated to an anchor with supremal type and attributes.
 Note that we use `\upd' from \MAP\ and write long argument lists vertically
 instead of horizontally.}
{\simplefundef
{\addanchorpp n c{\VMmkhdpp{d}{A}{L}{att}{a}} \PPeq \VMmkhdpp d{\updpp n c A}L{att}a
\\~~~\Longleftarrow\ (n\in\domainSpp A) \PPeq \false 
\\\addanchorpp{n}{c}{\VMmkhdpp{d}{A}{L}{att}{a}} \PPeq
\\~~{\begin{array}{@{}l@{}l@{}}
\mkhd(&d,
\\&{\begin{array}{@{}l@{}l@{}}
\upd(
&n 
\\&{\begin{array}{@{}l@{}l@{}}
\mkanchor(
&\locSanchorpp c,
\\&\maxSanchorpp c{c'},
\\&\concatpp{\attSanchorpp c}{\attSanchorpp{c'}}),
\end{array}}
\\&A),
\end{array}}
\\&L,
\\&{att},
\\&a)
\end{array}}
\\~~~\Longleftarrow\ (n\in\domainSpp A) = \true \und \anchorShdpp n A=c' \und \locSanchorpp c=\locSanchorpp{c'}}}

\comfundef
{\delanchorpp n h=h'}
{$h'$ is the hyperdocument after the anchor with the name $n$ 
 has been removed from the hyperdocument $h$.}
{\delanchorpp{n}{\VMmkhdpp{d}{A}{L}{att}{a}} \PPeq 
 \VMmkhdpp{d}{\rempp{n}{A}}{L}{att}{a}
\pagebreak
}

\comfundef
{\VMaddlinkpp l h=h'}
{\math{h'} is the hyperdocument after the link \math l has been added 
 to the set of links in \math h.
\\
 A link of the type `\mkunipp\VMreplacepp *' or `\mkunipp\VMnewpp *' 
 may be added when its source contains a specifier \math{sp}
 that refers to an anchor \math c in
 the the given hyperdocument of type `\sourcepp' or `\labelpp'.
 This is expressed in the first four rules.
\\
 For a link of the type `\mkunipp\embedpp\userpp' we additionally require
 that this anchor \math c 
 must point to a location that may carry an embed link. 
 This is expressed in the next two rules.
 Note that `\embedlinkokS' comes from \BASISP.
\\
 For a link of the type `\mkunipp\embedpp\autopp' we additionally require
 that the link has exactly one target.
 This is expressed in the next two rules.
\\
 Finally, a link of the type `\bipp' 
 may be added when its source contains a specifier \math{sp}
 that refers to an anchor \math c in
 the the given hyperdocument of type `\labelpp'.}
{\VMaddlinkpp{l}{\VMmkhdpp{d}{A}{L}{att}{a}} \PPeq 
 \VMmkhdpp{d}{A}{\VMinsertpp{l}{L}}{att}{a}
\\~~~\Longleftarrow~\typeSlinkpp{l}\boldequal\mkunipp{\VMreplacepp}{act}
 \und 
 sp\tightin\sourceSlinkpp l
 \und 
\\\phantom{~~~\Longleftarrow~}
 \addSlinkpp{sp}\boldequal\integratepp{a}
 \und
 \anchorShdpp{\anchSlinkpp{sp}}{A}\boldequal c\und
 \typeSanchorpp c\boldequal\sourcepp
\\\VMaddlinkpp{l}{\VMmkhdpp{d}{A}{L}{att}{a}} \PPeq 
 \VMmkhdpp{d}{A}{\VMinsertpp{l}{L}}{att}{a}
\\~~~\Longleftarrow~\typeSlinkpp{l}\boldequal\mkunipp{\VMreplacepp}{act}
 \und 
 sp\tightin\sourceSlinkpp l
 \und 
\\\phantom{~~~\Longleftarrow~}
 \addSlinkpp{sp}\boldequal\integratepp{a}
 \und
  \anchorShdpp{\anchSlinkpp{sp}}{A}\boldequal c\und
  \typeSanchorpp c\boldequal\labelpp
\\\VMaddlinkpp{l}{\VMmkhdpp{d}{A}{L}{att}{a}} \PPeq 
 \VMmkhdpp{d}{A}{\VMinsertpp{l}{L}}{att}{a}
\\~~~\Longleftarrow~\typeSlinkpp{l}\boldequal\mkunipp\VMnewpp{act}
 \und 
 sp\tightin\sourceSlinkpp l
 \und 
\\\phantom{~~~\Longleftarrow~}
 \addSlinkpp{sp}\boldequal\integratepp{a}
 \und
  \anchorShdpp{\anchSlinkpp{sp}}{A}\boldequal c\und
  \typeSanchorpp c\boldequal\sourcepp
\\\VMaddlinkpp{l}{\VMmkhdpp{d}{A}{L}{att}{a}} \PPeq 
 \VMmkhdpp{d}{A}{\VMinsertpp{l}{L}}{att}{a}
\\~~~\Longleftarrow~\typeSlinkpp{l}\boldequal\mkunipp\VMnewpp{act}
 \und 
 sp\tightin\sourceSlinkpp l
 \und 
\\\phantom{~~~\Longleftarrow~}
 \addSlinkpp{sp}\boldequal\integratepp{a}
 \und
  \anchorShdpp{\anchSlinkpp{sp}}{A}\boldequal c\und
  \typeSanchorpp c\boldequal\labelpp
\\\VMaddlinkpp{l}{\VMmkhdpp{d}{A}{L}{att}{a}} \PPeq 
  \VMmkhdpp{d}{A}{\VMinsertpp{l}{L}}{att}{a}
\\~~~\Longleftarrow~
  \typeSlinkpp l\boldequal\mkunipp\embedpp\userpp
  \und 
  sp\tightin\sourceSlinkpp l
  \und 
\\\phantom{~~~\Longleftarrow~}
  \addSlinkpp{sp}\boldequal\integratepp{a}
  \und
  \anchorShdpp{\anchSlinkpp{sp}}{A}\boldequal c\und
  \typeSanchorpp c\boldequal\sourcepp\und
\\\phantom{~~~\Longleftarrow~}
  \embedlinkokSpp{\locSanchorpp c}d
\\\VMaddlinkpp{l}{\VMmkhdpp{d}{A}{L}{att}{a}} \PPeq 
  \VMmkhdpp{d}{A}{\VMinsertpp{l}{L}}{att}{a}
\\~~~\Longleftarrow~
  \typeSlinkpp l\boldequal\mkunipp\embedpp\userpp
  \und 
  sp\tightin\sourceSlinkpp l
  \und 
\\\phantom{~~~\Longleftarrow~}
  \addSlinkpp{sp}\boldequal\integratepp{a}
  \und
  \anchorShdpp{\anchSlinkpp{sp}}{A}\boldequal c\und
  \typeSanchorpp c\boldequal\labelpp\und
\\\phantom{~~~\Longleftarrow~}
  \embedlinkokSpp{\locSanchorpp c}d
\\\VMaddlinkpp{l}{\VMmkhdpp{d}{A}{L}{att}{a}} 
  \PPeq \VMmkhdpp{d}{A}{\VMinsertpp{l}{L}}{att}{a}
\\~~~\Longleftarrow~
  \typeSlinkpp l\boldequal\mkunipp\embedpp\autopp
  \und 
  sp\tightin\sourceSlinkpp l
  \und 
\\\phantom{~~~\Longleftarrow~}
  \addSlinkpp{sp}\boldequal\integratepp{a}
  \und
  \anchorShdpp{\anchSlinkpp{sp}}{A}\boldequal c\und
  \typeSanchorpp c\boldequal\sourcepp\und
\\\phantom{~~~\Longleftarrow~}
  \embedlinkokSpp{\locSanchorpp c}d
  \und
  \VMcardpp{\targetSlinkpp{l}}\boldequal 1
\\\VMaddlinkpp{l}{\VMmkhdpp{d}{A}{L}{att}{a}} 
  \PPeq \VMmkhdpp{d}{A}{\VMinsertpp{l}{L}}{att}{a}
\\~~~\Longleftarrow~
  \typeSlinkpp l\boldequal\mkunipp\embedpp\autopp
  \und 
  sp\tightin\sourceSlinkpp l
  \und 
\\\phantom{~~~\Longleftarrow~}
  \addSlinkpp{sp}\boldequal\integratepp{a}
  \und
  \anchorShdpp{\anchSlinkpp{sp}}{A}\boldequal c\und
  \typeSanchorpp c\boldequal\labelpp\und
\\\phantom{~~~\Longleftarrow~}
  \embedlinkokSpp{\locSanchorpp c}d
  \und
  \VMcardpp{\targetSlinkpp{l}}\boldequal 1
\\\VMaddlinkpp{l}{\VMmkhdpp{d}{A}{L}{att}{a}} \PPeq \VMmkhdpp{d}{A}{\VMinsertpp{l}{L}}{att}{a}
\\~~~\Longleftarrow~
  \typeSlinkpp l\boldequal\bipp
  \und 
  sp\tightin\sourceSlinkpp l
  \und 
\\\phantom{~~~\Longleftarrow~}
  \addSlinkpp{sp}\boldequal\integratepp{a}
  \und
  \anchorShdpp{\anchSlinkpp{sp}}{A}\boldequal c\und
  \typeSanchorpp c\boldequal\labelpp
}

\comfundef
{\dellinkpp l h=h'}
{$h'$ is the hyperdocument after the link $l$ is removed from the 
 hyperdocument $h$.}
{\dellinkpp{l}{\VMmkhdpp{d}{A}{L}{att}{a}} \PPeq 
 \VMmkhdpp{d}{A}{\removepp{l}{L}}{att}{a}}

\comfundef
{\addattributehdpp{att}h=h'}
{$h'$ is the hyperdocument after the hyperdocument $h$ is enriched with the 
 attributes $att$.}
{\addattributehdpp{att'}{\VMmkhdpp{l}{A}{L}{att}{a}} \PPeq 
 \VMmkhdpp{l}{A}{L}{\concatpp{att'}{att}}{a}}

\comfundef
{\delattributehdpp{att}h=h'}
{$h'$ is the hyperdocument after the attributes $att$ are removed from the 
 hyperdocument \nolinebreak$h$.}
{\delattributehdpp{att'} {\VMmkhdpp{l}{A}{L}{att}{a}} \PPeq 
 \VMmkhdpp{l}{A}{L}{\removeattpp{att'}{att}}{a}}

\comfundef
{\chaddrpp{a'}{h} = h'}
{$h'$ is the hyperdocument after the address of $h$ is 
 replaced by address $a'$.}
{\chaddrpp{a'}{\VMmkhdpp{d}{A}{L}{att}{a}}\PPeq 
 {\VMmkhdpp{d}{A}{L}{att}{a'}}}

\newpage


\subsection{Media Objects}\label{medienobject}

\noindent
\MO~=~\URI~\andspec~\ANCHORNAME~\andspec
~\SET[\renamesort\entrysort\anchornamesort]~\thenspec
\vissortsect{
  \mosort=\apptotuple\mosort{\pair\urisort\anchornamesort}
}
\conssect{
\\\commentsect
{Our interface to media-objects is quite simple because we are not
interested in modeling their internal behavior. The only thing we require
is that they have some unified resource identifier of sort `\urisort'
and a set of anchor identifiers to which links may refer.
Thus, a media-object basically introduces a legal set of specifiers
referring to it.}
\\\fundecl{\mkmo}{\urisort\times\setof\anchornamesort}\mosort
}

\newpage
\subsection{Hypermedia Document Level}

\subsubsection{Page}

\noindent
\SYMBOL{\LDONE}~=~\STRING~\thenspec
\vissortsect{
  \symbolisort{\ldonesort}
}

\yestop
\yestop
\yestop
\noindent
\LDONE~=~\MO~\andspec~\SYMBOL{\LDONE}~\andspec~{\ATTj \LDONE}~\andspec\\
\phantom{\LDONE~=~}\TREE[\renamesort\entrysort \structitypesort\ldonesort]~\andspec
\\\phantom{\LDONE~=~}%
\LIST[\renamesort\entrysort \ldonesort]~\andspec
\\\phantom{\LDONE~=~}%
\LISTPAIR[\renamesort{\entrysortwitharg{D1}}\natsort,
           \renamesort{\entrysortwitharg{D2}}\natsort,
           \renamesort{\entrysortwitharg   R}\natsort
]~\thenspec
\vissortsect{ 
  \ldonesort
\\\structitypesort\ldonesort
\\\identsorts{\locisort{\ldonesort}} {\mident{\listsort(\natsort)}}
}
\conssect{
  \basic,\symb,\emptypage,\hdlist,\hdtable,\hdtableline,\hdheadline,\hdpage,\hdtext,\\
  \hspace{5pt}\hdlinebreak,\hdfootnote,\hdparagraph,\hdcopyright~:~\structitypesort\ldonesort
\\\mtpage~:~\ldonesort
\\\fundecl{\imphdlow}{\mosort}\ldonesort
\\\fundecl{\impsymbol}{\symbolisort{\ldonesort}}\ldonesort
\\\fundecl{\mkld}{{\structitypesort\ldonesort}\times\listof{\ldonesort}\times\attjsort\ldonesort}\ldonesort
}
\deffunsect{
  \projections
\\\fundecl{\atomicS}{\ldonesort}\boolsort
\\\fundecl{\hasnthS}{\natsort\times\listof\ldonesort}\boolsort
\\\fundecl{\haslocationS}{\locisort{\ldonesort}\times\ldonesort}\boolsort
\\\fundecl{\includelinkokldoneS}{\locisort\ldonesort\times\ldonesort}\boolsort
\\\fundecl{\structSld}{\ldonesort}\treeof{\structitypesort\ldonesort}
\\\fundecl\pagesSld\ldonesort{\listof\ldonesort}
\\\fundecl\attSld\ldonesort{\attjsort\ldonesort}
\\\fundecl{\pnth}{\natsort\times\listof\ldonesort}\ldonesort 
\\\fundecl{\locate}{\locisort\ldonesort\times\ldonesort}\ldonesort
\\\fundecl\dimensionpagesymb{\ldonesort}\listof\natsort
\\\fundecl\dimensionpagelistsymb{\listof\ldonesort}\listof\natsort
\\\edfunctions
\\\fundecl{\changestruct}{\structitypesort\ldonesort\times\ldonesort}\ldonesort
\\\fundecl{\mklist}{\natsort}\ldonesort
\\\fundecl{\mktable}{\natsort\times\natsort}\ldonesort
\\\fundecl{\mktableline}{\natsort}\ldonesort
\\\fundecl{\insertatlde}{\ldonesort\times \locisort{\ldonesort}\times\ldonesort}\ldonesort
\\\fundecl{\insertatldee}{\ldonesort\times\natsort\times \locisort{\ldonesort}\times\listof{\ldonesort}\times\ldonesort}\listof{\ldonesort}
\\\fundecl{\insertatld}{\ldonesort\times\locisort\ldonesort\times\ldonesort}\ldonesort
\\\fundecl{\addattributeld}{\attjsort \ldonesort\times\ldonesort}\ldonesort
\\\fundecl{\delattributeld}{\attjsort \ldonesort\times\ldonesort}\ldonesort
}
\pagebreak
\varsect{
  \vardecl{h}{\mosort}
\\\vardecl{symb}{\symbolisort{\ldonesort}}
\\\vardecl{p,p',p'',p'''}{\ldonesort}
\\\vardecl{s,s'}{\structitypesort\ldonesort}
\\\vardecl{P}{\listof{\ldonesort}}
\\\vardecl{n}{\natsort}
\\\vardecl{o}{\locisort{\ldonesort}}
\\\vardecl{att,att'}{\attjsort\ldonesort}
}

\axiomsect{\phantomaxiomsect{
\projections
\\\atomicSpp{\mtpage} \PPeq \true
\\\atomicSpp{\imphdlowpp h} \PPeq \true
\\\atomicSpp{\impsymbolpp{symb}} \PPeq \true
\\\atomicSpp{\VMmkldpp s P{att}} \PPeq \false
}

\comfundef
{\haslocationSpp o p}
{Tests whether location \math o occurs in page $p$. 
 The empty location $\myemptylist$ means the whole page and therefore it 
 exists in every page. \hasnthSpp o P is an auxiliary function for it.}
{\simplefundef
{\hasnthSpp{\succpp 0}{\myemptylist}
&=
&\false
\\\hasnthSpp{\succpp 0}{\inslistpp p P}
&=
&\true
\\\hasnthSpp{\succpp{\succpp n}}{\inslistpp p P}
&=
&\hasnthSpp{\succpp n}P}
\\\simplefundef
{\haslocationSpp{\myemptylist}p 
  =\true
\\\haslocationSpp{\inslistpp{\succpp n}o}p = \false
~~~\\\Longleftarrow\ \atomicSpp p =\true
\\\haslocationSpp{\inslistpp{\succpp n}o}p = \false
~~~\\\Longleftarrow\ \atomicSpp p =\false
\und p = \VMmkldpp s P{att}
\und\hasnthSpp{\succpp n}P = \false
\\\haslocationSpp{\inslistpp{\succpp n}o}p
=\haslocationSpp o{\pnthpp{\succpp n}P}
~~~\\\Longleftarrow\ \atomicSpp p = \false
\und p = \VMmkldpp s P{att}
\und\hasnthSpp{\succpp n}P = \true}
}

\comfundef{\includelinkokldoneSpp o p}
{Returns `\true' if the location \math o exists in page $p$ and 
 the document located at \math o is an empty page \mtpage. 
 If location \math o does not exist in page $p$ it returns `\false'.}
{\simplefundef
{\includelinkokldoneSpp o p
&=\false
&\Longleftarrow~\haslocationSpp o p\boldequal\false
\\\includelinkokldoneSpp o p
&=\true
&\Longleftarrow~\haslocationSpp o p\boldequal\true\und
 \locatepp o p\boldequal\mtpage}}
\vfill\pagebreak

\comfundef
{\structSldpp p}
{Returns the tree of structures in page \math p.
Notice that it uses the function `\listmap' from \LIST\
that runs the function in its first argument over the list in its second
argument.}
{
\simplefundef
{\structSldpp{\mtpagepp{}}
&\PPeq
&\VMmktreepp\emptypage\myemptylist
\\\structSldpp{\imphdlowpp h}
&\PPeq
&\VMmktreepp\basic\myemptylist
\\\structSldpp{\impsymbolpp{symb}}
&\PPeq
&\VMmktreepp\symb\myemptylist
\\\structSldpp{\VMmkldpp s P{att}}
&\PPeq
&\VMmktreepp s{\listmappp\structSld P}}
}

\comfundef
{\pagesSldpp p = P}
{$P$ are the top level elements of page $p$.}
{\attSldpp{\VMmkldpp s{P}{att}} \PPeq P}

\comfundef
{\attSldpp p = att}
{$att$ are the top level attributes of page $p$.}
{\attSldpp{\VMmkldpp s{P}{att}} \PPeq att}

\comfundef
{\VMpnthpp{\succpp n}P}
{Computes the \nth n element of the list P,
 but starts with 1 (instead of 0).}
{\VMpnthpp{\succpp n}P \PPeq \nthpp{n}{P}}

\comfundef
{\VMlocatepp o p=p'}
{$p'$ is the the page located at position \math o in page \math p.}
{\VMlocatepp{\myemptylist}p\PPeq p
\\\VMlocatepp{\inslistpp{\succpp n}o}{\VMmkldpp s P{att}} \PPeq \VMlocatepp o{\VMpnthpp{\succpp n}P}}
\pagebreak

\comfundef
{\dimensionpagepp p}
{Returns the list of natural numbers of the sizes of 
 the page object \math p
 in all its dimensions. \Eg, a two dimensional 
 table with \math m lines and a maximum of
 \math n columns in one of these lines has a dimension of \math{(m,n)}.
 This means that the smallest two dimensional cube around it will
 have hight \math m and breadth \math n.
 A three dimensional table with dimension \math{(m,n,p)} will fill
 a cube of depth \math p.
 If the objects are not atomic, the element-wise maximum of its dimensions
 will be appended at the end of the dimension list of the table.
 Generally speaking, a page object represented as an \mkld-node tree
 of depth \nolinebreak\math d has the dimension \math{(n_1,\ldots,n_d)} where
 \math{n_i} is the maximum number of children of a node at depth \math i.
 Note that it uses the function `\listpairmapdefault' from \LISTPAIR\
 on page~\pageref{where is listpair}.}
{
\simplefundef
{\dimensionpagepp p = \myemptylist
\\~~~\Longleftarrow\ \atomicSpp p = \true
\\\dimensionpagepp p = \inslistpp{\lengthpp P}{\dimensionpagelistpp P}
\\~~~\Longleftarrow\ \atomicSpp p = \false\und p = \VMmkldpp s P{att}}
\\\simplefundef{\dimensionpagelistpp{\myemptylist} & \PPeq & \myemptylist
\\\dimensionpagelistpp{\inslistpp{p}{P}} & \PPeq & 
\\\multicolumn{3}{@{~~~}l}{\listpairmapdefaultpp00{\maxi}{\dimensionpagepp p}{\dimensionpagelistpp P}}}}

\yestop
\noindent
\edfunctions

\comfundef
{\changestructpp{s'}p=p'}
{$p'$ is the page containing the same documents and attributes as $p$, 
 but with a different structure $s'$.}
{\changestructpp{s'}{\VMmkldpp s{P}{att}} \PPeq \VMmkldpp{s'}{P}{att}}

\comfundef
{\mklistpp{n} = p}
{$p$ is a list with $n$ items, 
 containing an empty page $\mtpage$ in every item.}
{\mklistpp{n} \PPeq \VMmkldpp{\hdlist}{\reppp{n}{\mtpage}}{\emptyatt}}

\comfundef
{\VMmktablepp{m}{n} = p}
{$p$ is a \math{m\times n}-table, containing an empty page $\mtpage$ in every 
 cell. \VMmktablelinepp{n}  is an auxiliary function for it.}
{\VMmktablepp{m}{n} \PPeq \VMmkldpp{\hdtable}{\reppp{m}{\mktablelinepp{n}}}{\emptyatt}
\\\VMmktablelinepp{n} \PPeq \VMmkldpp{\hdtableline}{\reppp{n}{\mtpage}}{\emptyatt}}

\comfundef
{\insertatldepp{p'}o p=p''} 
{If the location \math o occurs in the page \math p, then
 \math{p''} is the page \math p with its part at location \math o
 replaced with the page \math{p'}.
\\
  If \math o does not exist in \math p 
  because a node \math\nu\ in \math p has not enough children,
  then \math p is first extended with sufficiently many child nodes
  for \math\nu.
  The type of these child nodes may depend on the parent node \math\nu.
  \Eg, if the parent node is a table then the child nodes will be of
  type table-line. If no special knowledge is given, the child nodes
  will be simply of type empty page (`\mtpage').
  The default child node is the last argument of a helper function
  `\insertatldee' that is very similar to `\insertatlde' but works on
  children lists instead of single nodes.}
{\insertatldepp{p'}\myemptylist p
=p'
\\\insertatldepp{p'}{\inslistpp n o}{\VMmkldpp s P{att}}
\\\phantom{\insertatldepp{p'}\myemptylist p}
=\VMmkldpp s{\insertatldeepp{p'}n o P{\VMmktablelinepp 0}}{att}
\\\phantom{\insertatldepp{p'}\myemptylist p}
\Longleftarrow~s \PPeq \hdtable
\\\insertatldepp{p'}{\inslistpp n o}{\VMmkldpp s{P}{att}}
\\\phantom{\insertatldepp{p'}\myemptylist p}
=\VMmkldpp s{\insertatldeepp{p'}n o P{\mtpage}}{att}
\\\phantom{\insertatldepp{p'}\myemptylist p}
\Longleftarrow~s \not \PPeq \hdtable
\\\simplefundef
{\insertatldeepp{p'}{\succpp 0}o{\inslistpp{p}{P}}{p''}
&=\inslistpp{\insertatldepp{p'}o p}P
\\\insertatldeepp{p'}{\succpp{\succpp n}}o{\inslistpp{p}{P}}{p''}
&=\inslistpp p{\insertatldeepp{p'}{\succpp n}o P{p''}}
\\\insertatldeepp{p'}{\succpp 0}o{\myemptylist}{p''}
&=\inslistpp{\insertatldepp{p'}o{p''}}{\myemptylist}
\\\insertatldeepp{p'}{\succpp{\succpp n}}o{\myemptylist}{p''}
&=\inslistpp{p''}{\insertatldeepp{p'}{\succpp n}o\myemptylist{p''}}}}

\comfundef
{\insertatldpp{p'}o p=p''}
{$p''$ is the page after $p'$ has been inserted at location \math o
 if \math o exists in $p$.}
{\insertatldpp{p'}o p\PPeq\insertatldepp{p'}o p~\Longleftarrow~\haslocationSpp o p~\PPeq~\true}

\comfundef
{\addattributeldpp{att}{p} = p'}
{$p'$ is the page after $p$ is enriched with the attributes $att$.}
{\addattributeldpp{att'}{\VMmkldpp s{P}{att}} \PPeq 
 \VMmkldpp s{P}{\concatpp{att'}{att}}}

\comfundef
{\delattributeldpp{att}{p} = p'}
{$p'$ is the page after the attributes $att$ are removed from $p$.}
{\delattributeldpp{att'}{\VMmkldpp s{P}{att}} \PPeq \VMmkldpp s{P}{\removeattpp{att'}{att}}}}
\newpage

\subsubsection{HyperMedia Document}

\yestop
\yestop
\yestop
\noindent
\ADDR{\HDONE}~=~\STRING~\thenspec
\vissortsect{
  \addrisort{\hdonesort}
}

\yestop\yestop
\yestop
\yestop
\noindent
\begin{tabbing}
\HDONE~=~ \= \= \HDCLASS \= \kill
\HDONE~=~ \> \LDONE~\andspec~\ADDR{\HDONE}~\andspec~
\\
\>\> \HDCLASS [{\renamesort\basisparamsort{\LDONE.\ldonesort}},\\
\>\>\> {~\renamesort\locationparamsort {\LDONE.\locisort\ldonesort}},\\
\>\>\> {~\renamesort\embedlinkokS{\LDONE.\includelinkokldoneS}},\\
\>\>\> {~\renamesort\addrparamsort{\ADDR\HDONE.\addrisort\hdonesort}}]
~\andspec
\\\>\MAPSET
[\renamesort{\entrysortwitharg1}\anchorsort,
 \renamesort{\entrysortwitharg2}{\LDONE.\locisort\ldonesort}]~\andspec
\\\>\MAPSET
[\renamesort{\entrysortwitharg1}\linksort,
 \renamesort{\entrysortwitharg2}\linksort]
\end{tabbing}
\thenspec
\vissortsect{ 
  \hdonesort=
  \apptotuple\hdsort{\trip
    {\LDONE.\ldonesort}
    {\LDONE.\locisort\ldonesort}
    {\ADDR\HDONE.\addrisort\hdonesort}}}
\deffunsect{
\edfunctions
\\\fundecl{\insertathmde}{\hdonesort\times\locisort{\ldonesort}\times\hdonesort\times\addrisort{\hdonesort}}\hdonesort
\\\fundecl{\insertathmd}{\hdonesort\times\locisort{\ldonesort}\times\hdonesort\times\addrisort{\hdonesort}}\hdonesort
\\\fundecl{\combinelink}{\addrisort{\hdonesort}\times\addrisort{\hdonesort}\times\addrisort{\hdonesort}\times\setof{\linksort}\times\setof{\linksort}}\setof{\linksort}
\\\fundecl{\sinklocation}{\locisort\ldonesort\times\anchorsort}\anchorsort
}
\varsect{
  \vardecl{m,n}{\natsort}
\\\vardecl{p,p'}{\ldonesort}
\\\vardecl{h,h',h''}{\hdonesort}
\\\vardecl{o,o'}{\locisort{\ldonesort}}
\\\vardecl{A,A'}{\mapof\anchornamesort\anchorsort}
\\\vardecl{L,L'}{\setof\linksort}
\\\vardecl{a,a',a''}{\addrisort{\hdonesort}}
\\\vardecl{t}{\anchortypesort}
\\\vardecl{att}{\attjsort\anchorsort}
}

\vfill\pagebreak
\axiomsect{

\yestop
\noindent
\edfunctions

\comfundef
{\insertathmdepp h o{h'}{a''}=h''}
{Replaces the part of hyperdocument \math{h'}
 at location \math o with hyperdocument \math h,
 resulting in a new hyperdocument \math{h''} under address \math{a''}.
 This is only possible when the names of the anchors in \math h and
 \math{h'} are disjoint and when \math{h'} does not have any
 anchors in the part replaced with \math h.
\\\sinklocationpp o c is an auxiliary function that appends \math o
to the front of the location of the anchor \math c, \ie\ it lets
\math c sink below the location \math o. Note that we can use 
`\sinklocation' as a unary function in the definition of `\insertathmde'
because we consider all functions to be curried and argument tupling
just to be syntactic sugar.
\\The functions `\concatmap' and `\composemap' are from 
\MAPof{\renamesort\domainsort\anchornamesort,
       \renamesort\rangesort\anchorsort} from \HDCLASS.
Note that the application of `\concatmap' 
is unproblematic here because the domains of 
\math A and \math{A'} are required to be disjoint.
\\`\combinelink' is an auxiliary function that changes all references
of links to \math h and \math{h'} to refer to \math{h''}.
It is defined via `\mapset' from \MAPSET
[\renamesort{\entrysortwitharg1}\linksort,
 \renamesort{\entrysortwitharg2}\linksort].
Moreover, `\mbox\replaceaddrlink' from \LINKCLASS\ is called (like
`\sinklocation') with one argument less than defined, 
in order to yield
a function of type `\math{\linksort\rightarrow\linksort}'.
\\Finally, note that in the condition of the definition
of `\insertathmde' the `\mapset'  is from 
\MAPSET
[\renamesort{\entrysortwitharg1}\anchorsort,
 \renamesort{\entrysortwitharg2}{\LDONE.\locisort\ldonesort}]
and the \mbox{`\existssymb'} is from 
\SETof{\renamesort\entrysort{\LDONE.\locisort\ldonesort}},
which again is part of 
\MAPSET
[\renamesort{\entrysortwitharg1}\anchorsort,
 \renamesort{\entrysortwitharg2}{\LDONE.\locisort\ldonesort}].}
{\insertathmdepp
   {\VMmkhdpp{p}{A}{L}{att}{a}}
   {o}
   {\VMmkhdpp{p'}{A'}{L'}{att'}{a'}}
   {a''} 
=
\\\begin{array}{@{~~~~}l l@{}}
\multicolumn{2}{@{~~~~}l@{}}{\VMmkhdpplong
  {\insertatldepp p o{p'}}
  {\composemappp 
    {\concatmappp A{\sinklocationmappp o}}
    {A'}}
  {\combinelinkpp a{a'}{a''}L{L'}}
  {\concatpp{att}{att'}}
  {a''}}
\\\Longleftarrow
&\intersectionpp{\domainSpp{A}}{\domainSpp{A'}}\boldequal\myemptyset\und
\\&\existspp{\VMisproperprefixpp o}{\mapsetpp{\locSanchor}{\ranpp{A}}}
\boldequal\false\end{array}
\\\\
\sinklocationpp o{\mkanchorpp{o'}t{att}}
=\mkanchorpp{\appendlistpp o{o'}}t{att}
\\\\
\combinelinkpp{a}{a'}{a''}{L}{L'} =
\\~~~~\mapsetpplong
    {\replaceaddrlink(\integratepp{a'},\integratepp{a''})}
    {\mapsetpplong
        {\replaceaddrlink(\integratepp{a},\integratepp{a''})}
        {\unionpp L{L'}}}
}

\comfundef
{\insertathmdpp h o{h'}{a''}=h''}
{$h''$ is the {\hmd} with address $a''$ after \math h 
 has been inserted at location \math o into {\hmd} \math{h'},
 provided that \math o exists in $h$.}
{\insertathmdpp h o{h'}{a''}=\insertathmdepp h o{h'}{a''}
\ \Longleftarrow\ \haslocationSpp o{\ldShdpp{h'}}=\true}}

\vfill\pagebreak
\subsection{Frameset Document Level}
\noindent\tightemph
{The following specifications are essentially
incomplete and have to be completed in the
future!!!}

\yestop\yestop\yestop
\subsubsection{Chapter}

\yestop\yestop\yestop\noindent
\SYMBOL{\LDTWO}~=
\vissortsect{
  \symbolisort{\ldtwosort}
}

\yestop\yestop\yestop\yestop\noindent
\LDTWO~=~\HDONE~\andspec~\SYMBOL{\LDTWO}~\andspec~{\ATTj \LDTWO}~\andspec\\
\phantom{\LDTWO~=~}\TREE[\renamesort\entrysort \structitypesort\ldtwosort]~\andspec\\
\phantom{\LDTWO~=~}\LIST[\renamesort\entrysort \ldtwosort]~\andspec\\
\phantom{\LDTWO~=~}\LIST[\renamesort\entrysort \natsort]~\thenspec
\vissortsect{ 
  \ldtwosort
\\\structitypesort\ldtwosort
\\\identsorts{\locisort{\hdtwosort}} {\mident{\listsort(\natsort)}}
}
\conssect{
  \hdhframeset,\hdvframeset,\hdaframeset~:~\structitypesort\hdtwosort
}

\yestop\yestop\yestop
\subsubsection{FrameSet Document}

\yestop\yestop\yestop\noindent
\ADDR{\HDTWO}~=~\STRING~\thenspec
\vissortsect{
  \addrisort{\hdtwosort}
}

\yestop\yestop\yestop\noindent
\begin{tabbing}
\HDTWO~=~ \= \= \HDCLASS \= \kill
\HDTWO~=~ \> \LDTWO~\andspec~\ADDR{\HDTWO}~\andspec\\
\>\> \HDCLASS [{\renamesort\basisparamsort {\LDTWO}.\ldtwosort},\\
\>\>\> {~\renamesort\locationparamsort {\LDTWO}.\locisort{\ldtwosort}},\\
\>\>\> {~\renamesort\embedlinkokS\LDTWO.\includelinkokldtwoS},\\
\>\>\> {~\renamesort\addrparamsort {\ADDR\HDTWO.\addrisort{\hdtwosort}}}]
\end{tabbing}
\thenspec
\vissortsect{ 
  \hdtwosort=
  \apptotuple\hdsort{\trip
    {\LDTWO.\ldtwosort}
    {\LDTWO.\locisort\ldtwosort}
    {\ADDR\HDTWO.\addrisort\hdtwosort}}}

\vfill\pagebreak
\subsection{Site Level}
\noindent\tightemph
{The following specifications are essentially 
incomplete and have to be completed in the
future!!!}

\yestop\yestop\yestop
\subsubsection{Book}

\yestop\yestop\yestop\noindent
\SYMBOL{\LDTHREE}~=
\vissortsect{
  \symbolisort{\ldthreesort}
}

\yestop\yestop\yestop\yestop\noindent
\LDTHREE~=~\HDTWO~\andspec~\SYMBOL{\LDTHREE}~\andspec~{\ATTj \LDTHREE}~\andspec\\
\phantom{\LDTHREE~=~}\TREE[\renamesort\entrysort \structitypesort\ldthreesort]~\andspec\\
\phantom{\LDTHREE~=~}\LIST[\renamesort\entrysort \ldthreesort]~\andspec\\
\phantom{\LDTHREE~=~}\LIST[\renamesort\entrysort \natsort]~\thenspec
\vissortsect{ 
  \ldthreesort
\\\structitypesort\ldthreesort
\\\identsorts{\locisort{\ldthreesort}} {\mident{\listsort(\natsort)}}
}
\conssect{
  \hdsitemap~:~\structitypesort\ldthreesort
}

\yestop\yestop\yestop
\subsubsection{Site}

\yestop\yestop\yestop\noindent
\ADDR{\HDTHREE}~=~\STRING~\thenspec
\vissortsect{
  \addrisort{\hdthreesort}
}

\yestop\yestop\yestop\noindent
\begin{tabbing}
\HDTHREE~=~ \= \= \HDCLASS \= \kill
\HDTHREE~=~ \> \LDTHREE~\andspec~\ADDR{\HDTHREE}~\andspec\\
\>\> \HDCLASS[{\renamesort\basisparamsort {\LDTHREE}.\ldthreesort},\\
\>\>\> {~\renamesort\locationparamsort {\LDTHREE}.\locisort{\ldthreesort}},\\
\>\>\> {~\renamesort\embedlinkokS{\LDTHREE.\includelinkokldthreeS}},\\
\>\>\> {~\renamesort\addrparamsort{\ADDR\HDTHREE.\addrisort{\hdthreesort}}}]
\end{tabbing}
\thenspec
\vissortsect{ 
  \hdthreesort=
  \apptotuple\hdsort{\trip
    {\LDTHREE.\ldthreesort}
    {\LDTHREE.\locisort\ldthreesort}
    {\ADDR\HDTHREE.\addrisort\hdthreesort}}}
\end{appendix}%

\flushbottom
\end{document}